\documentclass{article}%
\usepackage[utf8]{inputenc}%
\usepackage[x11names,dvipsnames,table]{xcolor}
\usepackage{hyperref}%
\hypersetup{%
colorlinks,%
linkcolor={red!50!black},%
citecolor={blue!50!black},%
urlcolor={blue!80!black}%
}%

\usepackage{amsmath}%
\usepackage{amssymb}%
\usepackage{amsfonts}%
\usepackage{amsthm}
\usepackage{nicefrac}
\usepackage{authblk}
\usepackage{doi}%
\usepackage{enumerate}%
\usepackage[toc,section=section]{glossaries}%
 \usepackage{fullpage}%
\usepackage{graphicx}%
\usepackage[square,comma,sort&compress]{natbib}%
\usepackage{hyperref}%
\usepackage{paralist}%
\usepackage{subfig}
\usepackage{url}%
\usepackage{comment}%
\usepackage{booktabs}%
\usepackage{pifont}%
\usepackage{tikz}
\usetikzlibrary{arrows,arrows.meta,babel,calc,intersections,backgrounds}
\usepackage{pgfplots}
\pgfplotsset{compat=1.14}

\newglossaryentry{nr}
{
  name={\ensuremath{n_r}},
  description={number of samples, repeats of one specific configuration},
  sort=nr
}
\newglossaryentry{A}
{
  name={\ensuremath{A}},
  description={optimization algorithm},
  sort=a
}
\newglossaryentry{a}
{
  name={\ensuremath{a}},
  description={optimization algorithm instance},
  sort=a
}

\newglossaryentry{p}
{
  name={\ensuremath{p}},
  description={problem, deterministic},
  sort=p
}
\newglossaryentry{pi}
{
  name={\ensuremath{\pi}},
  description={problem instance, random},
  sort=pi
}
\newglossaryentry{Pi}
{
  name={\ensuremath{\Pi}},
  description={problem class, i.e., set of problem instances},
  sort=Pi
}
\newglossaryentry{np}
{
  name={\ensuremath{n_p}},
  description={number of fixed problems $p$},
  sort=np
}
\newglossaryentry{npi}
{
  name={\ensuremath{n_{\pi}}},
  description={number of problem instance $\pi$},
  sort=npi
}
\newglossaryentry{na}
{
  name={\ensuremath{n_a}},
  description={number of algorithms $a$},
  sort=na
}
\newglossaryentry{nb}
{
  name={\ensuremath{n_b}},
  description={budget for one algorithm run, number of function evaluations},
  sort=nb
}
\newglossaryentry{ns}
{
  name={\ensuremath{n_s}},
  description={budget for the algorithm tuner},
  sort=ns
}
\newglossaryentry{Pr}
{
  name={\ensuremath{\text{Pr}}},
  description={probability},
  sort=Pr
}

\newglossaryentry{temp}
{
  name={\ensuremath{\texttt{temp}}},
  description={temperatur in simulated annealing, algorithm parameter},
  sort=temp
}
\newglossaryentry{y}
{
  name={\ensuremath{y}},
  description={algorithm performance},
  sort=y
}

\newacronym{aaai}{AAAI}{Association for the Advancement of Artificial Intelligence}
\newacronym{acm}{ACM}{Association for Computing Machinery}
\newacronym{aco}{ACO}{Ant Colony Optimization}
\newacronym{ANOVA}{ANOVA}{Analysis of Variance}
\newacronym{aslib}{ASlib}{Algorithm Selection Library}
\newacronym{bfgs}{BFGS}{Broyden-Fletcher-Goldfarb-Shanno}
\newacronym{bbob}{BBOB}{Black-Box-Optimization-Benchmarking}
\newacronym{cec}{CEC}{Congress on Evolutionary Computation}
\newacronym{cmaes}{CMA-ES}{Covariance Matrix Adaptation Evolution Strategy}
\newacronym{COCO}{COCO}{Comparing Continuous Optimizers} 
\newacronym{crn}{CRN}{Common Random Numbers}
\newacronym{crs4ea}{CRS4EA}{Chess Rating Systems for Evolutionary Algorithms}
\newacronym{dace}{DACE}{Design and Analysis of Computer Experiments}
\newacronym{DE}{DE}{Differential Evolution} 
\newacronym{doe}{DOE}{Design of Experiments}
\newacronym{ea}{EA}{Evolutionary Algorithm}
\newacronym{ec}{EC}{Evolutionary Computation}
\newacronym{ecdf}{ECDF}{Empirical Cumulative Distribution Function}
\newacronym{eda}{EDA}{Exploratory Data Analysis}
\newacronym{edalgo}{EDAlgo}{Estimation of Distribution Algorithm}
\newacronym{ego}{EGO}{Efficient Global Optimization}
\newacronym{ela}{ELA}{Exploratory Landscape Analysis}
\newacronym{ert}{ERT}{Expected Running Time}
\newacronym{es}{ES}{Evolution Strategy}
\newacronym{fe}{FE}{Function Evaluation}
\newacronym{ga}{GA}{Genetic Algorithm}
\newacronym{gecco}{GECCO}{Genetic and Evolutionary Computation Conference}
\newacronym{GenSA}{GenSA}{Generalized Simulated Annealing} 
\newacronym{GLG}{GLG}{Gaussian Landscape Generator}
\newacronym{gpd}{GPD}{Generalized Position-Distance}  
\newacronym{ieee}{IEEE}{Institute of Electrical and Electronics Engineers}
\newacronym{irace}{irace}{iterated racing}
\newacronym{labs}{LABS}{Low Auto-correlation Binary Sequence}
\newacronym{lhd}{LHD}{Latin Hypercube Design}
\newacronym{macoda}{MACODA}{Many Criteria Optimization and Decision Analysis}
\newacronym{maxsat}{MAX-SAT}{Maximum Satisfiability}
\newacronym{nflt}{NFLT}{no free lunch theorem}
\newacronym{MBO}{MBO}{Model-Based Optimization} 
\newacronym{MLE}{MLE}{Maximum Likelihood Estimation} 
\newacronym{moo}{MOO}{Multi-Objective Optimization}
\newacronym{nm}{NM}{Nelder Mead}
\newacronym{ofat}{OFAT}{One-factor-at-a-time}
\newacronym{paramils}{ParamILS}{Iterated Local Search in Parameter Configuration Space}
\newacronym{pbo}{PBO}{Pseudo-Boolean Optimization}
\newacronym{pdsc}{pDSC}{practical Deep Statistical Comparison}
\newacronym{PSO}{PSO}{Particle Swarm Optimization} 
\newacronym{roi}{ROI}{Region of Interest} 
\newacronym{sann}{SANN}{Simulated Annealing}
\newacronym{sat}{SAT}{Boolean Satisfiability}
\newacronym{smac}{SMAC}{Sequential Model-based Algorithm Configuration}
\newacronym{smoof}{smoof}{Single and Multi-Objective Optimization Test Functions} 
\newacronym{SPO}{SPO}{Sequential Parameter Optimization}
\newacronym{spot}{SPOT}{Sequential Parameter Optimization Toolbox}
\newacronym{tsp}{TSP}{Traveling Salesperson Problem}

\glstoctrue

\makeglossaries

\newtheorem{example}{Example}[section]

\colorlet{manuel}{OliveGreen}

\usepackage[normalem]{ulem}

\renewcommand\cite{\citep}

\title{Benchmarking in Optimization:\\ Best Practice and Open Issues}%
\author[1]{Thomas~Bartz-Beielstein}
\author[2]{Carola~Doerr} 
\author[3]{Daan~van~den~Berg} 
\author[6]{Jakob~Bossek} 
\author[1]{Sowmya~Chandrasekaran} 
\author[5]{Tome~Eftimov}
\author[1]{Andreas~Fischbach} 
\author[6]{Pascal~Kerschke}
\author[9]{William~La~Cava}
\author[7]{Manuel~L{\'o}pez-Ib{\'a}{\~n}ez} 
\author[8]{Katherine~M.~Malan}
\author[9]{Jason~H.~Moore}
\author[1]{Boris~Naujoks}
\author[9,10]{Patryk~Orzechowski} 
\author[11]{Vanessa~Volz}
\author[4]{Markus~Wagner}
\author[12]{Thomas~Weise}

\affil[1]{Institute for Data Science, Engineering, and Analytics, TH Köln, Germany}
\affil[2]{Sorbonne Universit\'e, CNRS, LIP6, Paris, France}
\affil[3]{Yamasan Science \& Education}
\affil[4]{Optimisation and Logistics, School of Computer Science, The University of Adelaide, Adelaide, Australia}
\affil[5]{Computer Systems Department, Jo\v{z}ef Stefan Institute, Ljubljana, Slovenia}
\affil[6]{Statistics and Optimization Group, University of M{\"u}nster, M{\"u}nster, Germany}
\affil[7]{School of Computer Science and Engineering, University of M{\'a}laga, M{\'a}laga, Spain}
\affil[8]{Department of Decision Sciences, University of South Africa, Pretoria, South Africa}
\affil[9]{Institute for Biomedical Informatics, University of Pennsylvania, Philadelphia, PA, USA}
\affil[10]{Department of Automatics, AGH University of Science and Technology, Krakow, Poland}
\affil[11]{modl.ai, Copenhagen, Denmark}
\affil[12]{Institute of Applied Optimization, School of Artificial Intelligence and Big Data, Hefei University, Hefei, China}

\affil[ ]{\textit {\href{mailto:benchmarkingbestpractice@gmail.com}{benchmarkingbestpractice@gmail.com}}}

\begin{document}
\maketitle
\vspace{-1cm}
\begin{center}
Version 2
\end{center}

\begin{abstract}
This survey compiles ideas and recommendations from more than a dozen researchers with different backgrounds and from different institutes around the world.
Promoting best practice in benchmarking is its main goal. 
The article discusses eight essential topics in benchmarking: clearly stated goals, well-specified problems, suitable algorithms, adequate performance measures, thoughtful analysis, effective and efficient designs, comprehensible presentations, and guaranteed reproducibility.
The final goal is to provide well-accepted guidelines (rules) that might be useful for authors and reviewers.
As benchmarking in optimization is an active and evolving field of research this manuscript is meant to co-evolve over time by means of periodic updates. 
\end{abstract}

\clearpage

\tableofcontents

\sloppy


\newpage

\section{Introduction}
\label{sec:sec01}

Introducing a new algorithm without testing it on a set of benchmark functions appears to be very strange to every optimization practitioner, unless there is a strong theoretical motivation justifying the interest in the algorithm. 
Taking theory-focused papers aside, from the very beginning in the 1960s nearly every publication in \gls{ec} was accompanied by benchmarking studies. 
One of the key promoters of the \gls{ec} research domain, Hans-Paul \citet{Schw75a}, wrote in his PhD thesis: 
\begin{quote}
The extremely large and constantly increasing number of optimization methods inevitably leads to the question of the best strategy. 
There does not seem to be a clear answer. Because, if there were an optimal optimization process, all other methods would be superfluous \ldots\footnote{German original quote: ``Die überaus große und ständig steigende Zahl von Optimierungsmethoden führt zwangsläufug zu der Frage nach der besten Strategie. Eine eindeutige Antwort scheint es nicht zu geben. Denn, gäbe es ein optimales Optimierungsverfahren, dann würden sich alle anderen Methoden erübrigen\ldots
''}
\end{quote}
Famous studies, e.g., from~\citet{More81a}, were performed in this period and established test functions that are today well known among algorithm developers. Some of them can still be found in the portfolio of recent benchmark studies, e.g., Rosenbrock's function~\citep{Rose60a}.
In the 1960s, experiments could be rerun only a very limited number of times, using different starting points or random seeds. This situation has changed drastically: nowadays, new algorithms can be run a hundred or even a thousand times. This enables very complex and sophisticated benchmark suites such as those available in the \gls{COCO}~\citep{cocoplat} platform or in Nevergrad~\cite{nevergrad}. However, the questions to be answered by benchmarking remain basically the same, e.g., 
\begin{itemize}
    \item how well does a certain algorithm perform on a given problem?
     \item why does an algorithm succeed/fail on a specific test problem?
\end{itemize}
Specifying the goal of a benchmark study is as important as the study itself, as it shapes the experimental setup -- i.e., the choice of problem instances, of the algorithm instances, the performance criteria, and the statistics. Typical goals that a user or a researcher wishes to answer through a benchmarking study are discussed in \autoref{sec:sec2}.

But not only computational power has increased significantly in the last decades. Theory made important progress as well.
In the 1980s, some researchers claimed that there is an algorithm that is able to outperform all other algorithms on average~\citep{Gol89a}.
A set of \glspl{nflt}, presented by \citet{Wolp97a} changed this situation~\citep{Deme19a}.
Statements about the performance of algorithms should be coupled with the problem class or even the problem instances. 
\citet{Brow07a} summarizes \gls{nflt} consequences and gives the following recommendations:  
\begin{quote}
1) bound claims of algorithm or parameter suitability to the problem instances being tested, \\
2) research into devising problem classes and matching suitable algorithms to classes is a good thing, \\
3) be cautious about generalizing performance to other problem instances, and \\
4) be very cautious about generalizing performance to other problem classes or domains.
\end{quote}
\citet{Haft16b} describes \gls{nflt} consequences as follows:
\begin{quote}
Improving an algorithm for one class of problem is likely to make it perform more poorly for other problems.
\end{quote}
Some authors claim that this statement is too general and should be detailed as follows: 
 improving the performance of an algorithm, e.g., via parameter tuning, for a subset of problems may make it perform worse for a different subset. This does not work so well for classes of problems, unless the classes are finite and small. It also does not work for any two arbitrary subsets, since the subsets may be correlated in precisely the way that leads to better performance of the algorithm.
A number of works discuss limitations for the consequences and the impact of \gls{nflt}, such as \citet{GarRodLoz2012soco} and \citet{McDermott2020nfl}.
For example, \citet{Cul1998ec} stated:  ``In the context of search problems, the NFL theorem strictly only applies if
arbitrary search landscapes are considered, while the instances of basically
any search problem of interest have compact descriptions and therefore cannot
give rise to arbitrary search landscapes ''.
 
Without doubt, \gls{nflt} has changed the way how benchmarking is \emph{considered}\/ in \gls{ec}.
Problems caused by \gls{nflt} are still subject of current research, e.g., \citet{Liu19b} discuss paradoxes in numerical comparison of optimization algorithms based on \gls{nflt}. 
\citet{Whit02b} examine the meaning and significance of benchmarks in light of theoretical results such as \gls{nflt}. 

Independently of the ongoing \gls{nflt} discussion, benchmarking gains a central role in current research, both for theory and practice. 
Three main aspects that need to be addressed in every benchmark study are the choice of 
\begin{enumerate}
\item the performance measures, 
\item the problem (instances), and 
\item the algorithm (instances).
\end{enumerate}

Excellent papers on how to set up a good benchmark test exist for many years. Hooker and Johnson are only two authors that published papers still worth reading today~\citep{Hook94a, Hoo1996joh, John89a, John91a, John96a}. 
\citet{McGe86a} can be considered as a milestone in the field of experimental algorithmics, which builds the cornerstone for benchmark studies.
\citet{Gent94a} stated that the empirical study of algorithms is a relatively immature field -- and we claim that this situation has unfortunately not significantly changed in the last 25 years.
Reasons for this unsatisfactory situation in \gls{ec} are manifold. 
For example, \gls{ec} has not agreed upon general methodology for performing benchmark studies like the fields of statistical \gls{doe} or data mining~\citep{Chap00a, Mont17a}. These fields provide a general methodology to encourage the practitioner to consider important issues before performing a study. 
Some journals provide explicit minimal standard requirements.\footnote{See \url{https://www.springer.com/cda/content/document/cda_downloaddocument/Journal+of+Heuristic+Policies+on+Heuristic+Search.pdf?SGWID=0-0-45-1483502-p35487524} for guidelines of the \emph{Journal of Heuristics} and \url{https://static.springer.com/sgw/documents/1593723/application/pdf/Additional_submission_instructions.pdf} for similar ones of the journal \emph{Swarm Intelligence}.} 

The question remains: why are minimum standards not considered in every paper submitted to \gls{ec} conferences and journals? Or, formulated alternatively: why have such best practices not become minimum required standards?
One answer might be: setting up a sound benchmark study is very complicated. There are many pitfalls, especially stemming from complex statistical considerations~\citep{Crep14a}. So, to do nothing wrong, practitioners oftentimes report only average values decorated with corresponding standard deviations, $p$-values, or boxplots.
Another answer might be: practical guidelines are missing. Researchers from computer science would apply these guidelines if examples were available. 
This paper is a joint initiative from several researchers in \gls{ec}. It presents best-practice examples with references to relevant publications and discusses open issues.
This joint initiative was established during the \href{https://www.dagstuhl.de/en/program/calendar/semhp/?semnr=19431}{Dagstuhl seminar 19431} on \emph{Theory of Randomized Optimization Heuristics}, which took place in October 2019.
Since then, we have been compiling ideas covering a broad range of disciplines, all connected to \gls{ec}.

We are aware that every version of this paper represents a snapshot, because the field is evolving. New theoretical results such as no-free lunch might come up from theory and new algorithms (quantum computing, heuristics supported by deep learning techniques, etc.) appear on the horizon, and new measures, e.g., based on extensive resampling (Monte Carlo), can be developed in statistics.

We consider this paper as a starting point, as a first trial to support the EC community in improving the quality of benchmark studies.
Surely, this paper cannot cover every single aspect related to benchmarking. 
Although this paper mainly focuses on single-objective, unconstrained optimization problems, its findings can be easily transferred to other domains, e.g, multi-objective or  constrained optimization. The objectives in other problem domains may differ slightly and may require different performance measures -- but the content of most sections  should be applicable.
Each of the following sections presents references to best-practice examples and discusses open topics.
The following aspects, which are considered relevant to every benchmark study, are covered in the subsequent sections:
\begin{enumerate}
    \item Goals: what are the reasons for performing benchmark studies (Section~\ref{sec:sec02})? 
    \item Problems: how to select suitable problems (Section~\ref{sec:sec3})?  
    \item Algorithms: how to select a portfolio of algorithms to be included in the benchmark study (Section~\ref{sec:sec4})?  
    \item Performance: how to measure performance (Section~\ref{sec:sec5})?  
    \item Analysis: how to evaluate results (Section~\ref{sec:sec6})? 
    \item Design: how to set up a study, e.g., how many runs shall be performed (Section~\ref{sec:07design})?  
    \item Presentation: how to describe results (Section~\ref{sec:08presentation})? 
    \item Reproducibility: how to guarantee scientifically sound results and how to guarantee a lasting impact, e.g., in terms of comparability (Section~\ref{sec:09reproducibility})?  
\end{enumerate}
The paper closes with a summary and an outlook in Section~\ref{sec:sec10}.

\paragraph{Generalization of benchmarking results.}  
As discussed above in the context of the \gls{nflt}, we recommend being very precise in the description of the algorithms and the problem instances that were used in the benchmark study. Performance extrapolation or generalization always needs to be flagged as such, and where algorithms are compared to each other, it should be made very clear what the basis for the comparison is. We suggest to very carefully distinguish between algorithms (e.g., ``the'' \gls{cmaes}~\cite{Hansen2000invariance}) and algorithm instances (e.g., the pycma-es~\cite{pycmaes} with population size 8, budget 100, restart strategy X, etc.).\footnote{Algorithm instances are also referred to as ``algorithm configurations'' in the literature~\citep{BirStuPaqVar02:gecco}.} 
A similar rule applies to the problems (e.g., ``the'' sphere function) vs. a concrete  problem instance (the five-dimensional sphere function $f:\mathbb{R}^5 \to \mathbb{R}, x \mapsto \alpha \sum_{i=1}^d{x_i^2}+\beta$ centered at at the origin, multiplicative scaling $\alpha$, and additive shift $\beta$). 
Going one step further, one may even argue that we only benchmark a certain \emph{implementation} of an algorithm instance, which is subject to a concrete choice of implementation language, compiler and operating system optimizations, and concrete versions of software libraries.

Algorithm and problem instances may be (and in the context of this survey often are) randomized, so that the performance of the algorithm instance on a given problem instance is a series of (typically highly correlated) random variables, one for each step of the algorithm. In practice, replicability is often achieved by fixing the random number generator and storing the random seed, which plays an important role in guaranteeing reproducibility as discussed in Sec.~\ref{sec:09reproducibility}.


\section{Goals of Benchmarking Activities}
\label{sec:sec02}
\label{sec:sec2} 

\begin{figure}[t]
    \centering
    \includegraphics[width=0.9\textwidth]{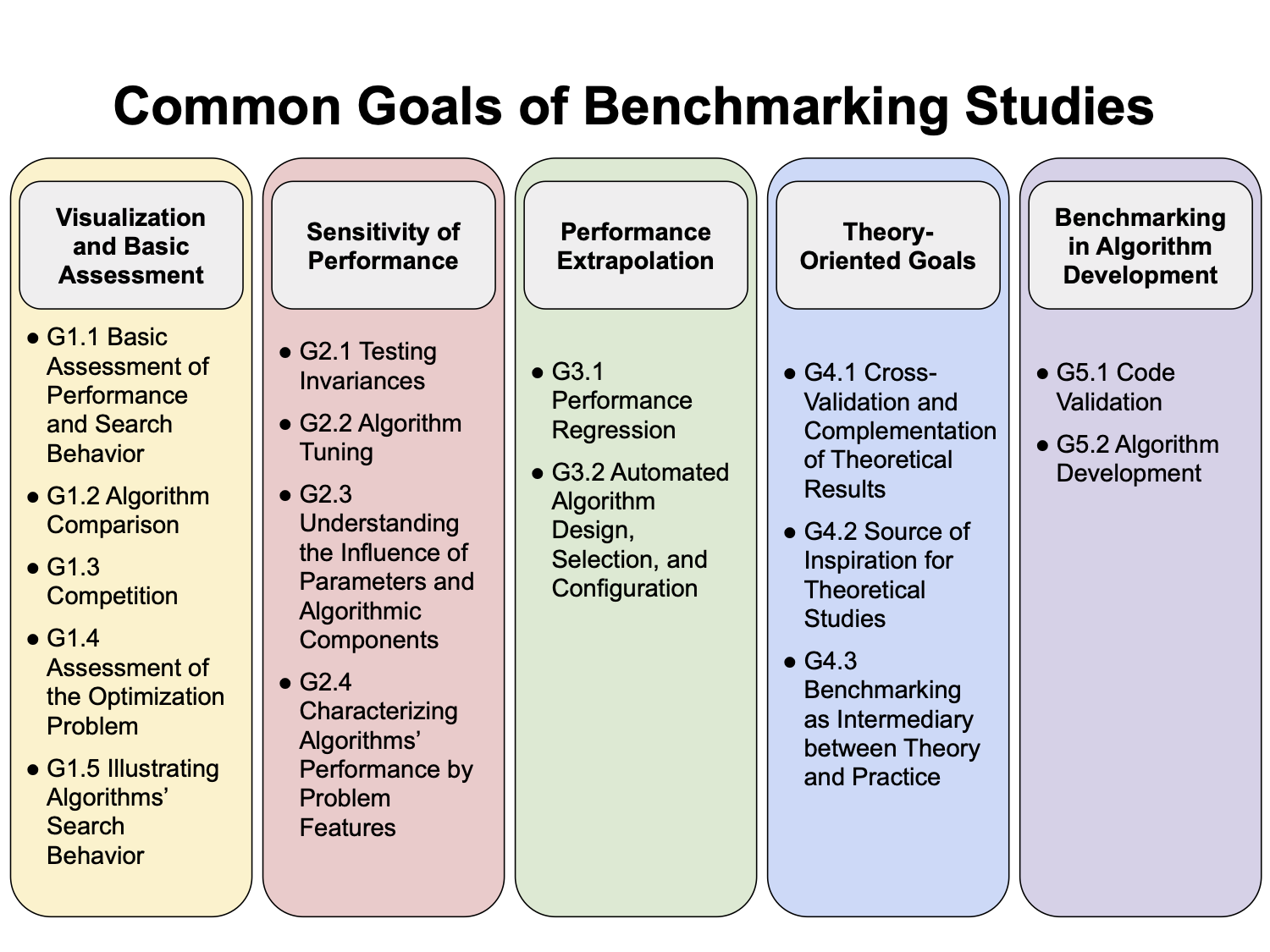}
    \caption{Summary of common goals of benchmark studies.}
    \label{fig:goals}
\end{figure}

The motivations for performing benchmark studies on optimization algorithms are as diverse as the algorithms and the problems that are being used in these studies. 
Apart from scientifically motivated goals, benchmarking can also be used as a means to popularize an algorithmic approach or a particular problem. In this section, we focus on summarizing the most common scientifically-motivated goals for benchmarking studies.  
Figure~\ref{fig:goals} summarizes these goals. The relevance of these goals can differ from study to study, and the proposed categorization is not necessarily unique, but should be understood as an attempt to find something that represents the benchmarking objectives well within the broader scientific community. Most of the goals listed below are, ultimately, aimed at contributing towards a better deployment of the algorithms in practice, typically through a better understanding of the interplay between the algorithmic design choices and the problem instance characteristics. However, benchmarking also plays an important role as intermediary between the scientific community and users of optimization heuristics and as intermediary between theoretically and empirically-guided streams within the research community.    

\subsection{Visualization and Basic Assessment of Algorithms and Problems}

\begin{enumerate}[({G1}.1)]
    
\item 
\label{goal:basic}
\emph{Basic Assessment of Performance and Search Behavior.}\newline The arguably most basic research question that one may want to answer with a benchmark study is how well a certain algorithm performs on a given problem. In the absence of mathematical analyses, and in the absence of existing data, the most basic approach to gain insight into the performance is to run one or more instances of the algorithm
(ideally several times, if the algorithm or the problem are stochastic) on one or more problem instances,
and to observe the behavior of the algorithm. With this data, one can analyze what a typical performance profile looks like on some problem instance, how the solution quality evolves over time,
 how robust the performance is, etc.
 The evaluation criteria can be diverse, as we shall discuss in Section~\ref{sec:performance}. But what is inherent to all studies falling into this goal~G1.\ref{goal:basic},
is that they are aimed to answer a rather basic question 
``How well does this particular algorithm perform on this particular problem instance?'' or ``How does a particular run of this algorithm on the given problem look like?''.
\item 
\label{goal:algorithm}
\emph{Algorithm Comparison.} \newline  The great majority of benchmark
  studies do not focus on a single algorithm, but rather
  \emph{compare} the performance and/or the search behavior of two or more algorithms. The comparison of algorithms serves, most notably, the purpose of understanding strengths and weaknesses of different algorithmic approaches for different types of problems or problem instances  during the different stages of the optimization process. These insights can be leveraged to design or to select, for a given problem class or instance, a most suitable algorithm instance.
\item 
\label{goal:competition}
\emph{Competition.} \newline One particular motivation to compare algorithms is to determine a ``winner'', i.e., an algorithm that performs better than any of its competitors, for a given performance measure and on a given set of problem instances.  Benchmarking is of great value in selecting the most adequate algorithm especially in real-world optimization settings~\citep{Beir17a}. The role of competitive studies for benchmarking is discussed quite
controversially~\citep{Hoo1996joh},
as competitive studies may promote algorithms that overstate the importance of the problems that they are tested upon, and thereby create over-fitting. At the same time, however, one cannot neglect that competitions can provide an important incentive to contribute to the development of new algorithmic ideas and better selection guidelines.
\item \emph{Assessment of the Optimization Problem.} In many real-world problems like scheduling, container packing, chemical plant control, or protein folding, the global optimum is unknown, while in other problems it is necessary to deal with limited knowledge, or lack of explicit formulas. In those situations, computer simulations or even physical experiments are required to evaluate the quality of a given solution candidate. In addition, even if a problem is explicitly modelled by a mathematical formula, it can nevertheless be difficult to grasp its structure or to derive a good intuition for what its fitness landscape looks like. Similarly, when problems consist of several instances, it can be difficult to understand in what respect these different instances are alike and in which aspects they differ. Benchmarking simple optimization heuristics can help to analyze and to visualize the optimization problem and to gain knowledge about its  characteristics. 

\item \emph{Illustrating Algorithms' Search Behavior.} \newline  
Understanding how an optimization heuristic operates on a problem can be difficult to grasp when only looking at the algorithm and problem description. One of the most basic goals that benchmarking has to offer are numerical and graphical illustrations of the optimization process.
With these numbers and visualizations, a first idea about the optimization process can be derived. This also includes an assessment of the stochasticity when considering several runs of a randomized algorithm or an algorithm operating upon a stochastic problem. In the same vein, benchmarking offers a hands-on way of visualizing effects that are difficult to grasp from mathematical descriptions. That is, where mathematical expressions are not easily accessible to everyone, benchmarking can be used to illustrate the effects that the mathematical expressions describe. 
\end{enumerate}

\subsection{Sensitivity of Performance in Algorithm Design and Problem Characteristics}

\begin{enumerate}[({G2}.1)]
  \item \emph{Testing Invariances.} \newline
  Several researchers argue that, ideally, the performance of an optimization algorithm should be invariant with respect to certain aspects of the problem embedding, such as the scaling and translation of the function values ~\citep{vrielink2019fireworks}, dimensional increase ~\citep{Dejonge2020sensitivity}, or a rotation of the search space (see~\citet{Hansen2000invariance} and references therein for a general discussion and \citet{LehreW12,ABB} for examples formalizing the notion of \emph{unbiased} algorithms). 
    
    Whereas certain invariances, such as comparison-baseness, are typically easily inferred from a pseudocode description of the algorithm, other invariances (e.g., invariance with respect to translation or rotation) might be harder to grasp. 
    In such cases, benchmarking can be used to test, empirically, whether the algorithm possesses the desired invariances. 

\item \emph{Algorithm Tuning.} 
\label{goal:tuning}
\newline
Most optimization heuristics are configurable, i.e, 
    we are able to adjust their search behavior (and, hence, performance) by modifying 
    their parameters. 
   Typical parameters of algorithms are the number of individuals kept
  in the memory (its `population size'), the number of individuals that are evaluated
  in each iteration, parameters determining the distribution
  from which new samples are generated (e.g., the mean, variance, and direction of the search), the selection of survivors for the next generation's population, and the stopping criterion. Optimization heuristics applied in practice often comprise tens of parameters that need to be tuned. 
    
    Finding the optimal configuration of an algorithm for a given problem instance is referred to as
    offline parameter tuning~\citep{Eibe02a, eibe03a}. Tuning can be done manually or with the help of automated configuration tools ~\citep{bergstra2013hyperopt, olson2016tpot, akiba2019optuna},
 Benchmarking is a core ingredient of the parameter tuning process. A proper design of experiment is an essential requirement for tuning studies~\citep{Bart06a, orzechowski2018we, orzechowski2020benchmarking}. 
     Parameter tuning is a necessary step before comparing a viable configuration of a method with others, as we are disregarding those combinations of parameters, which do not yield promising results.
     
 Benchmarking can help to shed light on suitable choices of parameters and algorithmic modules.
 Selecting a proper parameterization for a  given optimization problem is a tedious task~\citep{FialhoCSS10}. Besides the selection of the algorithm and the problem instance, tuning 
 requires the specification of a \emph{performance measure}, e.g., best solution 
 found after a pre-specified number of function evaluations  (to be discussed in Sec.~\ref{sec:performance}) 
 and a \emph{statistic}, i.e., number of repeats, which will be discussed in Sec.~\ref{sec:07design}.
 
Another important concern with respect to algorithm tuning is the robustness of the performance with respect to these parameters, i.e., how much does the performance deteriorate if the parameters are mildly changed? In this respect, parameter recommendations with a better robustness might be preferable over less robust ones, even if compromising on performance~\citep{Paen06a}. 

  \item \emph{Understanding the Influence of Parameters and Algorithmic
    Components.} \label{goal:understanding}
    \newline
        While algorithm tuning focuses on finding the best configuration for a given problem, 
    \emph{understanding} refers to the question: \emph{why} does one algorithm  perform better 
    than a competing one? 
    Understanding requires additional statistical tools, e.g., analysis of variance or regression techniques.
    Questions such as ``Does recombination have a significant effect on the performance?'' are considered in this approach. Several tools that combine methods from statistics and visualization are integrated in the software package \gls{spot}, which was designed for understanding the behavior of optimization algorithms. \gls{spot} provides a set of tools for model based optimization and tuning of algorithms. It includes surrogate models, optimizers and \gls{doe} approaches~\citep{Bart17parxiv}.
    
\item \emph{Characterizing Algorithms' Performance by Problem (Instance) Features and Vice Versa.} \label{goal:characterization}
\newline
Whereas \emph{understanding} as discussed in the previous paragraph tries to get a deep insight into the
elements and working principles of algorithms, \emph{characterization} refers to the 
relationship between algorithms and problems. That is, the goal is to link features of the problem with the performance of the algorithm(s). 
A classical example for a question answered by the characterization approach is how the performance of an algorithm scales with the number of decision variables. 

Problem instance features can be high-level features such as its dimensionality, its search constraints, its search space structure, and other basic properties of the problem. Low-level features of the problem, such as its multi-modality, its separability, or its ruggedness can either be derived from the problem formulation or via an exploratory sampling approach~\citep{Mers10a, mersmann_exploratory_2011, flacco2019, KerschkeT19, 13MAL00, munoz_exploratory_2015, munoz2015_as}.  

 \end{enumerate}

\subsection{Benchmarking as Training: Performance Extrapolation}
 
\begin{enumerate}[({G3}.1)]
    \item \emph{Performance Regression.} \newline
    The probably most classical hope associated with benchmarking is that the generated data can be used to extrapolate the performance of an algorithm for other, not yet tested problem instances. This extrapolation is highly relevant for \emph{selecting} which algorithm to choose and how to \emph{configure} it, as we shall discuss in the next section. Performance extrapolation requires a good understanding of how the performance depends on problem characteristics, the goal described in G2.\ref{goal:characterization}. 
    
    In the context of machine learning, performance extrapolation is also referred to as \textit{transfer learning}~\citep{Pan10a}. It can be done manually or via sophisticated regression techniques. 
    Regardless of the methodology 
    used to extrapolate performance data, an important aspect 
    in this regression task is a proper selection of the instances on which the algorithms/configurations are tested.
    For performance extrapolation based on supervised learning approaches, a suitable selection of feature extraction methods is another crucial requirement for a good fit between extrapolated and true performance.

  \item \emph{Automated Algorithm Design, Selection, and Configuration.} 
  \newline
  When the dependency of algorithms' performance with respect to relevant problem characteristics is known and performance can be reasonably well extrapolated to previously unseen problem instances, the benchmarking results can be used for designing, selecting, or configuring an algorithm for the problem at hand. That is, the goal of the benchmark study is to provide training data from which rules can be derived that help the user choose the best algorithm for her optimization task. These guidelines can be human-interpretable such as proposed in~\citet{Bart06a, shiwa} or they can be implicitly derived by AutoML techniques~\cite{HutterBook19,kerschke_automated_2019,KerschkeT19, olson2016tpot}. 
 \end{enumerate}

\subsection{Theory-Oriented Goals}

\begin{enumerate}[({G4}.1)]
\item \emph{Cross-Validation and Complementation of Theoretical Results.}
\label{goal:crossvalidation}
\newline
  Theoretical results in the context of optimization are often expressed in terms of asymptotic running time bounds~\cite{DoerrN20,AugerD11,NeumannW10}, so that it is typically not possible
  to derive concrete performance values from them, e.g., for a concrete dimension, target values, etc. To analyze the behavior
  in small dimensions and/or to extend the regime for which the theoretical
  bounds are valid, a benchmarking study can be used to complement existing
  theoretical results.  
\item\emph{Source of Inspiration for Theoretical Studies.} 
\label{goal:source}
\newline
Notably,
  empirical results derived from benchmarking studies are an important
  source of inspiration for theoretical works. In particular when
  empirical performance does not match our intuition, or when we
  observe effects that are not well understood by mathematical means,
  benchmarking studies can be used to pinpoint these effects, and to
  make them accessible to theoretical studies, see~\cite{DoerrDL19} for an example. 
\item\emph{Benchmarking as Intermediary between Theory and Practice.}
\newline
  The last two goals, G4.\ref{goal:crossvalidation}  and G4.\ref{goal:source}, 
  together with G1.\ref{goal:basic} and G1.\ref{goal:algorithm} 
  highlight the role of benchmarking as an important
  intermediary between empirically-oriented and 
  mathematically-oriented sub-communities within the domain of heuristic
  optimization~\citep{Muel10a}. In this sense, benchmarking plays a similar role for optimization heuristics as \emph{Algorithm Engineering}~\citep{KliemannS16} does for classical algorithmics.  
\end{enumerate}

\subsection{Benchmarking in Algorithm Development}

\begin{enumerate}[({G5}.1)]
    \item \emph{Source Code Validation.} 
    \newline
    Another important aspect of benchmarking is that it can be used to verify that a given program performs as it is expected to. To this end, algorithms can be assessed on problem instances with known properties. If the algorithm consistently does not behave as expected, a source code review might be necessary. 
\item \emph{Algorithm Development.} \newline
In addition to
    \emph{understanding} performances, benchmarking is also used to
    identify weak spots with the goal to develop better performing
    algorithms. This also includes first empirical comparisons of new
    ideas to gain first insights into whether or not it is worth investigating
    further. This can result in a loop of empirical and theoretical
    analysis.  
    A good example for this is parameter control: it has been observed
    early on that a dynamic choice of algorithms' parameters can be
    beneficial over static ones~\citep{Kara15b}. 
    This led to the above mentioned loop of evaluating parameters empirically and stimulated theoretical investigations. 

\end{enumerate}

\subsection{Open Issues and Challenges} 
Several of the goals listed above require fine-grained records about the traces of an algorithm, raising the issue of storing, sharing, and re-using the data from the benchmark studies. Several benchmark environments offer a data repository to allow users to re-use previous experimental results. However, compatibility between the data formats of different platforms is rather weak, and a commonly agreed-upon standard would be highly desirable, both for a better comparability and for a resource-aware benchmarking culture. As long as such standards do not exist, tools that can flexibly interpret different data formats can be used. For example, the performance assessment module IOHanalyzer of the IOHprofiler benchmarking environment~\cite{IOHprofiler} can deal with various different formats, including those from the two most widely adopted benchmarking environments in \gls{ec}, Nevergrad~\cite{nevergrad} and \gls{COCO}~\cite{cocoplat}. 

Coming back to a resource-aware benchmarking culture, we repeat a statement already made in the introduction: two of the most important steps of a benchmarking study are the formulation of a clear research question that shall be answered, and the design of an experimental setup that best answers this question through a well-defined set of experiments. It is often surprising to see how many scientific reports do not clearly explain the main research question that the study aims to answer, (n)or how the reported benchmarking data supports the main claims.   

Finally, we note that also the goals themselves undergo certain ``trends'', which are not necessarily stable over time. The above collection of goals should therefore be seen as a snapshot of what we observe today, some of the goals mentioned above may gain or lose in relevance. 

\section{Problem Instances}
\label{sec:sec3}
A critical element of algorithm benchmarking is the choice of problem instances, because it can heavily influence the results of the benchmarking. Assuming that we (ultimately) aim at solving real-world problems, ideally, the problem set should be representative of the real-world scenario under investigation, otherwise it is not possible to derive general conclusions from the results of the benchmarking. In addition, it is important that problem sets are continually updated to prevent the over-tuning of algorithms to particular problem sets. 

This section discusses various aspects related to problem sets used in benchmarking. The four questions we address are: 
\begin{enumerate}
\item What are the desirable properties of a good problem set?
\item How to evaluate the quality of a problem set?
\item What benchmark problem sets are publicly available?
\item What are the open problems in research related to problem sets for benchmarking?
\end{enumerate}

\subsection{Desirable Characteristics of a Problem Set}
This section describes some of the general properties that affect the usefulness of suites of problems for benchmarking, see~\citet{WhitleyRDM96} and \citet{ShirDB18} for position statements. 

\begin{enumerate}[({B1}.1)]
\item
\emph{Diverse.}\newline A good benchmark suite should contain problems with a range of difficulties \cite{olson2017pmlb}. However, what is difficult for one algorithm could be easy for another algorithm and for that reason, it is desirable for the suite to contain a wide variety of problems with different characteristics. In this way, a good problem suite can be used to highlight the strengths and weaknesses of different algorithms. Competition benchmark problems are frequently distinguished based on a few simple characteristics such as modality and separability, but there are many other properties that can affect the difficulty of problems for search \citep{13MAL00, flacco2019, munoz2015_as} and the instances in a problem suite should collectively capture a wide range of characteristics. 

\item
\emph{Representative.}\newline At the end of a benchmarking exercise, claims are usually made regarding algorithm performance. The more representative the benchmarking suite is of the class of problems under investigation, the stronger the claim about algorithm performance will hold. The problem instances should therefore include the difficulties that are typical of real world instances of the problem class under investigation.  

\item
\emph{Scalable and tunable.}\newline Ideally a benchmark set/framework includes the ability to tune the characteristics of the problem instances. For example, it could be useful to be able to set the dimension of the problem, the level of dependence between variables, the number of objectives, and so on.

\item
\emph{Known solutions / best performance.}\newline If the optimal solution(s) of a benchmark problem are known, then it makes it easier to measure exact performance of algorithms in relation to the known optimal performance. There are, however, simple problems for which optimal solutions are not known even for relatively small dimensions (e.g. the \gls{labs} problem~\cite{Packebusch_2016_LABS}). In these cases it is desirable to have the best known performance published for particular instances.
\end{enumerate}

\subsection{Evaluating the Quality of a Problem Set}
Although it is trivial to assess whether a problem suite provides information on the optimal solution or is tunable, it is not as obvious to assess whether a problem set is diverse or representative. 
In this section, we provide a brief overview of existing ways of evaluating the quality of problem sets.

\begin{enumerate}[({B2}.1)]

\item
\emph{Feature space.}
One of the ways of assessing the diversity of a set of problem instances is to consider how well the instances cover a range of different problem characteristics. When these characteristics are measurable in some way, then we can talk about the instances covering a wide range of feature values. \citet{GARD2014} use a self-organizing feature map to cluster and analyse the \gls{bbob} and \gls{cec} problem sets based
on fitness landscape features (such as ruggedness and the presence of multiple funnels). In a similar vein, \citet{SEK2020UTPS} use \gls{ela} features~\citep{mersmann_exploratory_2011} combined with clustering and a $t$-distributed stochastic neighbor embedding visualization approach to analyse the distribution of problem instances across feature space.

\item
\emph{Performance space.}\newline Simple statistics such as mean and best performance aggregate much information without always enabling the discrimination of two or more algorithms. For example, two algorithms can be very similar (and thus perform comparably) or they might be structurally very different but the aggregated scores might still be comparable. From the area of algorithm portfolios, we can employ ranking-based concepts such as the marginal contribution of an individual algorithm to the total portfolio, as well as the Shapley values, which consider all possible portfolio configurations~\citep{Frechette2016shapley}. Still, for the purpose of benchmarking and better understanding of the effect of design decisions on an algorithm's performance, it might be desirable to focus more on instances that enable the user to tell the algorithms apart in the performance space.

This is where the targeted creation of instances comes into play. Among the first articles that evolved small \gls{tsp} instances that are difficult or easy for a single algorithm is that by \citet{Mersmann2013tspinstances}, which was then followed by a number of articles also in the continuous domain as well as for constrained problems. Recently, this was extended to the explicit discrimination of pairs of algorithms for larger \gls{tsp} instances~\cite{Bossek2019tspEvolve}, which required more disruptive mutation operators. 

\item
\emph{Instance space.}\newline Smith-Miles and colleagues~\citep{ST2012MAF} introduced a methodology called {\it instance space analysis}\footnote{\url{https://matilda.unimelb.edu.au/matilda/}} for visualizing problem instances based on features that are correlated with difficulty for particular algorithms. This approach can be seen as combining problem feature and algorithm performance into a single space. Regions of good performance (so called `footprints') in the instance space indicate the types of problems that specific algorithms can relatively easily solve. Visualizations of instance spaces can also be useful for indicating the spread of a set of problem instances across the space of features and can therefore be used to assess whether a benchmark suite covers a diverse range of instances for the algorithms under study. Example applications of the methodology include analysis of the \gls{tsp}~\citep{ST2012MAF} and continuous black-box optimisation problems \citep{MS2017PAOC}. An interesting extension of the approach is to evolve problem instances that fill the gaps in instance space left open by existing problem instances~\cite{SB2015GNTI}, or to directly evolve diverse sets of instances~\cite{Neumann2019diversity}.

\end{enumerate}

\subsection{Available Benchmark Sets}
Over the years, competitions and special sessions at international conferences have provided a wealth of resources for benchmarking of optimization algorithms. Some studies on metaheuristics have also made problems available to be used as benchmarks. This section briefly outlines some of these resources, mostly in alphabetical order of their key characteristic. We have concentrated on benchmark problems that are fundamentally different in nature, and that have documentation and code available online. It is also due to our focus on fundamental differences that we typically do not go into the details of configurable instances and parameterized instance generators.

It is worth mentioning upfront that a number of the benchmark problems mentioned below are available within the optimization software platform Nevergrad~\cite{nevergrad}.

\begin{enumerate}[({B3}.1)]
\item
\emph{Artificial discrete optimization problems.}\newline 
Subjectively, this area is among those with the largest number of benchmark sets. Here, many are inspired by problems encountered in the real world, which then have given rise to many fundamental problems in computer science. Noteworthy subareas of discrete optimization are combinatorial optimization, integer and constraint programming---and for many of them large sets of historically grown sets of benchmarks exist. 
Examples include the Boolean satisfiability and maximum satisfiability competitions\footnote{\url{http://www.satcompetition.org/}, \url{https://maxsat-evaluations.github.io/}}, travelling salesperson problem library\footnote{\url{http://comopt.ifi.uni-heidelberg.de/software/TSPLIB95/} and \url{http://www.math.uwaterloo.ca/tsp/index.html}} and the mixed integer programming library of problems\footnote{\url{https://miplib.zib.de/}}. 

In contrast to these instance-driven sets are the more abstract models that define variable interactions at the lowest level (i.e., independent of a particular problem) and then construct an instance based on fundamental characteristics. Noteworthy examples here are (for binary representations) the NK landscapes~\citep{93KAU00} (which has the idea of tunable ruggedness at its core), the W-Model~\citep{Weise2018wmodel} (with configurable features like length, neutrality, epistasis, multi-objectivity, objective values, and ruggedness), and the \gls{pbo} suite of 23 binary benchmark functions by \citet{DoerrYHWSB20}, which covers a wide range of landscape features and which extends the W-model in various ways (in particular, superposing its transformations to other base problems).

\item
\emph{Artificial real-parameter problems.}\newline Benchmark suites have been defined for special sessions, workshops and competitions at both the \gls{acm} \gls{gecco} and the \gls{ieee} \gls{cec}. Documentation and code are available online---for \gls{gecco} \gls{bbob}\footnote{\url{https://coco.gforge.inria.fr}}, and for \gls{cec}\footnote{\url{https://github.com/P-N-Suganthan/2020-Bound-Constrained-Opt-Benchmark}}. 

\item
\emph{Artificial mixed representation problems.}\newline
Benchmark suites combining discrete and continuous variables include mixed-integer NK landscapes \cite{LI2006}, mixed-binary and real encoded multi-objective problems \cite{MCCL2011}, mixed-integer problems based on the \gls{cec} functions \citep{LiaSocMonStuDor2014}, and a mixed-integer suite based on the \gls{bbob} functions (bbob-mixint) with a bi-objective formulation (bbob-biobj-mixint) \citep{Tusar2019}.


\item
\emph{Black-box optimization problems.}\newline 
For all benchmarks listed here, the problem formulation and the instances typically are publicly available, which inevitably leads to a specialization of algorithms to these. The Black-Box Optimization Competition\footnote{\url{https://www.ini.rub.de/PEOPLE/glasmtbl/projects/bbcomp/}} has attempted to address this shortcoming with its single- and multi-objective, continuous optimization problems. Having said this, in 2019, the evaluation code has been made available.

\item
\emph{Constrained real-parameter problems.}\newline Most real-parameter benchmark problems are unconstrained (except for basic bounds on variables) and there is a general lack of constrained benchmark sets for \gls{ec}. Exceptions include a set of 18 artificial scalable problems for the \gls{cec} 2010 Competition on Constrained Real-Parameter Optimization \footnote{\url{https://github.com/P-N-Suganthan/CEC2010-Constrained}}, six constrained real-parameter multi-objective real-world
problems presented by ~\citet{TI2020AETURWMOOPS} and a set of 57 real-world constrained problems\footnote{\url{https://github.com/P-N-Suganthan/2020-RW-Constrained-Optimisation}} for both the \gls{gecco} and \gls{cec} 2020 conferences. 

\item
\emph{Dynamic single-objective optimization problems.}\newline
Benchmark problems for analysing \glspl{ea} in dynamic environments should ideally allow for the nature of the changes (such as severity and frequency) to be configurable. A useful resource on benchmarks for dynamic environments is the comprehensive review by \citet{Nguyen2012}.   


\item
\emph{Expensive optimization problems.}\newline The \gls{gecco} 2020 Industrial Challenge provides a suite of discrete-valued electrostatic precipitator problems with expensive simulation-based evaluation\footnote{\url{https://www.th-koeln.de/informatik-und-ingenieurwissenschaften/gecco-challenge-2020_72989.php}}. An alternative approach to benchmarking expensive optimization (used by \gls{cec} competitions) is to limit the number of allowed function evaluations for solving existing benchmark problems. 

\item
\emph{Multimodal optimization (niching).}\newline Benchmark problem sets for niching include the \gls{gecco} and \gls{cec} competitions on niching methods for multimodal optimization problems\footnote{\url{http://epitropakis.co.uk/gecco2020/}} and the single-objective multi-niche competition problems\footnote{\url{https://github.com/P-N-Suganthan/CEC2015-Niching}}.

\item
\emph{Noisy.}\newline The original version of the Nevergrad platform~\cite{nevergrad}\footnote{\url{https://github.com/facebookresearch/nevergrad}} had a strong focus on noisy problems, but the platform now also covers discrete, continuous, mixed-integer problems with and without constraints, with and without noise, explicitly modelled problems and true black-box problems, etc. The electroencephalography (EEG) data optimization problem set of the \gls{cec} Optimization of Big Data 2015 Competition\footnote{\url{http://www.husseinabbass.net/BigOpt.html}} also includes noisy versions of the problem.

\item
\emph{Problems with interdependent components.}\newline While much research tackles combinatorial optimization problems in isolation, many real-world problems are combinations of several sub-problems~\cite{Bonyadi2019opportunities}. The Travelling Thief Problem~\cite{Bonyadi2013ttpcec} has been created as an academic platform to systematically study the effect of the interdependence, and the $9\,720$ instances~\cite{Polyakovskiy2014ttplib}\footnote{\url{https://cs.adelaide.edu.au/~optlog/research/combinatorial.php}} vary in four dimensions. A number of single- and multi-objective as well as static and dynamic extensions of the Travelling Thief Problem have been proposed since then~\cite{sachdeva2020dynamic}.

\item
\emph{Real-world discrete optimization.}\newline
The \gls{gecco} competition on the optimal camera placement problem (OCP) and the unicost set covering problem (USCP) include a set of discrete real-world problem instances\footnote{\url{http://www.mage.fst.uha.fr/brevilliers/gecco-2020-ocp-uscp-competition/}}. Other real-world problems include the Mazda benchmark problem\footnote{\url{http://ladse.eng.isas.jaxa.jp/benchmark/}}, which is a scalable, multi-objective, discrete-valued, constrained problem based on real-world car structure design, and a benchmark suite of combinatorial logic circuit design problems \cite{DESO2020} that cover a range of characteristics influencing the difficulty of the problem.

\item
\emph{Real-world numerical optimization.}\newline
A set of 57 single-objective real-world constrained problems were defined for competitions at a number of conferences\footnote{\url{https://github.com/P-N-Suganthan/2020-RW-Constrained-Optimisation}}. Other benchmarks include electroencephalography (EEG) data optimization problems~\cite{Goh2015}, sum-of-squares clustering benchmark problem set~\cite{GALL2016}, the Smart Grid Problems Competitions for real-world problems in the energy domain\footnote{\url{http://www.gecad.isep.ipp.pt/ERM-competitions/home/}}, and the Game Benchmark for \glspl{ea}~\cite{volz2019single} of test functions inspired by game-related problems\footnote{\url{http://www.gm.fh-koeln.de/~naujoks/gbea/gamesbench.html}}.

\end{enumerate}

\subsection{Open Issues} 
We see a number of opportunities for research on problems sets for benchmarking.

First, the number of real-world benchmark seems to be orders of magnitude smaller than the actual number of real-world optimisation problems that are tackled on a daily basis---this is especially true for continuous optimization. When there are some proper real-world problems available (e.g. data sets for combinatorial problems, or the \gls{cec} problems mentioned), they are often single-shot optimizations, i.e., only a single run can be conducted, which then makes it difficult to retrieve generalizable results. 
Having said this, a recent push towards a collection and characterization has been made with a survey\footnote{\url{https://sites.google.com/view/macoda-rwp/home}} by the \gls{macoda} working group.

Second, the availability of diverse instances and of source code (of fitness functions, problem generators, but also of algorithms) leaves much to be desired. Ideal are large collections of instances, their features, algorithms, and their performance---the \gls{aslib}\footnote{\url{https://github.com/coseal/aslib_data}}~\cite{Bischl2016aslib} has such data, although for a different purpose. As a side effect, these (ideally growing) repositories offer a means against the reinvention of the wheel and the benchmarking against so-called ``well-established'' algorithms that are cited many times---but maybe just cited many times because they can be beaten easily. 

Third, and this is more of an educational opportunity: we as the community need to make sure that we watch our claims when benchmarking.  
This includes that we not only make claims like ``my approach is better than your approach'', but that we also investigate what we can learn about the problem and about the algorithms (see e.g. the discussion in \cite{Agrawal2020duo} in the context of data mining), so that we can inform again the creation of new instances. Or to paraphrase this: we need to clarify what conclusions can we actually attempt to draw, given the performance comparison is always ``with respect to the given benchmark suite''.

Fourth, it is an advantage of test problem suites that they can provide an objective means of comparing systems. However, there are also problems related to test problem suites: \citet{Whit02b} discuss the potential disadvantage  that systems can become overfitted to work well on benchmarks and therefore that good performance on benchmarks does not generalize to real-world problems.
\citet{Fisc18a} list and discuss several drawbacks of these test suites, namely: (i) problem instances are somehow artificial and have no direct link to real-world settings; (ii) since there is a fixed number of test instances, algorithms can be fitted or tuned to this specific and very limited set of test functions; (iii) statistical tools for comparisons of several algorithms on several test problem instances are relatively complex and not easy to analyze.
    
Last, while for almost all benchmark problems and for a wide range of real-world problems the fitness of a solution is deterministic, there are also many problems out there where the fitness evaluations are conducted under noise. Hence, the adequate handling of noise can be critical so as to allow algorithms to explore and exploit the search space in a robust manner. \citet{Bra01} discuss  strategies for coping with noise, and  \citet{Jin05} present a good survey.  While noise (in computational experiments) is often drawn from relatively simple distributions, real-world noise can be non-normal, time-varying, and even be dependent on system states. 
To validate experimental outcomes from such noisy environments, mechanisms way beyond ``do $n$ many repetitions'' are needed, and \citet{Bokhari2020validationNoise} compare five such approaches.

\section{Algorithms}
\label{sec:sec4}

To understand strengths and weaknesses of different algorithmic ideas, it is important to select a suitable set of algorithms that is to be tested within the benchmark study. While the algorithm portfolio is certainly one of the most subjective choices in a benchmarking study, there are nevertheless a few design principles to respect. In this section we summarize the most relevant of these guidelines.  

\subsection{Algorithm Families}
To assess the quality of different algorithmic ideas, it is useful to compare algorithm instances from different families. For example, one may want to add solvers from the families of 
\begin{itemize}
    \item one-shot optimization algorithms (e.g., pure random search, \gls{lhd}~\cite{LHS}, or quasi-random point constructions),
    \item greedy local search algorithms (e.g., randomized local search,  \gls{bfgs} algorithm~\citep{shanno1970}, conjugate gradients~\citep{fletcher1976}, and Nelder-Mead~\citep{Neld65a})
     \item non-greedy local search algorithms (e.g., \gls{sann}~\cite{Kirkpatrick83}, Threshold Accepting~\cite{DueckThreshold}, and \emph{Tabu Search}~\citep{glov89a})
    \item single-point global search algorithms (e.g., $(1+\lambda)$ Evolution Strategies~\cite{eibe03a} and Variable Neighborhood Search~\cite{VariableNeighborhoodSearch})
    \item population-based algorithms (e.g., \gls{PSO}~\citep{kenn95a,shi98a}, ant colony optimization~\citep{DorBirStu06:ci,soch08a}, most \glspl{ea}~\cite{back97a,eibe03a}, and \glspl{edalgo}~\cite{EDAMuehlenbein,Larranaga2002EDA} such as the \gls{cmaes}~\citep{hans03a})
    \item surrogate-based algorithms (e.g., \gls{ego} algorithm~\citep{Jones1998} and other Bayesian optimization algorithms)
\end{itemize}
Note that the ``classification'' above is by no means exhaustive, nor stringent. In fact, classification schemes for optimization heuristics always tend to be fuzzy, as hybridization between one or more algorithmic ideas or components is not unusual, rendering the attribution of algorithms to the different categories subjective; see~\citep{BoussaidLS13ClassificationSBOH,BirattariPS03ClassificationSBOH,stork18a} for examples.

\subsection{Challenges and Guidelines for the Practitioner}\label{sec:4challenges}

The following list summarizes considerations that should guide the selection of the algorithm portfolio. This list is not meant to recommend certain algorithms to solve a given problem class, but to give an overview of aspects that should be taken into consideration before a benchmark study starts. 
\begin{enumerate} [({C4}.1)]
    
    \item \emph{Problem features}\newline
    The arguably most decisive criterion for the algorithm portfolio is the type of problems that are to be benchmarked. Where information such as gradients are available, gradient-based search methods should be included in the benchmark study. Where mixed-integer decision spaces are to be explored, different algorithms are relevant than for purely numerical or purely combinatorial problems. Also, other characteristics such as the degree of variable interaction, the (supposed) shape of the objective value landscape etc. should determine the algorithm portfolio. 
    
    We recommend gathering and using all available information about the problem, e.g., its landscape features~\cite{kerschke_automated_2019} and algorithms performances on the problem class from the past~\cite{kerschke_automated_2019}. 
    Even if the goal of the benchmark study does not have a competitive character, a deeper look into results of preceding competitions or workshops can give useful hints which algorithms to select. Benchmark data repositories such as those collected in~\cite{IOHanalyzer} are designed to support the user in these tasks. 

    \item \emph{Budget and convergence}\newline
     Compute power and the availability of resources to interpret the benchmarking results has a strong impact on the \emph{number of algorithms} that can be compared, whereas the \emph{budget} that can be allocated to each algorithm is typically driven by the research question or application of interest. The budget can have a decisive influence on the selection of algorithms. For example, surrogate-assisted algorithms tend to be algorithms of choice for small budgets, whereas evolution strategies tend to be more competitive for mid- and large-sized budgets.       

    \item \emph{State of the art}\newline
     Results of a benchmark study can easily be biased if only outdated algorithms are added.  We clearly recommend familiarizing oneself with the state-of-the-art algorithms for the given problem type, where state of the art may relate to performance on a given problem class or the algorithm family itself. 
    Preliminary experiments may lead to a proper pre-selection (i.e., exclusion) of algorithms. 
    The practitioner should be certain to compare versus the best algorithms. 
    Consequently, always compare to the most current versions and implementations of algorithms. 
    This also counts for the programming platform and its versions. 
    For algorithm implementations on programming platforms and operating systems the practitioner is not familiar with, nowadays there exist methods and technologies to solve this inconvenience, e.g., container (like Docker) or virtualizations.
    For details about considerations regarding the experimental design, e.g., the number of considered algorithms, the number of repetitions, the number of problem instances, the number of different parameter settings, or sequential designs if the state of the art is unknown, please see Section~\ref{sec:07design}. 
    
    \item \emph{Hyperparameter handling}\newline
    All discussed families of algorithms require one or several control parameters.
    To enable a fair comparison of their performances and to judge their efficiency, it is crucial to avoid bad parameter configurations and to properly tune the algorithms under consideration~\cite{eibe12a, Beir17a}.
    Even a well-working parameter configuration for a certain setup, i.e., a fixed budget, may work comparably worse on a significantly different budget.
    As mentioned in Section~\ref{sec:sec2} under goal (G2.\ref{goal:tuning}), the robustness of algorithms with respect to their hyperparameters can be an important characteristic for users, in which case this question should be integrated into (or even be the subject of) the benchmarking study. 
    Furthermore, the practitioner should be certain that the algorithm implementation is properly using the parameter setup. 
    It may occur that some implementations do not warn the user if the parameter setting is out of their bounds. 
       
    Several tools developed for automatic parameter configuration are available, e.g., \gls{irace}~\cite{LopDubPerStuBir2016irace}, \gls{paramils}~\cite{HutHooLeyStu2009jair}, \gls{spot}~\cite{bartz05s}, \gls{smac}~\cite{hutt11}, GGA~\cite{GGA}, and hyperband~\cite{hyperband} to name a few.
    As manual tuning can be biased, especially for algorithms unknown to the experimenter, automated tuning is state of the art and highly recommended.
    Giving rise to a large amount of research in the field of automated algorithm configuration and hyperparameter optimization, there exist several related benchmarking platforms, like the algorithm configuration library (ACLib)~\citep{HutLopFaw2014lion} or the hyperparameter optimization library (HPOlib)~\cite{eggensperger2013}, which deal particularly with this topic.
    
    \item \emph{Initialization} \newline
    A good benchmarking study should ensure that the results achieved do not happen by chance. 
    It can be important to consider that algorithm performances are not erroneously rated due to the (e.g., random) initialization of the algorithm. 
    According to a given problem instance, a random seed-based starting point can be beneficial for algorithms if they are placed near or at one or more local optima.
    Consequently, the practitioner should be aware that the performance of algorithms can be biased by the initialization of algorithms with respect to, e.g., their random seeds, the starting points, the sampling strategy, combined with the difficulty of the chosen problem instance. 
    
    We recommend letting all candidate algorithms use the same starting points, especially when the goal of the benchmarking study is to compare (goals G1.\ref{goal:algorithm} and G1.\ref{goal:competition}) or to analyze the algorithms search behavior (G1.\ref{goal:basic}). 
    This recommendation also extents to the comparison with historical data. 
    Additionally, the design of experiment (see Section~\ref{sec:07design}) can reflect the considerations by properly handling the number of problem instances, repetitions, sampling strategies (in terms of the algorithm parametrization), and random seeds.
    For (random) seed handling and further reproducibility handling, we refer to Section~\ref{sec:09reproducibility}.
    
    \item \emph{Performance assessment} \newline
    Not all algorithms support the configuration of the same stopping criteria, which may influence the search~\cite{Beir17a} and which has to be taken into account in the interpretation of the results. For example, implementation of algorithms may not respect the given number of objective function evaluations. 
    If not detected by the practitioner, this can largely bias the evaluation of the benchmark. 
\end{enumerate}

\subsection{Challenges and Open Issues}\label{sec:challenges}

The selection of the algorithms to be included in a benchmarking study depends to a great extent on the users experience, the availability of off-the-shelf or easy-to-adjust implementations, the availability of data about the problems and/or algorithms, etc. Identifying relevant algorithms, data sets, research papers often requires a major effort. Even where data and implementations are easily available, formats can greatly differ between different studies, hindering their efficient use. We therefore believe that common data formats, common benchmark interfaces, and a better compatibility between existing software to assist benchmarking of optimization heuristics is greatly needed. 

Another major issue in the current benchmarking landscape concerns a lack of detail in the description of the algorithms. Especially for complex, say, surrogate-assisted optimization heuristics, not all parameters and components are explicitly mentioned in the paper. Where code is available in an open access mode, the user can find these details there, but availability of algorithms implementations is still a major bottleneck in our community.

\section{How to Measure Performance?}
\label{sec:sec5}
\label{sec:performance}

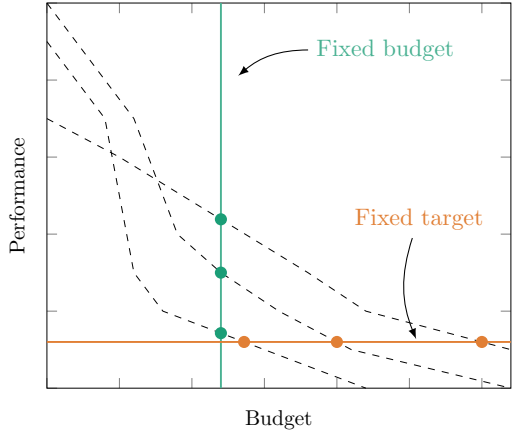
\begin{figure}
    \centering
    \begin{tikzpicture}[scale=0.9, every node/.style={scale=0.9}]
\definecolor{brewergreen}{HTML}{1B9E77}
\definecolor{brewerorange}{HTML}{D95F02}
  \begin{axis}[
    xmin=0, xmax=32, ymin=0, ymax=10,
    xlabel=Budget,
    ylabel=Performance,
    yticklabels=\empty,
    xticklabels=\empty]
    \addplot[black,dashed,name path=run1] coordinates {(0, 10) (4,8) (6,7) (7,6) (9,4) (12,3) (16,2) (21,1) (32,0)};
    
    \addplot[black,dashed,name path=run2] coordinates {(0, 9) (4,7) (6,3) (8,2) (22,0)};
    
    \addplot[black,dashed,name path=run3] coordinates {(0, 7) (5,6) (18,3) (22,2) (32,1)};
    
    \addplot[brewergreen!80,solid,thick,name path=vertcut] coordinates {(12, 10) (12,0)};
    
    \addplot[brewerorange!80,solid,thick,name path=horcut] coordinates {(0, 1.2) (32,1.2)};
    
    \path [name intersections={of=run1 and vertcut}]; 
    \node[fill=brewergreen,circle,inner sep=2pt] (OP1) at (intersection-1) {};
    
    \path [name intersections={of=run1 and horcut}]; 
    \node[fill=brewerorange!80,circle,inner sep=2pt] (OP2) at (intersection-1) {};
    
    \path [name intersections={of=run2 and vertcut}]; 
    \node[fill=brewergreen,circle,inner sep=2pt] (OP1) at (intersection-1) {};
    
    \path [name intersections={of=run2 and horcut}]; 
    \node[fill=brewerorange!80,circle,inner sep=2pt] (OP2) at (intersection-1) {};
    
    \path [name intersections={of=run3 and vertcut}]; 
    \node[fill=brewergreen,circle,inner sep=2pt] (OP1) at (intersection-1) {};
    
    \path [name intersections={of=run3 and horcut}]; 
    \node[fill=brewerorange!80,circle,inner sep=2pt] (OP2) at (intersection-1) {};
    
  \end{axis}
  \node (vertcutdummy) at (2.7,4.5) {};
  \node (vertcutlabel) at (5,5) {\textcolor{brewergreen!80}{Fixed budget}};
  
  \draw (vertcutlabel.west) edge[->,bend right=20,>=latex] (vertcutdummy);
  
  \node (horcutdummy) at (5.5,0.6) {};
  \node (horcutlabel) at (5.5,2.5) {\textcolor{brewerorange!80}{Fixed target}};
  
  \draw (horcutlabel) edge[->,bend right=20,>=latex] (horcutdummy);
\end{tikzpicture}
\caption{Visualization of a fixed-budget perspective (vertical, green line) and a fixed-target perspective (horizontal, orange line) inspired by Figure~4 in~\cite{HAFR2012RPBBOBES}. Dashed lines show three exemplary performance trajectories.}
\label{fig:performance_cuts}
\end{figure}

The performance of algorithms can be measured with regard to several objectives, of which \emph{solution quality} and \emph{consumed budget} are the most obvious two (see Figure~\ref{fig:performance_cuts}). In fact, when benchmarking algorithms one usually examines them with regard to one of the following two questions:
\begin{compactitem}
\item ``How fast can the algorithms achieve a given solution quality?'' (Section~\ref{sec:performance:budget}) or
\item ``What solution quality can the algorithms achieve with a given budget?'' (Section~\ref{sec:performance:quality})
\end{compactitem}
These two scenarios correspond to vertical and horizontal cuts in a performance diagram as discussed by~\citet{HAFR2012RPBBOBES} and \citet{FHRA2015CDR1}, respectively (see Figure~\ref{fig:performance_cuts}). The fixed-budget scenario (vertical cut) comes with the benefit that its results are well-defined as any real computation has a limited budget.
Whereas fixing the desired solution quality (horizontal cut) allows to draw conclusions that are easier to interpret; statements such as ``algorithm instance~$b$ is ten times faster than algorithm instance~$a$ in solving this problem'' are likely more tangible compared to ``the solution quality achieved by algorithm instance~$b$ is 0.2\% better than the one of algorithm instance~$a$.'' However, as not all algorithm runs may hit the chosen target, users of \emph{fixed-target measures} need to define how they treat such non-successful runs.

Depending on the chosen time budgets or targeted objective values, different algorithms may yield better results or shorter run times, respectively. Therefore, instead of measuring the objectives using a fixed value, algorithms can also be assessed regarding their \emph{anytime behaviour}~\citep{jesus2020,BossekKT2020AnytimeBehavior}. In those cases, the performance does not correspond to a singular point, but instead to an entire curve in the time-quality diagrams. Note that all three views have different implications, and each of them has its justification. As a result, it depends on the application at hand, which perspective should be focussed.

In addition, the \emph{robustness of the found solution} -- which might be affected by the algorithm's stochasticity, a noisy optimization problem, or the smoothness of the landscape in a solution's vicinity -- can also be in a study's focus. However, as outlined in Section~\ref{sec:performance:robustness}, measuring this objective can be very challenging.

\subsection{Measuring Time}%
\label{sec:performance:budget}%

It should be noted that \emph{time} can be measured in different ways with clock or CPU time being the most intuitive.
In several combinatorial optimization problems like solving \gls{tsp}~\cite{KerschkeKBHT18} or \gls{sat} problems~\cite{XuHutHooLey2008jair}. CPU time is the default. However, as it is highly sensitive to a variety of external factors -- such as hardware, programming language, work load of the processors -- results from experiments that relied on CPU time are much less replicable and thus hardly comparable. In an attempt of mitigating this issue, \citet{JMG2004EAOHFTS} proposed a normalized time, which is computed by dividing the runtime of an algorithm by the time a standardized implementation of a standardized algorithm requires for the same (or at least a comparable) problem instance. 

An alternative way of measuring time are \glspl{fe}, i.e., the number of fully evaluated candidate solutions. In fact, in case of sampling-based optimization, like in classical continuous optimization, this machine-independent metric is \emph{the} most common way of measuring algorithmic performances~\cite{HansenABTT16}. Yet, from the perspective of actual clock time, they risk giving a wrong impression as the \glspl{fe} of different algorithms might be of different time complexity~\citep{WCTLTCMY2014BOAAOSFFTTSP}.
In such cases, counting algorithm steps in a domain-specific method -- e.g., the number of distance evaluations on the \gls{tsp} \cite{WCTLTCMY2014BOAAOSFFTTSP} or bit flips on the \gls{maxsat} problem \cite{HWHC2013HILSFM} -- may be useful.
Nevertheless, within the \gls{ec} community, counting \glspl{fe} is clearly the most commonly accepted way to  measure the efforts spend by an algorithm to solve a given problem instance.

From a practical point of view, both options have their merits.
If the budget is given by means of clock time -- e.g., models have to be trained until the next morning, or they need to be adjusted within seconds in case of predictions at the stock market -- then results relying on CPU time are more meaningful.
On the other hand, in case single \glspl{fe} are expensive -- e.g., in case of physical experiments or cost-intensive numerical simulations -- the number of required \glspl{fe} is a good proxy for clock time, and a more universal measure, as discussed above. 
To satisfy all perspectives, best practice would be to report both: \glspl{fe} \emph{and} the CPU time. Moreover, in situations in which single \glspl{fe} are expensive, CPU time should ideally be separated into the part used by the expensive \gls{fe}, and the part used by the algorithm at each iteration.

In surrogate-based optimization, algorithms commonly slow down over time due to an ever-increasing complexity of their surrogate models~\cite{ueno2016combo,bliek2020black}.
In this case, it is even possible that the algorithm becomes more expensive than the expensive \gls{fe} itself.
By measuring the CPU time used by the algorithm separately from the CPU time used by the \glspl{fe}, it can be verified that the number of \glspl{fe} is indeed the limiting factor.
At the same time, this reveals more information about the benchmark, namely how expensive it is exactly, and whether all \glspl{fe} have the same cost.
However, if the \glspl{fe} are expensive not because of the computation time but due to some other costs (deteriorating biological samples, the use of expensive equipment, human interaction, etc.), then just measuring the number of \glspl{fe} could be sufficient.

Noticeably, many papers in the \gls{ec} community also use \emph{generations} as machine-independent time measures.
However, it might not be a good idea to \emph{only} report generations, because the exact relationship between \glspl{fe} and generations is not always clear.
This makes results hard to compare with, e.g., local search algorithms, so if generations are reported, \glspl{fe} should be reported as well.


\subsection{Measuring Solution Quality}%
\label{sec:performance:quality}%

There are several ``natural'' quality metrics, e.g., fitness in continuous optimization, the tour length in \gls{tsp}, the accuracy of a classification algorithm in a machine learning task, or the number of ones in a binary bit string in case of OneMax. However, interpreting these objective values on their own is usually quite difficult and also very specific to the respective problem instance. Instead, one could ideally try to use more intuitive and less problem-dependent alternatives as described below~\citep{J2002ATGTTEAOA,T2009MFDTI}.

If an instance's optimal solution is known, the (absolute or relative) difference to the optimal target quality could be used.
Alternatively, a best-known lower bound for the optimal objective value could be used for normalizing the results.
For instance, in case of the \gls{tsp}, results are often compared to the Held-Karp lower bound~\cite{J2002ATGTTEAOA}. As \emph{absolute} differences are very specific to the scaling of the problem's objective values, it is highly recommended to rather look at the \emph{relative} excess over the optimal solution -- something that has been common practice in solving \glspl{tsp} for decades~\cite{christofides1976}. Noticeably, despite the varying scales of the objective values across the different problem instances, absolute differences are in the focus of continuous optimization benchmarks like BBOB \cite{cocoplat}.

Another alternative is to use results of a clearly specified and ideally simple heuristic for normalization~\citep{J2002ATGTTEAOA}.
The relative excess over the best-known solution is also often reported.
This is typical in the Job Shop Scheduling domain, as shown by many references listed in~\citet{W2019JRDAIOTJSSP}.
However, this requires an exact knowledge of the related work and may be harder to interpret later in the future.
For some problems, reference solutions may be available and the excess over their quality can be reported.

\paragraph{Constraint optimization} Under constraint optimization, a solution is either feasible or not, which is decided based on a set of constraints.
Here, the absolute violations of each constraint can be summed up as a performance metric~\cite{HB2018BEAFSORVCOACR,KWZMSD2020GFRWSOCOC}.%

\subsection{Measuring Robustness}%
\label{sec:performance:robustness}%

In terms of robustness analyses, one can differentiate between three reasons for volatility among the results: (i) stochastic search behavior of the considered algorithm (e.g., in randomized search heuristics), (ii) noisy problems, (iii) ruggedness or smoothness of the problem landscape.

From a practical point of view, rugged landscapes can be highly problematic for the outcome of an optimization problem. For instance, when controlling parameters of an airplane or conducting medical surgeries, the global optimum will likely not be targeted, if slight variations (in search space) can have hazardous effects on the objective space and thus the on the whole system \cite{tsutsui1996robust, branke1998creating}. In such scenarios, one likely is much more interested in finding (local) optima, whose objective values are still very close to the global optimum, but change only slightly when perturbing the underlying solutions.

Another common issue in real-world applications is the existence of noise. In particular, in case of physical experiments or stochastic simulations models, the outcome of an experiment may vary despite using the same candidate solution \cite{arnold2012noisy,cauwet2016noisy}.

The third issue related to robustness investigations is the stochastic nature of the algorithms themselves. In fact, many sampling-based optimization algorithms are nowadays randomized search heuristics and as such their performances will vary if the experiment is repeated, i.e., if the algorithm is executed again using the same input. Therefore, it is common to use performance metrics that aggregate the results of several (ideally independent) runs to provide reliable estimates of the algorithm performance.

\paragraph{Location: Measures of Central Behaviors}%
In case of a fixed budget (i.e., vertical cut) approach, solution qualities are usually aggregated using the arithmetic mean.
But of course, other location measures like the median 
or the geometric mean \cite{Flem86a} can be useful alternatives when interested in robust metrics or when aggregating normalized benchmark results, respectively.

In scenarios, in which the primary goal is to achieve a desired target quality (horizontal cut), it might be necessary to aggregate successful and failed runs. In this case, two to three metrics are mostly used for aggregating performances across algorithm runs, and we will list them below. 

The gold standard in (single-objective) continuous optimization is the \gls{ert}~\cite{P1997DEVTFOT2I,AH2005PEOAALSEA}, which computes the ratio between the sum of consumed budget across all runs and the number of successful runs~\cite{HAFR2012RPBBOBES}. Thereby, it estimates the average running time an algorithm needs to find a solution of the desired target quality (under the assumption of independent restarts every~$T$ time units until success).

In other optimization domains, like \gls{tsp}, SAT, etc., the \emph{Penalized Average Runtime (PAR)}~\cite{Bischl2016aslib} is more common. It penalizes unsuccessful runs with a multiple of the maximum allowed budget -- penalty factors ten (PAR10) and two (PAR2) are the most common versions -- and afterwards computes the arithmetic mean of the consumed budget across all runs. The \emph{Penalized Quantile Runtime (PQR)}~\cite{BKTAMOPOPAAASOOA,KBT2018POSOTAPIARSBOITS} works similarly, but instead of using the arithmetic mean for aggregating across the runs, it utilizes quantiles---usually the median -- of the (potentially penalized) running times. In consequence, PQR provides a robust alternative to the respective PAR scores.

\paragraph{Spread and Reliability}%
A common measure of reliability is the \emph{estimated success probability}, i.e., the fraction of runs that achieved a defined goal.
\citet{BKTAMOPOPAAASOOA} use it to take a multi-objective view by combining the probability of success and the average runtime of successful runs. Similarly, \citet{HB2018BEAFSORVCOACR} aggregate the two metrics in a single ratio, which they called \emph{SP}.

As measures of dispersion of a given single performance metric, statistics like standard deviations as well as quantiles are used, whereas the latter are more robust.

For constraint optimization, a \emph{feasibility rate} (FR)~\cite{KWZMSD2020GFRWSOCOC,WMS2017PDAECFTCCOCRPO} is defined as the fraction of runs discovering at least one feasible solution.
The number of constraints violated by the median solution~\cite{KWZMSD2020GFRWSOCOC} and the mean amount of constraint violation over the best results of all runs~\cite{HB2018BEAFSORVCOACR} can also be used.

\subsection{Open Issues}%

Although each of the different optimization domains has established its preferable performance metric, research in this field is still facing open issues.
For instance, so far performance is mostly measured using fixed values (budget or target). However, depending on the use case, comparing the anytime behavior of algorithms might occasionally be of interest as well.

Aside from facing challenges like measuring quality \emph{and} time simultaneously, we also have to integrate costs for violating constraints (constraint optimization), quantify variation or uncertainty (robust/noisy optimization), measure the spread across the local optima (multimodal optimization), or capture the proximity of the population to the local and/or global optima of the problem.

\section{How to Analyze Results?}
\label{sec:sec6}

\subsection{Three-Level Approach}
Once the performance measure for the algorithm's performance is selected by the user and all data related to it is collected in experiments, the next step is to analyse the data and draw conclusions from it.
From the detailed characterization of possible benchmark goals in Section~\ref{sec:sec2}, we will focus on goals (G1.\ref{goal:algorithm}) and (G1.\ref{goal:competition}),  
i.e., algorithm comparison and competition of several algorithms. Therefore, we will consider:
\begin{itemize}
    \item single-problem analysis and
    \item multiple-problem analysis.
\end{itemize}
In both scenarios, multiple algorithms will be considered, i.e., following the notation introduced in Section~\ref{sec:sec2}, 
there are at least two different algorithm instances, say, $a_j$ and $a_k$ from algorithm $A$ or at least two 
different algorithm instances $a_j \in A$ and $b_k \in B$, where $A$ and $B$ denote the corresponding algorithms.
Single-problem analysis is a scenario where the data consists of multiple runs of the algorithms on a single problem instance $\pi_i \in \Pi$. 
This is necessary because many optimization algorithms are stochastic in nature, 
so there is no guarantee that the result will be the same for every run. 
Additionally, the path leading to the final solution is often different. 
For this reason, it is not enough to perform just a single algorithm run per problem, 
but many runs are needed to make a conclusion.
In this scenario, the result from the analysis will give us a conclusion which algorithm performs the best on that specific problem.

Otherwise, in the case of multiple-problem analysis, focusing on (G1.\ref{goal:algorithm}), 
we are interested in comparing the algorithms on a set of benchmark problems.
The best practices of how to select a representative value for multiple-problem analysis will be described 
in Section~\ref{sec:07design}. 

No matter of what we are performing, i.e., single-problem or multiple-problem analysis, 
the best practices analyzing the results of the experiments suggest that the analysis can be done as a three-level approach, which consists of the following three steps:
\begin{enumerate}
    \item \gls{eda}
    \item Confirmatory Analysis
    \item Relevance Analysis
\end{enumerate}

This section focuses on analyzing the empirical results of an experiment using descriptive, graphical, 
and statistical tools, which can be used for the three-level approach for analysis. 
More information about various techniques and best practices analyzing  the results of experiments can be found in \citet{Crow79a}, \citet{Gold86a}, \citet{Barr95a}, \citet{Bart04a},  \citet{Chia07a}, \citet{Garc09a}, \citet{Bart10a}, \citet{Derr11a}, \citet{eftimov2017novel}, \citet{Beir17a}. 
\citet{Mers10a}, and more recently \citet{KerschkeT19}, present methods based on \gls{ela} to answer two basic questions that arise when benchmarking optimization algorithms. The first one is: which algorithm is the ‘best’ one? and the second one: which algorithm should I use for my real world problem? 
In the following, we summarize the most accepted and standard practices to evaluate the considered algorithms stringently. These methods, if adhered to, may lead to wide acceptance and applicability of empirically tested algorithms and may be a useful guide in the jungle of statistical tools and methods.

\subsection{Exploratory Data Analysis}

\subsubsection{Motivation}

Exploratory Data Analysis (\Gls{eda}) is an elementary tool that employs descriptive and graphical techniques to better understand and explore empirical results. It must be performed to validate the underlying assumptions about the distribution of the results, e.g., normality or independence, before implementing any statistical technique that will be discussed in Section~\ref{sec:sa}. 

We recommend starting with \gls{eda} to understand basic patterns in the data. It is useful to prepare (statistical) hypotheses, which are the basis of confirmatory analysis.
In \gls{eda}, visual tools are preferred, whereas confirmatory analysis is based on probabilistic models. \gls{eda} provides a flexible way to analyze data without preconceptions.
Its tools stem from descriptive statistics and use an inductive approach, because in the beginning, there is no theory that has to be validated.
One common saying is "let the data speak", so data suggest interesting questions, e.g., unexpected outliers might indicate a severe bug in the algorithm.
\gls{eda} is a very flexible way to generate hypotheses, which can be analyzed in the second step (confirmatory analysis).
Although \gls{eda} might provide deeper understanding of the algorithms, it does not always provide definitive answers. Then, the next step (confirmatory analysis) is necessary. And, there is also the danger of overfitting: focussing on very specific experimental designs and results might cause a far too pessimistic (or optimistic) bias. Finally, it is based on experience, judgement, and artistry. So there is no standard cookbook available, but many recipes.

The following are the key tools available in \gls{eda}. It can provide valid conclusions that are graphically presented, without requiring further statistical analysis.
For further reading about \gls{eda}, the reader is referred to~\cite{Tuke77a}. 

\subsubsection{The Glorious Seven}\label{sec:gloriousseven}
Descriptive statistics include the mean, median, best and worst (minimum and maximum, respectively), first and third quartile, and standard deviation of the performance measures of the algorithms. 
These seven so-called summary statistics measure the central tendency and the variability of the results. Note, they might be  sensitive to outliers, missing, or biased data. Most importantly, they do not provide a complete analysis of the performance, because they are based on a very specific data sample.
For example, mean and standard deviation are affected by outliers, which might exist because of an algorithms' poor runs and variability.
Both can be caused by an inadequate experimental design, e.g., selection of improper starting points for the algorithm or too few function evaluations.
The median is more robust statistic than the mean if sufficiently many data points are available. The best and the worst value of the performance measure provide insights about the performance data, but they consider only one out of $n$ data points, i.e., they are determined by one single data and therefore not very robust compared to the mean or median that are considering all data points.
The quantiles are cut points which split a probability distribution into continuous intervals with equal probabilities. Similar to the median, they require a certain amount of data points and are probably meaningless for small data.
\citet{Bart06a} presents a detailed discussion of these basic statistics.

\subsubsection{Graphical Tools} 
\paragraph{Visualising final results.}
Graphical tools can provide more insight into the results and their distributions. 
The first set of graphical tools can be used to analyse the final results of the optimization runs.
\emph{Histograms} and \emph{boxplots} are simple but effective tools and provide more information for further analysis of the results. 
Box plots visualize the distribution of the results. They illustrate the  statistics introduced in Section~\ref{sec:gloriousseven} in a very compact and comprehensive manner and provide means to detect outliers.
Histograms provide information about the shape of the distribution. Because the shape of histograms is highly affected by the size of the boxes, we strongly recommend combining histograms with density plots.

\paragraph{Visualising run-time behaviour.}
The second set of tools can be used to analyse the algorithm performance over time, i.e., information about the performance for every $l$th iteration is required. 
Suitable for the analysis of the performances of the optimization algorithms are \emph{convergence plots} in which the performance of the algorithm can be evaluated against the number of function evaluations. This helps us to understand the dynamics of multiple algorithms in a single plot.

Histograms and box plots are also used in the graphical multiple problem analysis. Besides these common tools, specific tools for the multiple problem analysis were developed, e.g., \emph{performance profiles} proposed in~\citet{Dola02a}. They have emerged as an important tool to compare the performances of optimization algorithms based on the  cumulative distribution function of a performance metric (CPU time, achieved optimum). It is the ratio of a performance metric obtained by each algorithm versus the best value the performance metric among all algorithms that is being compared. Such plots help to visualize the advantages (or disadvantages) of each competing algorithm graphically. Performance profiles are not applicable, if the (true or theoretical) optimum is unknown. However, there are solutions for this problem, e.g., using the best known solution so far or a guessed (most likely) optimum based on the user's experience; however, the latter is likely to be error-prone.
  
As performance profiles are not evaluated against the number of function evaluations, they cannot be used to infer the percentage of the test problems that can be solved with some specific number of function evaluations. To attain this feature, the \emph{data profiles} were designed for fixed-budget derivative free optimization algorithms~\cite{More09a}. It is appropriate to compare the best possible solutions obtained from various algorithms within a fixed budget.

\subsection{Confirmatory Analysis}\label{sec:sa}
\subsubsection{Motivation}
The second step in the three-level approach is referred to as \emph{confirmatory analysis}, which is based in inferential statistics, because it implements a deductive approach: a given assumption (statistical hypothesis) is tested using the experimental data.
Since the assumptions are formulated as statistical hypotheses, confirmatory analysis heavily relies on probability models. Its final goal is to provide  definite answers to specific questions, i.e., questions for a specific experimental design. Because it uses probability models, its emphasis is on complex numerical calculations. Its main ingredients are hypothesis tests and confidence intervals. Confirmatory analysis usually generates more precise results for a specific context than \gls{eda}. But, if the context is not suitable, e.g., statistical assumptions are not fulfilled, a misleading impression of precision might occur. 

Often, \gls{eda} tools are not sufficient to clearly analyze the differences in the performances of algorithms, mainly when the differences are of smaller magnitude. 
The need  to perform statistical analysis and various procedures involved in making decisions about selecting the best algorithm are widely discussed in \cite{Gold86a,Amin93a,Barr95a,Mcge96a, Chia07a, Garc09a, Carr11a, eftimov2017novel}. The basic idea of statistical analysis is based on hypothesis testing. 
Before analysing the performance data, we should define two hypotheses i) the null hypothesis $H_{0}$ and ii) the alternative hypothesis $H_{1}$. 
The null hypothesis states that there is no significant statistical difference between the two algorithms' performances, while the alternative hypothesis directly contradicts the null hypothesis by indicating the statistical significance between the algorithms' performances. Hypothesis testing can be two-sided  or one-sided.
We will consider the one-sided case in the following, because it allows us to ask if algorithm instance $a$ is better than algorithm instance $b$. Let $p(a)$ denote the performance of algorithm $a$. In the context of minimization, smaller performance values will be better, because we will compare the best solutions or the run times.
The statement "$a$ outperforms $b$" is equivalent to "$p(a) < p(b)$", which can be formulated as the statistical hypothesis $H_1:\/ p(b) - p(a) > 0 $. It is a common agreement in hypotheses testing that this hypothesis $H_1$ will be tested against the null hypothesis $H_0:\/ p(b) - p(a) \leq 0$, which states that $a$ is not better than $b$.

After the hypotheses are defined, we should select an appropriate statistical test, say $T$, for the analysis. The test statistic $T$ which is a function of a random sample that allows researchers to determine the likelihood of obtaining the outcomes if the null hypothesis is true.
The mean of the best found values from $n$ repeated runs of an algorithm is a typical example of a test statistic.
Additionally, a significance level $\alpha$ should be selected. Usually, a significance level of 95\% is used. However, the selection of this value depends on the experimental design and the scientific question to be answered. 
 
\subsubsection{Assumptions for the Safe Use of the Parametric Tests} 
There are \emph{parametric} and \emph{non-parametric} statistical tests.
To select between them, there are assumptions for the safe use of the parametric tests. Common assumptions include independence, normality, and homoscedasticity of variances. 
The independence assumption is directly met as the results of independent runs of the algorithm with randomly generated initial seeds are being compared. 
To check the normality assumption several tests can be performed including \emph{Kolmogorov-Smirnov test} \cite{Shes03a}, \emph{Shapiro-Wilk test} \cite{Shap65a}, and \emph{Anderson Darling test} \cite{Ande52a}. 
The normality assumption can be also checked by using graphical representation of the data using \emph{histograms}, \emph{empirical distribution functions} and \emph{quantile-quantile} plots (Q-Q plots) \cite{devore2011probability}. 
The  \emph{Levene’s test} \cite{Leve61a} and \emph{Bartlett’s test} \cite{Bart37a} can be performed to check if the assumption of equality of variances is violated. We should also mention that there are transformation approaches that may help to attain the normality, but this should be done with a great care, since we are changing the decision space. If the required assumptions are satisfied then we are selecting a parametric test since it has higher power than a non-parametric one, otherwise we should select a non-parametric one.
 
 Additionally to the assumptions for the safe use of the parametric tests, before selecting an appropriate statistical test, we should take care if the performance data is paired or unpaired. Paired data is data in which natural or matched couplings occur. This means that each data value in one sample is uniquely paired to a data value in the other sample.  The choice between paired and unpaired samples depends on experimental
design, and researchers need to be aware of this when designing their experiment. Using \gls{crn} is a well-known technique for generating paired samples. If the same seeds are used during the optimization, \glspl{crn} might reduce the variances and lead to more reliable statistical conclusions~\cite{Nazz11a, Klei88a}.

\subsubsection{A Pipeline for Selecting an Appropriate Statistical Test}
\begin{figure}
    \centering
    \includegraphics[scale=0.5]{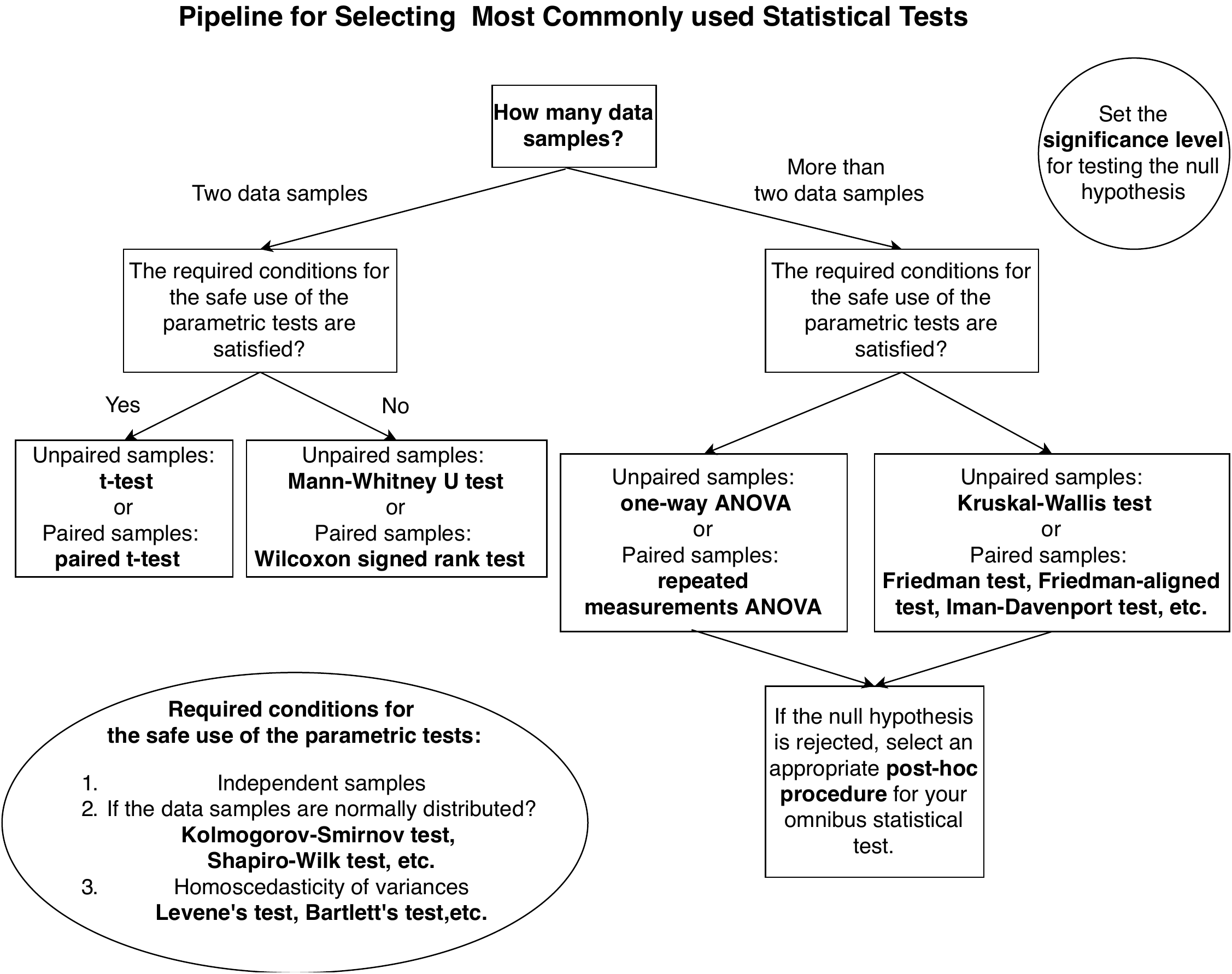}
    \caption{A pipeline for selecting an appropriate statistical test \cite{eftimov2020dsctool}.}
    \label{fig:StatisicalTests}
\end{figure}

A pipeline for selecting an appropriate statistical test for benchmarking optimization algorithms is presented in Figure \ref{fig:StatisicalTests}. Further, we are going to explain some of them depending upon the benchmarking scenario (i.e., single-problem or multiple-problem analysis). 

\paragraph{Single-problem analysis.} As we previously mentioned, in this case, the performance measure data is obtained using multiple runs of $k$ algorithm instances $a_1, \ldots, a_k$ on one selected problem instance $\pi_j$. 

The comparison of samples in pairs is called a \emph{pairwise comparison}. 
Note, that a pairwise comparison of algorithms 
does not necessarily mean that the corresponding samples are paired. In fact, most pairwise comparisons use unpaired samples, because the setup for pairwise sampling is demanding, e.g., implementing random number streams etc.
If more than two samples are compared at the same time, a \emph{multiple comparison} is performed.

For pairwise comparison, the $t$ test \cite{Shes03a} is the appropriate parametric one, while its non-parametric version is the \emph{Mann-Whitney U test} (i.e., \emph{Wilcoxon-rank sum test}) \cite{hart2001mann}. In the case when more than two algorithms are involved, the parametric version is the one-way\emph{ANOVA} \cite{Lind74a, Mont17a}, while its appropriate non-parametric test is the \emph{Kruskal-Wallis rank sum test} \cite{Krus52a}. Here, if the null hypothesis is rejected, then we should continue with a post-hoc procedure to define the pairs of algorithms that contribute to the statistical significance.

\paragraph{Multiple-problem analysis.} 
The mean of the performance measure from multiple runs can be used as representative value of each algorithm on each problem. However, as stated above, averaging is sensitive to outliers, which needs to be considered especially because  optimization algorithms could have poor runs. For this reason, the median of the performance measure from the multiple runs can also be used as more robust statistic.

Both mean and median are sensitive to errors inside some $\epsilon$-neighborhood (i.e., small difference between their values that is not recognized by the ranking schemes of the non-parametric tests), which can additionally affect the statistical result. For these reasons, Deep Statistical Comparison (DSC) for comparing evolutionary algorithms was proposed \cite{eftimov2017novel}. Its main
contribution is its ranking scheme, which is based on the whole distribution, instead of using only one statistic to describe the distribution, such as mean or median.

The impact of the selection of the three above-presented transformations, which can be used to find a representative value for each algorithm on each problem, to the final result of the statistical analysis in the multiple-problem analysis is presented in \cite{eftimov2018impact}. 

\paragraph{Statistical tests.}
No matter which transformation is used, once the data for analysis is available, the next step is to select an appropriate statistical test. For pairwise comparison, the \emph{t test}  is the appropriate parametric one~\cite{Shes03a}, while its relevant non-parametric version is the  \emph{Wilcoxon signed rank test} \cite{Wilc45a}. 
In the case when more than two algorithms are involved, the parametric version is the repeated measurements \emph{ANOVA} \cite{Lind74a, Mont17a}, while its appropriate non-parametric tests are the \emph{Friedman rank-based test} \cite{Frie37a}, \emph{Friedman-aligned test} \cite{Garc09a}, and \emph{Iman-Davenport test} \cite{Garc09a}. Additionally, if the null hypothesis is rejected, same as the single-problem analysis, we should continue with a post-hoc procedure to define the pairs of algorithms that contribute to the statistical significance. 

Other non-parametric tests are the non-parametric rank based tests, which are suitable when the distribution assumptions are questionable~\cite{Shes03a}. Using them, the data is ranked and then the p-value is calculated for the ranks and not the actual data. This ranking helps to eliminate the problem of skewness and in handling extreme values. The \emph{permutation test} \cite{Pesa01a} estimates the permutation distribution by shuffling the data without replacement and identifying almost all possible values of the test statistic. The \emph{Page’s trend test} \cite{Derr14a} is also a non-parametric test that can be used to analyse convergence performance of evolutionary algorithms.

\paragraph{Post-hoc procedures.}
When we have more than two algorithms that are involved in the comparison, 
the appropriate statistical test can find statistical significance between the algorithms' performances, but it is not able to define the pairs of algorithms that contribute to this statistical significance. For this reason, if the null hypothesis is rejected, we should continue with a \emph{post-hoc} test.

The \emph{post-hoc} testing can be done in two scenarios: i) all pairwise comparisons and ii) multiple comparisons with a control algorithm. 
Let us assume that we have $k$ algorithms involved in the comparison, so in the first scenario we should perform $k(k-1)/2$ comparisons, and in the second one $k-1$. 

In the case of all pairwise comparisons, the post-hoc test statistic should be calculated. It depends on the appropriate statistical test that is used to compare all algorithms together, which rejected the null hypothesis. After that, the obtained p-values are corrected with some post-hoc procedure. For example, if the null hypothesis in the \emph{Friedman test}, \emph{Friedman aligned-ranks test}, or \emph{Iman–Davenport test}, is rejected, we can use the \emph{Nemenyi}, \emph{Holm}, \emph{Shaffer}, and \emph{Bergmann} correction to adapt the p-values and handle multiple-testing issues.  

The multiple comparisons with a control algorithm is the scenario when our newly developed is the control algorithm, and we are comparing it with state-of-the-art algorithms. Same as the previous scenario, the post-hoc statistic depends on the appropriate statistical test that is used to compare all algorithms together, which rejected the null hypothesis, and the obtained p-values are corrected with some post-hoc procedure. In the case of \emph{Friedman test}, \emph{Friedman aligned-ranks test} , or \emph{Iman–Davenport test}, appropriate post-hoc procedures are: \emph{Bonferroni}, \emph{Holm}, \emph{Hochberg}, \emph{Hommel}, \emph{Holland}, \emph{Rom}, \emph{Finner}, and \emph{Li}.

Another way to the multiple comparisons with a control algorithms is to perform all comparisons between the control algorithm and each other algorithm using some pairwise test. In this case, we should be careful when making a conclusion, since we are losing the control on the Family-Wise Error Rate~(FWER) when performing multiple pairwise comparisons. All obtained p-values will come from independent pairwise comparisons. The calculation of the true statistical significance for combining pairwise comparisons is presented in \cite{Garc09a,eftimov2017novel}.

More information about different post-hoc procedures and their application in benchmarking theory in evolutionary computation is presented in \cite{Garc09a}.

\subsection{Relevance Analysis}
\subsubsection{Motivation}
The third step of the recommended approach is related to the practical relevance of our statistical findings: are the differences really meaningful in practice or are they only statistical "artifacts" caused by an inadequate experimental design?
A typical example for these artifacts is a difference in performance, say $\delta$, which is statistically significant but of no practical relevance, because a value as small as $\delta$ cannot be measured in real-world scenarios. So, there is still a gap when transferring the learned knowledge from theory to practice.  
This happens because the statistical significance that exists is not scientifically meaningful in a practical sense. 
\begin{example}[Assembly line]
Let us assume that two optimization algorithms that should minimize the average time of a production process, e.g., an assembly line, are compared. The mean difference in performance is $\delta = 10^{-14}$, which is statistically significant. However, this difference has no meaning in reality, because it is far below the precision of the assembly line timer.
\end{example}
 For this reason, when we are performing a statistical analysis, we should also try to find the relevance of the statistical significance to real world applications. We should also mention that the practical significance depends on the specific problem being solved.
Additionally, this is also true in benchmarking performed for scientific publications, where the comparisons of the performance measures can be affected by several factors such as computer accuracy (i.e., floating points), variable types (4-byte float, 8-byte float, 10-byte float), or even the stopping criteria that is the error threshold when the algorithms are stopped. All these factors can result in different values, which does not represent the actual performance of the algorithms even if statistical significance is found.

\subsubsection{Severity: Relevance of Parametric Test Results}
In order to probe the meaningfulness of the statistically significant result, it is suggested to perform a post-data analysis. One such post data analysis is the severity measure, a meta statistical principle \cite{Mayo06a,Bart10a}. Severity describes the degree of support to decisions made using classical hypothesis testing. Severity takes into account the data and performs a post-data evaluation to scrutinize the decisions made by analyzing how well the data fits the testing framework. 
The severity is the actual power attained in the post data analysis and can be described separately for the decision of either rejecting or not rejecting the null hypothesis. 

The conclusions obtained from the hypothesis testing is dependent on sample size and can suffer from the \emph{problem of large $n$}. Severity deals with this problem directly~\citep{Mayo06a}.

 \subsubsection{Multiple-Problem Analysis} 
 We present two approaches that investigate the scientific meaningfulness of statistically significant results in the multiple-problem setting. 
 One approach is the \gls{crs4ea}, which is an empirical algorithm for comparing and ranking evolutionary algorithms~\cite{vevcek2014chess}. It makes a chess tournament where the optimization algorithms are considered as chess players and a comparison between the performance measures of two optimization algorithms as the outcome of a single game. A draw limit that defines when two performance measure values are equal should be specified by the user and it is a problem specific. At the end, each algorithm has its own rating which is a result from the tournament and the statistical analysis is performed using confidence intervals calculated using the algorithms rating. 
 
 The second approach is the \gls{pdsc}, which is a modification of the DSC approach used for testing for statistical significance~\citep{eftimov2019identifying}. The basic idea is that the data on each problem should be pre-processed with some practical level specified by a user, and after that involved with DSC to find relevant difference. Two pre-processing steps are proposed: i) sequential pre-processing, which pre-processes the performance measures from multiple runs in a sequential order, and ii) a Monte-Carlo approach, which pre-processes the performance measure values by using a Monte-Carlo approach to avoid the dependence of the practical significance on the order of the independent runs. A comparison between the \gls{crs4ea} and \gls{pdsc} is presented in \cite{eftimov2019identifying}. Using these two approaches, the analysis is made for a multiple-problem scenario. Additionally, the rankings from \gls{pdsc} obtained on a single-problem level can be used for single-problem analysis.

\subsection{Open Issues} 
 
An important aspect not addressed in this iteration of the document is the analysis of the benchmark problems themselves rather than the performance of the algorithms operating thereon. That is, which means exist for investigating structural characteristics of the benchmarking problem at hand? How can one (automatically) extract its most relevant information? How should this information be interpreted? There exists a variety of approaches for this, and each of them helps to improve the understanding of the respective problem, and in consequence may facilitate the design, selection and/or configuration of a suitable algorithm.

Linked to the above is a discussion of methods for visualizing problem landscapes. Visualizing the landscape of a continuous problem, or plotting approximate tours for a given \gls{tsp} instance, usually improves our understanding of its inherent challenges and reveals landscape characteristics such as multimodality. Moreover, such visualizations also help to study the search behavior of the algorithms under investigation. Unfortunately, the vast majority of works treats the issue of visualizing problems very poorly, so we will make sure to address this particular issue in the continuation of this document.

\section{Experimental Design}
\label{sec:07design}

\subsection{Design of Experiments (DoE)}\label{s:doe}
Unfortunately, many empirical evaluations of optimization algorithms are performed and reported without addressing basic experimental design considerations~\citep{Brow07a}. 
An important step to make this procedure more transparent and more objective is to use \gls{doe} and related techniques. 
They provide an algorithmic procedure to make comparisons in benchmarking more transparent.
Experimental design provides an excellent way of deciding which and how many algorithm runs should be performed 
so that the desired information can be obtained with the least number of runs. 

\gls{doe} is planning, conducting, analyzing, and interpreting controlled tests to evaluate the influence of the varied factors on the outcome of the experiments. The importance and the benefits of a well designed planned experiment have been summarized by \citet{Hoo1996joh}. 
\citet{John96a} suggests to report not only the run time of an algorithm, but also explain the corresponding adjustment process (preparation or tuning before the algorithm is run) in detail, and therefore to include the time for the adjustment in all reported running times to avoid a serious underestimate. 

The various key implications involved in the \gls{doe} are clearly explained in \citet{Klei11a}. A comprehensive list of the recent publications on design techniques can be found in \citet{Klei17a}. The various design strategies in the \gls{dace} are discussed by \citet{Sant03a}. 
\citet{Wagn10a} discusses important experimental design topics, e.g., ``How many replications of each design should be performed?'' or 
``How many  algorithm runs should be evaluated?''

This section discusses various important practical aspects of formulating the design of experiments for a stochastic optimization problem. 
The key principles are outlined. 
For a detailed reading of the \gls{doe}, the readers are referred to \citet{Mont17a} and \citet{Klei15a}. 

\subsection{Design Decisions}
Design decisions can be based on geometric or on statistical criteria~\citep{Puke93a, Sant03a}.
Regarding geometric criteria, 
two different design techniques can be distinguished:
The samples can be placed
either (1) on the boundaries, or
(2) in the interior 
of the design space.
The former technique is used in \gls{doe}
whereas \gls{dace} uses the latter approach.
An experiment is called \emph{sequential\/} if the experimental conduct at any stage depends on the results obtained so far.
Sequential approaches exist for both variants.

We recommend using factorial designs or space-filling designs instead of the commonly used \gls{ofat} designs.
When several factors are involved in an experiment, the \gls{ofat} design strategy is inefficient as it suffers from various limitations including huge number of experimental runs and inability to identify the interactions among the factors involved. It is highly recommended using a multi-factorial design \citep{Mont17a}. The Factorial designs are robust and faster when compared with \gls{ofat}. For a complete insight into Fully-Factorial and Fractional Factorial designs the readers are redirected to \citet{Mont17a}. 
The Taguchi design \cite{Roy01a} is a variation of the fractional factorial design strategy, which provides robust designs at better costs with fewer evaluations. The Plackett and Burman design \cite{Plac46a} are recommended for screening. 
The modern space-filling designs are sometimes more efficient and require fewer evaluations than the fractional designs, especially in case of non-linearity. Further information about space-filling designs can be found in \citet{Sant03a}. 

However, it is still an open question which design characteristics are important: ``$\ldots$ extensive empirical studies would be useful for better understanding what sorts of designs perform well and for which models''~\cite[p.~161]{Sant03a}.

\subsection{Designs for Benchmark Studies}
In the context of \gls{doe} and \gls{dace}, runs of an optimization algorithm instance will be treated as experiments. 
There are many degrees of freedom when an optimization algorithm instance is run.
In many cases optimization algorithms  require the determination of 
parameters (e.g., the population size in \glspl{es})  
 before the optimization run is performed.
From the viewpoint of an experimenter, design variables (factors) are the parameters that can be changed during an experiment. 
Generally, there are two different types of factors that influence the behavior of an optimization algorithm:
\begin{enumerate}
\item problem-specific factors, i.e., the objective function,
\item algorithm-specific factors, i.e., the population size of an \gls{es} and other parameters which need to be set to derive an executable algorithm instance.
\end{enumerate}
We will consider experimental designs that comprise problem-specific factors and algorithm-specific factors.
Algorithm-specific factors will be considered first.
{\em Implicit parameters\/} can be distinguished from {\em explicit parameters\/} (synonymously referred to as \emph{endogeneous} and \emph{exogeneous} in~\citep{BeySch2002:es}).
The latter are explicitly exposed to the user, whereas the former are often hidden, i.e., either made inaccessible to the user (e.g., when the algorithm code is not made available) or simply ``hidden'' in the implementation and not easily identifiable as a parameter that can be optimized.

An \emph{algorithm design\/} is a set of parameters, 
each representing one specific setting of the design variables of an algorithm and defining an algorithm instance.
A design can be specified by defining ranges of values for the design parameters.
Note that a design can contain none, one, several, or even infinitely many design points, each point representing an algorithm instance. 
Consider the set of explicit strategy parameters for \gls{PSO} algorithms with the following values:
swarm size $s=10$, cognitive parameter $c_1 \in [1.5, 2]$, social parameter $c_2 = 2$, starting value of the inertia weight $w_{\max} = 0.9$, final value of the inertia weight $w_{\text{scale}} = 0$, percentage of iterations for which $w_{\max}$ is reduced $w_{\text{iterscale}} =1$, and maximum value of the step size $v_{\max} = 100$.
This algorithm design 
contains infinitely many design points, because $c_1$ is not fixed.

\emph{Problem designs\/} provide information 
related to the optimization problem, such as the available resources (number of function evaluations) or 
the problem's dimension.

An \emph{experimental design\/} consists of a problem design and an algorithm design.
Benchmark studies require complex experimental designs, because they are combinations of several problem and algorithm designs. Furthermore, as discussed in Section~\ref{sec:sec5}, one or several performance measures must be specified.

\subsection{How to Select a Design for Benchmarking}
The following points have to be considered when designing an benchmark study\footnote{At the moment, this is only a list, which will be extended in forthcoming versions of this survey.}:
\begin{itemize}
\item What are the main goals of the experiment? (see Section~\ref{sec:sec2}) 
\item What is/are the test problem(s) and which (type of) instances do we select?  (see Section~\ref{sec:sec3}) 
\item How many algorithms are to be tested? (see Section~\ref{sec:sec4}) 
\item How many test problems/test classes are relevant for the study? (see Section~\ref{sec:sec3}) 
\item How tuning of algorithms has to be performed? (see Section~\ref{sec:sec4}) 
\item What validation procedures are considered to evaluate the results of the experiment? (see Section~\ref{sec:sec5}) 
\item How will the results be analyzed? (see Section~\ref{sec:sec6}) 
\item How will the results be presented? (see Section~\ref{sec:08presentation})
\item How are randomization and replicability of the experiment achieved? (see Section~\ref{sec:09reproducibility}) 
\end{itemize}

\subsection{Tuning Before Benchmarking}
\citet{Brow07a} discusses the importance of tuning an algorithm before benchmarking. \citet{Bart09e} state that comparisons of tuned versus untuned algorithms are not fair and should be avoided. 
During a benchmark study 
the employed parameter settings are extremely important
as they largely define the obtained performance. Depending
on the availability of code for the algorithms under scope
and time for parameter searches,
there are different possibilities to make a fair comparison:
\begin{itemize}
\item In the best case, the code for all methods is available.
It is then possible to perform a parameter search for each
problem and each algorithm via a tuning method.
Taking the best parameter sets for each method for each
problem ensures comparing the algorithms at their peak
performance.
\item If algorithm runs on the chosen problems take too
long for a full tuning process, one may however perform a simple
space-filling design on the parameter space, e.g., a
\gls{lhd} or a low-discrepancy point set~\cite{Mat99} with only a few design points
and repeats. This prevents misconfigurations of algorithms
as one probably easily gets into the ``ball park''~\citep{Dejo07a}
of relatively good parameter settings. Most likely, neither algorithm works
at its peak performance level, but the comparison is still fair. 
\item If no code other than for one's own algorithm is available,
one has to resort to comparing with default parameter values. For a new
algorithm, these could be determined by a common tuning process over
the whole problem set. Note however, that such comparison deliberately
abstains from setting good parameters for specific problems, even if
this would be attempted for any real-world application. 
\end{itemize}

\subsection{Open Issues} 

\begin{enumerate}[({O7}.1)]
    \item \emph{Best Designs.} \newline
    Same authors consider \glspl{lhd} as the default choice, even if for numerous applications a superiority of other space-filling or low-discrepancy designs has been demonstrated~\citet{Sant03a}. The question when to prefer i.i.d. uniform sampling, \glspl{lhd}, low-discrepancy point sets, other space-filling designs, or sets minimizing some other diversity criterion is largely open.  
\item \emph{Multiple Objectives.} \newline
    Sometimes properties of the objective function are used to determine the quality of a design. Therefore, it remains unclear how to measure the quality in settings where the objective function is unknown. Furthermore, problems occur if wrong assumptions about the objective function, e.g., linearity, are made. And, last but not least, in \gls{moo}, where no single objective can be specified, finding the optimal design can be very difficult~\citep{Sant03a}.
\end{enumerate}

\section{How to Present Results?}
\label{sec:08presentation}

\subsection{General Recommendations}
Several papers have been published in the last years, which give recommendations on how to report results.
As \citet{Gent94a} already stated in 1994, after having generated some good results in your benchmark study,
there are still many mistakes to make. They give the following recommendations:
\begin{enumerate}
    \item present statistics, i.e., statements such as ``algorithm $a$ outperforms $b$'' should be accompanied with suitable test results as described in Section~\ref{sec:sec6},
    \item do not push deadlines, i.e., do not reduce the quality of the report, because the deadline is approaching soon. Invest some time in planning: number of experiments, algorithm portfolio, hardness of the problem instances as discussed in Section~\ref{sec:07design},
    \item  and report negative results \label{neg-res}, i.e., present and discuss problem instances on which the algorithms fail (this is a key component of a good scientific report as discussed in this section).
\end{enumerate}

\citet{Barr95a}  in their classical work on reporting empirical results of heuristics specify a loose experimental setup methodology with the following steps:
\begin{enumerate}
    \item define the goals of the experiment,
    \item  select measure of performance and factors to explore,
    \item design and execute the experiment, 
    \item  analyze the data and draw conclusions, and finally 
    \item report the experimental results. 
\end{enumerate}
They then suggest eight guidelines for reporting results, in summary they are; reproducibility, specify all influential factors (code, computing environment, etc.), be precise regarding measures, specify parameters, use statistical experimental design, compare with other methods, reduce variability of results, ensure results are comprehensive. They then go on to clarify these points with examples.

\subsection{Reporting Methodologies}
Besides recommendations, that provide valuable hints on how to report results, there exist also methodologies, which employ a scientific methodology, e.g., based on hypothesis testing~\citep{Popp59a, Popp79a}. 
Such a methodology was proposed by \citet{Bart09e}. They propose organizing the presentation of experiments into seven parts, as follows:

\begin{enumerate}[({R}.1)]
\item \emph{Research question}\\ Briefly names the matter dealt with, the (possibly
very general) objective, preferably in one
sentence. This is used as the report's ``headline'' and related to the primary
model.
\label{enum-exp-struct:focus}
\item \emph{Pre-experimental planning}\\
Summarizes the first---possibly explorative---program runs, leading
to task and setup (R-\ref{enum-exp-struct:task} and
R-\ref{enum-exp-struct:setup})\label{enum-exp-struct:pre-experimental}.
Decisions on employed benchmark problems or performance measures should
be taken according to the data collected in preliminary runs.
The report on pre-experimental planning should also include negative results,
e.g., modifications to
an algorithm that did not work or a test problem that turned out
to be too hard, if they provide new insight.
\item \emph{Task}\\ Concretizes the question in focus and states
scientific claims and derived statistical hypotheses to test.
Note that one scientific claim may require several,
sometimes hundreds, of statistical hypotheses.
In case of a purely explorative study, as with the first test
of a new algorithm, statistical tests may not be applicable.
Still, the task should be formulated as precisely as possible. This step is
related to the experimental model.
\label{enum-exp-struct:task}
\item \emph{Setup}\\ Specifies problem design and algorithm design,
including the investigated algorithm, the controllable and
the fixed parameters, and the chosen
performance measuring. It also includes information about the  computational environment (hard- and software specification, e.g., the packages or libraries used). 
The information provided in this part
should be sufficient to replicate an experiment.
\label{enum-exp-struct:setup}
\item \emph{Results/Visualization}\\
Gives raw or produced (filtered) data on the experimental outcome and
additionally provides basic visualizations where meaningful. This is related to
the data model.
\label{enum-exp-struct:experiment-visualization}
\item \emph{Observations}\\
Describes exceptions from the expected, or unusual patterns
noticed, without subjective assessment or explanation.
As an example, it may be worthwhile to look at parameter
interactions.
Additional visualizations may help to clarify what happens.
\label{enum-exp-struct:observations}
\item \emph{Discussion}\\
Decides about the hypotheses specified in R-\ref{enum-exp-struct:task},
and provides necessarily subjective interpretations
of the recorded observations. Also places the results in a
wider context. The leading question here is: What did we learn?
\label{enum-exp-struct:conclusions}
\end{enumerate} 

This methodology was extended and refined in~\citet{Preu15a}.
It is important to divide parts
R-6 and R-7,
to facilitate different conclusions drawn by others, based
on the same results/observations.
This distinction into parts of
increasing subjectiveness is similar to
the suggestions of \citet{Barr95a},  who
distinguish between results, their analysis, and
the conclusions drawn by the experimenter.

Note that all of these parts are already included
in current good experimental reports. However,
they are usually not separated but wildly mixed.
Thus, we only suggest inserting labels into the
text to make the structure more obvious.

We also recommend keeping a journal of experiments
with single reports according to the above scheme
to enable referring to previous experiments later on.
This is useful even if single experiments do not
find their way into a publication, as it improves the
overview of subsequent experiments 
and helps to avoid repeated tests.

\subsection{Open Issues} 
Reporting negative results has many benefits, e.g., to demonstrate what has been done and does not work, so someone else will not do the same in the future. And they are valuable tools for illustrating the limitations of new approaches.
The presentation of negative results discussed above in \ref{neg-res} is not adequately accepted in the research community (cf. \citet{Gent94a}).  
Whereas a paper improving existing experimental results or outperforming another algorithm regularly gets accepted for publication, papers presenting negative results regularly will not.

\section{How to Guarantee Reproducibility?}
\label{sec:09reproducibility}

Reproducibility has been a topic of interest in the experimental analysis of
algorithms for many decades. Classical works~\cite{J2002ATGTTEAOA} advise
\emph{ensuring reproducibility}, but caution that the classical understanding
of reproducibility in computer science, i.e., running exactly the same code on
the same machine returns exactly the same measurements, differs substantially
from the understanding in other experimental sciences, i.e., a different
implementation of the experiment under similar conditions returns measurements
that lead to the same conclusions.

For example, the ``Reproducibility guidelines for AI research''\footnote{\url{http://folk.idi.ntnu.no/odderik/reproducibility_guidelines.pdf}}
intended to be adopted by the \gls{aaai} are clearly focused on the concept of reproducibility in computer science.

Trying to clearly define various reproducibility concepts, the \gls{acm} distinguishes among:\footnote{Quoting from:\\
\url{https://www.acm.org/publications/policies/artifact-review-and-badging-current}}

\begin{description}
\item[Repeatability] (Same team, same experimental setup) The measurement can
be obtained with stated precision by the same team using the same measurement
procedure, the same measuring system, under the same operating conditions, in
the same location on multiple trials.
For computational experiments, this
means that a researcher can reliably repeat her own computation.
\item[Reproducibility] (Different team, same experimental setup) The measurement
can be obtained with stated precision by a different team using the same
measurement procedure, the same measuring system, under the same operating
conditions, in the same or a different location on multiple trials.
For computational experiments, this means that an independent group can obtain the
same result using the author’s own artifacts.
\item[Replicability] (Different team, different experimental setup) The
measurement can be obtained with stated precision by a different team, a
different measuring system, in a different location on multiple trials.
For
computational experiments, this means that an independent group can obtain the
same result using artifacts which they develop completely independently.
\end{description}

The above classification helps to identify various levels of reproducibility,
reserving the term ``Replicability'' to the most scientifically useful, yet
hardest to achieve. There are many practical guidelines and software systems
available to achieve repeatibility and
reproducibility~\cite{GenGraMac1997hownotto, J2002ATGTTEAOA}, including code
versioning tools (Subversion and Git), data repositories (Zenodo), reproducible
documents (Rmarkdown and Jupyter notebooks), and reproducible software
environments (OSF\footnote{\url{https://osf.io/}}, CodeOcean and
Docker). 

Unfortunately, it is not so clear how to successfully achieve Replicability. For achieving replicability, one must give up on exactly reproducing the results and provide statistical guidelines that are commonly accepted by the field to provide sufficient evidence for a conclusion, even under different, but similar, experimental conditions.
What constitutes \emph{similar} experimental conditions depends on the experiment and there is no simple answer when benchmarking algorithms.
One step towards better replicability is to pre-register experimental designs~\cite{NosEbeHav2018preregistration} to fix the hypothesis and design of experiments. 
Preregistration reduces the risk of spurious results due to adaptations to data analysis. 
However, it is much harder to systematically control for adaptive computational experiments because, unlike randomized controlled trials, they are much easier to run and re-run prior to registration.


\section{Summary and Outlook}
\label{sec:sec10}
This survey compiles ideas and recommendations from more than a dozen researchers with different backgrounds and from different institutions around the world.
Its main goal is the promotion of best practice in benchmarking.
This version is the result of long and fruitful discussions among the authors. The authors agreed on eight essential topics, that should be considered in every benchmark study: goals, problems, algorithms, performance, analysis, design, presentation, and reproducibility.
These topics defined the section structure of this article.

While it is definitely not a textbook that explains every single approaches in detail, we hope it is a good starting point for setting up benchmark studies. It is basically a guide (similar to the famous hitch-hiker's guide to \gls{ec}~\citep{heit94a}) and has a long list of references, which covers classical papers as well as the most recent ones. Every section presents recommendations, best practice examples, and open issues.

As mentioned above, this survey is only the beginning of a wonderful journey. 
It can serve as a starting point for many activities that improve the quality of benchmark studies and enhance the quality of research in \gls{ec} and related fields.
Next steps can be as follows:
\begin{enumerate}
    \item offering tutorials and organizing workshops,
    \item compiling videos, which explain how to set up the experiments, analyze results, and report important findings,
    \item providing software tools,
    \item developing a comprehensible check-list, especially for beginners in benchmarking,
    \item including a discussion section in every section, which describes controversial topics and ideas. 
\end{enumerate}

Our final goal is to provide well-accepted guidelines (rules) that might be useful for authors, reviewers, and others. Consider the following (rudimentary and incomplete) checklist, that can serve as a guideline for authors and reviewers:
\begin{enumerate}
    \item goals: did the authors clearly state the reasons for this study?
    \item problems: is the selection of problem instances well motivated and justified? 
    \item algorithms: do comparisons include relevant competitors?
    \item performance: is the choice of the performance measure adequate?
    \item analysis: are standards from statistics considered?
    \item design: does the experimental setup enable efficient and fair experimentation? What measures are taken to avoid ``cherry-picking results''?
    \item presentation: are the results well organized and explained?
    \item reproducibility: data and code availability?
\end{enumerate}

Transparent, well accepted standards will improve the review process in \gls{ec} and related fields significantly. These common standards might also accelerate the review process, because they improve the quality of submissions and helps reviewers to write objective evaluations. 
Most importantly, it is not our intention to dictate specific test statistics, experimental designs, or performance measures. Instead, we claim that publications in \gls{ec} would improve, if authors explain, \emph{why} they have chosen this specific measure, tool, or design. And, last but not least, authors should describe the goal of their study.

Although we tried to include the most relevant contributions, we are aware that important contributions are missing.
Because the acceptance of the proposed recommendations is crucial,  we would like to invite more researchers to share their knowledge with us. 
Moreover, as the field of benchmarking is constantly changing, this article will be regularly updated and published on arXiv~\citep{Bart20gArxiv}. 
To get in touch, interested readers can use the associated e-mail address for this project:
\href{mailto:benchmarkingbestpractice@gmail.com}{benchmarkingbestpractice@gmail.com}.

There are several other initiatives that are trying to improve benchmarking standards in query-based optimization fields, e.g., 
the {Benchmarking Network}\footnote{\url{https://sites.google.com/view/benchmarking-network/}},  an initiative built to consolidate and to stimulate activities on benchmarking iterative optimization heuristics~\citep{Wein20a}.

In our opinion, starting and maintaining this public discussion is very important. Maybe, this survey poses more questions than answers, which is fine. Therefore, we conclude this article with a famous saying that is attributed to Richard Feynman\footnote{\url{https://en.wikiquote.org/w/index.php?title=Talk:Richard_Feynman&oldid=2681873\#\%22I_would_rather_have_questions_that_cannot_be_answered\%22}}:
\begin{quote}
    I would rather have questions that can't be answered than answers that can't be questioned.
\end{quote}


\subsection*{Acknowledgments}
{\footnotesize
This work has been initiated at Dagstuhl seminar 19431 on Theory of Randomized Optimization Heuristics,\footnote{\url{https://www.dagstuhl.de/19431}} and we gratefully acknowledge the support of the Dagstuhl seminar center to our community. 

We thank Carlos~M.~Fonseca for his important input and our fruitful discussion, which helped us shape the section on performance measures. We also thank participants of the Benchmarking workshops at GECCO 2020 and at PPSN 2020 for several suggestions to improve this paper. We thank Nikolaus Hansen for providing feedback on an earlier version of this survey.   

\medskip

C.~Doerr acknowledges support from the Paris Ile-de-France region and from a public grant as part of the Investissement d'avenir project, reference ANR-11-LABX-0056-LMH, LabEx LMH. 

J.~Bossek acknowledges support by the Australian Research Council (ARC) through grant DP190103894.

J.~Bossek and P.~Kerschke acknowledge support by the European Research Center for Information Systems (ERCIS).

S.~Chandrasekaran and T.~Bartz-Beielstein acknowledge support from the Ministerium f{\"u}r Kultur und Wissenschaft des Landes Nordrhein-Westfalen in the funding program FH Zeit f{\"u}r Forschung under the grant number 005-1703-0011 (OWOS). 

T.~Eftimov acknowledges support from the Slovenian Research Agency under research core funding No. P2-0098 and project No. Z2-1867.

A.~Fischbach and T.~Bartz-Beielstein acknowledge support from the German Federal Ministry of Education and Research in the funding program Forschung an Fachhochschulen under the grant number 13FH007IB6 (KOARCH). 

W.~La~Cava is supported by NIH grant K99-LM012926 from the National Library of Medicine. 

M.\@ L\'opez-Ib\'a\~nez is a ``Beatriz Galindo'' Senior Distinguished Researcher (BEAGAL 18/00053) funded by the Ministry of Science and Innovation of the Spanish Government.

K.M.~Malan acknowledges support by the National Research Foundation of South Africa (Grant Number: 120837).

B.~Naujoks and T.~Bartz-Beielstein acknowledge support from the European Commission’s H2020 programme, 
 H2020-MSCA-ITN-2016 UTOPIAE (grant agreement No.\ 722734),  as well as the DAAD (German Academic Exchange Service), Project-ID: 57515062 “Multi-objective Optimization for Artificial Intelligence Systems in Industry”. 

M.~Wagner acknowledges support by the ARC projects DE160100850, DP200102364, and DP210102670.

T.~Weise acknowledges support from the National Natural Science Foundation of China under Grant 61673359 and the Hefei Specially Recruited Foreign Expert program. 

We also acknowledge support from COST action 15140 on Improving Applicability of Nature-Inspired Optimisation by Joining Theory and Practice (ImAppNIO).
}

\clearpage
\printglossaries

\clearpage
\addcontentsline{toc}{section}{References}
\bibliographystyle{plainnat} 
\begin{small}
\bibliography{optbib/abbrev,optbib/journals,optbib/authors,optbib/biblio,optbib/crossref,bibliography}  

\providecommand{\MaxMinAntSystem}{{$\cal MAX$--$\cal MIN$} {A}nt {S}ystem}
  \providecommand{\Rpackage}[1]{#1} \providecommand{\SoftwarePackage}[1]{#1}
  \providecommand{\proglang}[1]{#1}
\begin{thebibliography}{212}
\providecommand{\natexlab}[1]{#1}
\providecommand{\url}[1]{\texttt{#1}}
\expandafter\ifx\csname urlstyle\endcsname\relax
  \providecommand{\doi}[1]{doi: #1}\else
  \providecommand{\doi}{doi: \begingroup \urlstyle{rm}\Url}\fi

\bibitem[Adam et~al.(2019)Adam, Alexandropoulos, Pardalos, and
  Vrahatis]{Deme19a}
Stavros~P Adam, Stamatios-Aggelos~N Alexandropoulos, Panos~M Pardalos, and
  Michael~N Vrahatis.
\newblock {No Free Lunch Theorem: A Review}.
\newblock In \emph{Approximation and Optimization}, pages 57~--~82. Springer,
  2019.

\bibitem[Agrawal et~al.(2020)Agrawal, Menzies, Minku, Wagner, and
  Yu]{Agrawal2020duo}
Amritanshu Agrawal, Tim Menzies, Leandro~L. Minku, Markus Wagner, and Zhe Yu.
\newblock Better software analytics via ``duo'': Data mining algorithms
  using/used-by optimizers.
\newblock \emph{Empirical Software Engineering}, 25\penalty0 (3):\penalty0
  2099--2136, 2020.
\newblock \doi{10.1007/s10664-020-09808-9}.
\newblock URL \url{https://doi.org/10.1007/s10664-020-09808-9}.

\bibitem[Akiba et~al.(2019)Akiba, Sano, Yanase, Ohta, and
  Koyama]{akiba2019optuna}
Takuya Akiba, Shotaro Sano, Toshihiko Yanase, Takeru Ohta, and Masanori Koyama.
\newblock Optuna: A next-generation hyperparameter optimization framework.
\newblock In \emph{Proceedings of the 25th ACM SIGKDD International Conference
  on Knowledge Discovery \& Data Mining}, pages 2623--2631, 2019.

\bibitem[Amini and Barr(1993)]{Amin93a}
Mohammad~M Amini and Richard~S Barr.
\newblock Network reoptimization algorithms: A statistically designed
  comparison.
\newblock \emph{ORSA Journal on Computing}, 5\penalty0 (4):\penalty0 395--409,
  1993.

\bibitem[Anderson and Darling(1952)]{Ande52a}
Theodore~W Anderson and Donald~A Darling.
\newblock Asymptotic theory of certain" goodness of fit" criteria based on
  stochastic processes.
\newblock \emph{The annals of mathematical statistics}, pages 193--212, 1952.

\bibitem[Ans\'{o}tegui et~al.(2015)Ans\'{o}tegui, Malitsky, Samulowitz,
  Sellmann, and Tierney]{GGA}
Carlos Ans\'{o}tegui, Yuri Malitsky, Horst Samulowitz, Meinolf Sellmann, and
  Kevin Tierney.
\newblock Model-based genetic algorithms for algorithm configuration.
\newblock In \emph{Proc.~of International Conf.~on Artificial Intelligence
  (IJCAI'15)}, pages 733--739. AAAI, 2015.

\bibitem[Arnold(2012)]{arnold2012noisy}
Dirk~V. Arnold.
\newblock \emph{{Noisy Optimization with Evolution Strategies}}, volume~8.
\newblock {Springer}, 2012.

\bibitem[Auger and Doerr(2011)]{AugerD11}
Anne Auger and Benjamin Doerr.
\newblock \emph{Theory of Randomized Search Heuristics}.
\newblock World Scientific, 2011.

\bibitem[Auger and Hansen(2005)]{AH2005PEOAALSEA}
Anne Auger and Nikolaus Hansen.
\newblock Performance evaluation of an advanced local search evolutionary
  algorithm.
\newblock In \emph{{Proceedings of the IEEE Congress on Evolutionary
  Computation}}, pages 1777--1784. {IEEE}, 2005.

\bibitem[{B{\"a}ck} et~al.(1997){B{\"a}ck}, Fogel, and Michalewicz]{back97a}
Thomas {B{\"a}ck}, David~B. Fogel, and Zbigniew Michalewicz.
\newblock \emph{Handbook of Evolutionary Computation}.
\newblock IOP Publishing Ltd., GBR, 1st edition, 1997.

\bibitem[Barr et~al.(1995)Barr, Golden, Kelly, Resende, and Stewart]{Barr95a}
Richard~S Barr, Bruce~L Golden, James~P Kelly, Mauricio~GC Resende, and
  William~R Stewart.
\newblock Designing and reporting on computational experiments with heuristic
  methods.
\newblock \emph{Journal of Heuristics}, 1\penalty0 (1):\penalty0 9--32, 1995.

\bibitem[Bartlett(1937)]{Bart37a}
Maurice~Stevenson Bartlett.
\newblock Properties of sufficiency and statistical tests.
\newblock \emph{Proceedings of the Royal Society of London. Series
  A-Mathematical and Physical Sciences}, 160\penalty0 (901):\penalty0 268--282,
  1937.

\bibitem[Bartz-Beielstein(2006)]{Bart06a}
Thomas Bartz-Beielstein.
\newblock \emph{{Experimental Research in Evolutionary Computation---The New
  Experimentalism}}.
\newblock Natural Computing Series. Springer, 2006.

\bibitem[Bartz-Beielstein and Preuss(2010)]{Bart09e}
Thomas Bartz-Beielstein and Mike Preuss.
\newblock {The Future of Experimental Research}.
\newblock In Thomas Bartz-Beielstein, Marco Chiarandini, Luis Paquete, and Mike
  Preuss, editors, \emph{Experimental Methods for the Analysis of Optimization
  Algorithms}, pages 17--46. Springer, Berlin, Heidelberg, New York, 2010.

\bibitem[Bartz-Beielstein et~al.(2004)Bartz-Beielstein, Parsopoulos, and
  Vrahatis]{Bart04a}
Thomas Bartz-Beielstein, Konstantinos~E Parsopoulos, and Michael~N Vrahatis.
\newblock Design and analysis of optimization algorithms using computational
  statistics.
\newblock \emph{Applied Numerical Analysis \& Computational Mathematics},
  1\penalty0 (2):\penalty0 413--433, 2004.

\bibitem[Bartz-Beielstein et~al.(2005)Bartz-Beielstein, Lasarczyk, and
  Preuss]{bartz05s}
Thomas Bartz-Beielstein, Christian~WG Lasarczyk, and Mike Preuss.
\newblock {Sequential Parameter Optimization}.
\newblock In \emph{{Proceedings of the 2005 IEEE Congress on Evolutionary
  Computation}}, volume~1, pages 773~--~780. IEEE, 2005.

\bibitem[Bartz-Beielstein et~al.(2010)Bartz-Beielstein, Chiarandini, Paquete,
  and Preuss]{Bart10a}
Thomas Bartz-Beielstein, Marco Chiarandini, Lu{\'\i}s Paquete, and Mike Preuss.
\newblock \emph{Experimental methods for the analysis of optimization
  algorithms}.
\newblock Springer, 2010.

\bibitem[Bartz-Beielstein et~al.(2017)Bartz-Beielstein, Gentile, and
  Zaefferer]{Bart17parxiv}
Thomas Bartz-Beielstein, Lorenzo Gentile, and Martin Zaefferer.
\newblock In a nutshell: Sequential parameter optimization.
\newblock Technical report, TH K{\"o}ln, 2017.

\bibitem[{Bartz-Beielstein} et~al.(2020){Bartz-Beielstein}, {Doerr}, {Bossek},
  {Chandrasekaran}, {Eftimov}, {Fischbach}, {Kerschke}, {Lopez-Ibanez},
  {Malan}, {Moore}, {Naujoks}, {Orzechowski}, {Volz}, {Wagner}, and
  {Weise}]{Bart20gArxiv}
Thomas {Bartz-Beielstein}, Carola {Doerr}, Jakob {Bossek}, Sowmya
  {Chandrasekaran}, Tome {Eftimov}, Andreas {Fischbach}, Pascal {Kerschke},
  Manuel {Lopez-Ibanez}, Katherine~M. {Malan}, Jason~H. {Moore}, Boris
  {Naujoks}, Patryk {Orzechowski}, Vanessa {Volz}, Markus {Wagner}, and Thomas
  {Weise}.
\newblock {Benchmarking in Optimization: Best Practice and Open Issues}.
\newblock \emph{arXiv e-prints}, art. arXiv:2007.03488, {July} 2020.

\bibitem[Beiranvand et~al.(2017)Beiranvand, Hare, and Lucet]{Beir17a}
Vahid Beiranvand, Warren Hare, and Yves Lucet.
\newblock {Best Practices for Comparing Optimization Algorithms}.
\newblock \emph{{Optimization and Engineering}}, 18\penalty0 (4):\penalty0
  815~--~848, 2017.

\bibitem[Bergstra et~al.(2013)Bergstra, Yamins, and Cox]{bergstra2013hyperopt}
James Bergstra, Dan Yamins, and David~D Cox.
\newblock Hyperopt: A python library for optimizing the hyperparameters of
  machine learning algorithms.
\newblock In \emph{Proceedings of the 12th Python in science conference},
  volume~13, page~20. Citeseer, 2013.

\bibitem[Beyer and Schwefel(2002)]{BeySch2002:es}
Hans-Georg Beyer and Hans-Paul Schwefel.
\newblock Evolution strategies: A comprehensive introduction.
\newblock \emph{Natural Computing}, 1:\penalty0 3--52, 2002.

\bibitem[Birattari et~al.(2002)Birattari, St{\"u}tzle, Paquete, and
  Varrentrapp]{BirStuPaqVar02:gecco}
Mauro Birattari, Thomas St{\"u}tzle, Lu{\'\i}s Paquete, and Klaus Varrentrapp.
\newblock A racing algorithm for configuring metaheuristics.
\newblock In W.~B. Langdon et~al., editors, \emph{Proceedings of the Genetic
  and Evolutionary Computation Conference, GECCO 2002}, pages 11--18. Morgan
  Kaufmann Publishers, San Francisco, CA, 2002.

\bibitem[Birattari et~al.(2003)Birattari, Paquete, and
  Stützle]{BirattariPS03ClassificationSBOH}
Mauro Birattari, Luis Paquete, and Thomas Stützle.
\newblock Classification of metaheuristics and design of experiments for the
  analysis of components.
\newblock
  \url{https://www.researchgate.net/publication/2557723_Classification_of_Metaheuristics_and_Design_of_Experiments_for_the_Analysis_of_Components},
  2003.
\newblock technical report.

\bibitem[Bischl et~al.(2016)Bischl, Kerschke, Kotthoff, Lindauer, Malitsky,
  Fr{\'e}chette, Hoos, Hutter, {Leyton-Brown}, Tierney, and
  Vanschoren]{Bischl2016aslib}
Bernd Bischl, Pascal Kerschke, Lars Kotthoff, Thomas~Marius Lindauer, Yuri
  Malitsky, Alexandre Fr{\'e}chette, Holger~H. Hoos, Frank Hutter, Kevin
  {Leyton-Brown}, Kevin Tierney, and Joaquin Vanschoren.
\newblock {ASlib: A Benchmark Library for Algorithm Selection}.
\newblock \emph{{Artificial Intelligence (AIJ)}}, 237:\penalty0 41~--~58, 2016.

\bibitem[Bliek et~al.(2020)Bliek, Verwer, and de~Weerdt]{bliek2020black}
Laurens Bliek, Sicco Verwer, and Mathijs de~Weerdt.
\newblock Black-box mixed-variable optimisation using a surrogate model that
  satisfies integer constraints.
\newblock \emph{arXiv preprint arXiv:2006.04508}, 2020.

\bibitem[Bokhari et~al.(2020)Bokhari, Alexander, and
  Wagner]{Bokhari2020validationNoise}
Mahmoud~A. Bokhari, Brad Alexander, and Markus Wagner.
\newblock Towards rigorous validation of energy optimisation experiments.
\newblock In \emph{Proceedings of the 2020 Genetic and Evolutionary Computation
  Conference}, GECCO ’20, page 1232–1240, New York, NY, USA, 2020.
  Association for Computing Machinery.
\newblock ISBN 9781450371285.
\newblock \doi{10.1145/3377930.3390245}.
\newblock URL \url{https://doi.org/10.1145/3377930.3390245}.

\bibitem[Bonyadi et~al.(2013)Bonyadi, Michalewicz, and
  Barone]{Bonyadi2013ttpcec}
Mohammad~Reza Bonyadi, Zbigniew Michalewicz, and Luigi Barone.
\newblock {The Travelling Thief Problem: The First Step in the Transition from
  Theoretical Problems to Realistic Problems}.
\newblock In \emph{{2013 IEEE Congress on Evolutionary Computation}}, pages
  1037~--~1044. IEEE, 2013.

\bibitem[Bonyadi et~al.(2019)Bonyadi, Michalewicz, Wagner, and
  Neumann]{Bonyadi2019opportunities}
Mohammad~Reza Bonyadi, Zbigniew Michalewicz, Markus Wagner, and Frank Neumann.
\newblock {Evolutionary Computation for Multicomponent Problems: Opportunities
  and Future Directions}.
\newblock In \emph{{Optimization in Industry: Present Practices and Future
  Scopes}}, pages 13~--~30. Springer, 2019.
\newblock \doi{10.1007/978-3-030-01641-8_2}.

\bibitem[Bossek et~al.(2019)Bossek, Kerschke, Neumann, Wagner, Neumann, and
  Trautmann]{Bossek2019tspEvolve}
Jakob Bossek, Pascal Kerschke, Aneta Neumann, Markus Wagner, Frank Neumann, and
  Heike Trautmann.
\newblock {Evolving Diverse TSP Instances by Means of Novel and Creative
  Mutation Operators}.
\newblock In \emph{{Proc. of the 15th ACM/SIGEVO Conference on Foundations of
  Genetic Algorithms}}, pages 58~--~71. {ACM}, 2019.

\bibitem[Bossek et~al.(2020{\natexlab{a}})Bossek, Kerschke, and
  Trautmann]{BKTAMOPOPAAASOOA}
Jakob Bossek, Pascal Kerschke, and Heike Trautmann.
\newblock {A Multi-Objective Perspective on Performance Assessment and
  Automated Selection of Single-Objective Optimization Algorithms}.
\newblock \emph{{Applied Soft Computing Journal ({ASOC})}}, 88:\penalty0
  105901, {March} 2020{\natexlab{a}}.

\bibitem[Bossek et~al.(2020{\natexlab{b}})Bossek, Kerschke, and
  Trautmann]{BossekKT2020AnytimeBehavior}
Jakob Bossek, Pascal Kerschke, and Heike Trautmann.
\newblock {Anytime Behavior of Inexact TSP Solvers and Perspectives for
  Automated Algorithm Selection}.
\newblock In \emph{Proc.~of the IEEE Congress on Evolutionary Computation}.
  IEEE, 2020{\natexlab{b}}.
\newblock A preprint of this manuscript can be found at
  \url{https://arxiv.org/abs/2005.13289}.

\bibitem[Boussa{\"{\i}}d et~al.(2013)Boussa{\"{\i}}d, Lepagnot, and
  Siarry]{BoussaidLS13ClassificationSBOH}
Ilhem Boussa{\"{\i}}d, Julien Lepagnot, and Patrick Siarry.
\newblock A survey on optimization metaheuristics.
\newblock \emph{Information Sciences}, 237:\penalty0 82--117, 2013.
\newblock \doi{10.1016/j.ins.2013.02.041}.
\newblock URL \url{https://doi.org/10.1016/j.ins.2013.02.041}.

\bibitem[Branke(1998)]{branke1998creating}
J{\"u}rgen Branke.
\newblock {Creating Robust Solutions by Means of Evolutionary Algorithms}.
\newblock In \emph{{International Conference on Parallel Problem Solving from
  Nature}}, pages 119~--~128. Springer, 1998.

\bibitem[Branke et~al.(2001)Branke, Schmidt, and Schmeck]{Bra01}
J{\"u}rgen Branke, Christian Schmidt, and Hartmut Schmeck.
\newblock {Efficient Fitness Estimation in Noisy Environments}.
\newblock In \emph{Genetic and Evolutionary Computation Conference (GECCO'01)},
  pages 243~--~250. Morgan Kaufmann, 2001.

\bibitem[Brownlee(2007)]{Brow07a}
Jason Brownlee.
\newblock {A Note on Research Methodology and Benchmarking Optimization
  Algorithms}.
\newblock Technical report, {Complex Intelligent Systems Laboratory (CIS),
  Centre for Information Technology Research (CITR), Faculty of Information and
  Communication Technologies (ICT), Swinburne University of Technology,
  Victoria, Australia, Technical Report ID 70125}, 2007.

\bibitem[Carrano et~al.(2011)Carrano, Wanner, and Takahashi]{Carr11a}
Eduardo~G Carrano, Elizabeth~F Wanner, and Ricardo~HC Takahashi.
\newblock A multicriteria statistical based comparison methodology for
  evaluating evolutionary algorithms.
\newblock \emph{IEEE Transactions on Evolutionary Computation}, 15\penalty0
  (6):\penalty0 848--870, 2011.

\bibitem[Cauwet and Teytaud(2016)]{cauwet2016noisy}
Marie-Liesse Cauwet and Olivier Teytaud.
\newblock {Noisy Optimization: Fast Convergence Rates with Comparison-based
  Algorithms}.
\newblock In \emph{{Proceedings of the Genetic and Evolutionary Computation
  Conference 2016}}, pages 1101~--~1106, 2016.

\bibitem[Chapman et~al.(2000)Chapman, Clinton, Kerber, Khabaza, Reinartz,
  Shearer, and Wirth]{Chap00a}
Pete Chapman, Julian Clinton, Randy Kerber, Thomas Khabaza, Thomas Reinartz,
  Colin Shearer, and R{\"u}diger Wirth.
\newblock {CRISP-DM 1.0: Step-by-Step Data Mining Guide}.
\newblock Technical report, SPSS Inc., 2000.

\bibitem[Chiarandini et~al.(2007)Chiarandini, Paquete, Preuss, and
  Ridge]{Chia07a}
Marco Chiarandini, Luis Paquete, Mike Preuss, and Enda Ridge.
\newblock {Experiments on Metaheuristics: Methodological Overview and Open
  Issues}.
\newblock Technical report, Institut for Matematik og Datalogi Syddansk
  Universitet, 2007.

\bibitem[Christofides(1976)]{christofides1976}
Nicos Christofides.
\newblock {The Vehicle Routing Problem}.
\newblock \emph{Revue fran{\c{c}}aise d'automatique, d'informatique et de
  recherche op{\'e}rationnelle (RAIRO). Recherche op{\'e}rationnelle},
  10\penalty0 (1):\penalty0 55~--~70, 1976.
\newblock URL \url{http://www.numdam.org/item?id=RO_1976__10_1_55_0}.

\bibitem[{\v C}repin{\v s}ek et~al.(2014){\v C}repin{\v s}ek, Liu, and
  Mernik]{Crep14a}
Matej {\v C}repin{\v s}ek, Shih-Hsi Liu, and Marjan Mernik.
\newblock {Replication and Comparison of Computational Experiments in Applied
  Evolutionary Computing: Common Pitfalls and Guidelines to Avoid Them}.
\newblock \emph{{Applied Soft Computing}}, 19:\penalty0 161~--~170, {June}
  2014.

\bibitem[Crowder et~al.(1979)Crowder, Dembo, and Mulvey]{Crow79a}
Harlan Crowder, Ron~S Dembo, and John~M Mulvey.
\newblock On reporting computational experiments with mathematical software.
\newblock \emph{ACM Transactions on Mathematical Software (TOMS)}, 5\penalty0
  (2):\penalty0 193--203, 1979.

\bibitem[Culberson(1998)]{Cul1998ec}
Joseph~C. Culberson.
\newblock On the futility of blind search: An algorithmic view of ``no free
  lunch''.
\newblock \emph{Evolutionary Computation}, 6\penalty0 (2):\penalty0 109--127,
  1998.
\newblock \doi{10.1162/evco.1998.6.2.109}.

\bibitem[De~Jong(2007)]{Dejo07a}
Kenneth De~Jong.
\newblock {Parameter Setting in EAs: a 30 Year Perspective}.
\newblock In \emph{{Parameter Setting in Evolutionary Algorithms}}, pages
  1~--~18. Springer, 2007.

\bibitem[De~Jonge and van~den Berg(2020)]{Dejonge2020sensitivity}
Marleen De~Jonge and Daan van~den Berg.
\newblock Parameter sensitivity patterns in the plant propagation algorithm.
\newblock In \emph{IJCCI}, page 92–99, 2020.

\bibitem[de~Souza et~al.(2020)de~Souza, da~Silva, Chaves, and
  Bernardino]{DESO2020}
Lucas Augusto~M\"{u}ller de~Souza, Jos{\'{e}} Eduardo~Henriques da~Silva,
  Luciano~Jerez Chaves, and Heder~Soares Bernardino.
\newblock A benchmark suite for designing combinational logic circuits via
  metaheuristics.
\newblock \emph{Applied Soft Computing}, 91:\penalty0 106246, {June} 2020.
\newblock \doi{10.1016/j.asoc.2020.106246}.
\newblock URL \url{https://doi.org/10.1016/j.asoc.2020.106246}.

\bibitem[Derrac et~al.(2011)Derrac, Garc{\'\i}a, Molina, and Herrera]{Derr11a}
Joaqu{\'\i}n Derrac, Salvador Garc{\'\i}a, Daniel Molina, and Francisco
  Herrera.
\newblock A practical tutorial on the use of nonparametric statistical tests as
  a methodology for comparing evolutionary and swarm intelligence algorithms.
\newblock \emph{Swarm and Evolutionary Computation}, 1\penalty0 (1):\penalty0
  3--18, 2011.

\bibitem[Derrac et~al.(2014)Derrac, Garc{\'\i}a, Hui, Suganthan, and
  Herrera]{Derr14a}
Joaqu{\'\i}n Derrac, Salvador Garc{\'\i}a, Sheldon Hui, Ponnuthurai~Nagaratnam
  Suganthan, and Francisco Herrera.
\newblock Analyzing convergence performance of evolutionary algorithms: A
  statistical approach.
\newblock \emph{Information Sciences}, 289:\penalty0 41--58, 2014.

\bibitem[Devore(2011)]{devore2011probability}
Jay~L Devore.
\newblock \emph{Probability and Statistics for Engineering and the Sciences}.
\newblock Cengage learning, 2011.

\bibitem[Doerr and Neumann(2020)]{DoerrN20}
Benjamin Doerr and Frank Neumann.
\newblock \emph{Theory of Evolutionary Computation -- Recent Developments in
  Discrete Optimization}.
\newblock Springer, 2020.

\bibitem[Doerr et~al.(2019)Doerr, Doerr, and Lengler]{DoerrDL19}
Benjamin Doerr, Carola Doerr, and Johannes Lengler.
\newblock Self-adjusting mutation rates with provably optimal success rules.
\newblock In \emph{Proceedings of the Genetic and Evolutionary Computation
  Conference}, pages 1479~--~1487. ACM, 2019.

\bibitem[{Doerr} et~al.(2018){Doerr}, {Wang}, {Ye}, {van Rijn}, and
  {B{\"a}ck}]{IOHprofiler}
Carola {Doerr}, Hao {Wang}, Furong {Ye}, Sander {van Rijn}, and Thomas
  {B{\"a}ck}.
\newblock {IOHprofiler: A Benchmarking and Profiling Tool for Iterative
  Optimization Heuristics}.
\newblock \emph{arXiv e-prints}, art. arXiv:1810.05281, Oct 2018.
\newblock Wiki page of IOHprofiler is available at
  \url{https://iohprofiler.github.io/}.

\bibitem[Doerr et~al.(2020)Doerr, Ye, Horesh, Wang, Shir, and
  B{\"{a}}ck]{DoerrYHWSB20}
Carola Doerr, Furong Ye, Naama Horesh, Hao Wang, Ofer~M. Shir, and Thomas
  B{\"{a}}ck.
\newblock {Benchmarking Discrete Optimization Heuristics with IOHprofiler}.
\newblock \emph{Applied Soft Computing}, 88:\penalty0 106027, 2020.

\bibitem[Dolan and Mor{\'e}(2002)]{Dola02a}
Elizabeth~D Dolan and Jorge~J Mor{\'e}.
\newblock Benchmarking optimization software with performance profiles.
\newblock \emph{Mathematical programming}, 91\penalty0 (2):\penalty0 201--213,
  2002.

\bibitem[Dorigo et~al.(2006)Dorigo, Birattari, and St{\"u}tzle]{DorBirStu06:ci}
Marco Dorigo, Mauro Birattari, and Thomas St{\"u}tzle.
\newblock Ant colony optimization: Artificial ants as a computational
  intelligence technique.
\newblock \emph{IEEE Computational Intelligence Magazine}, 1\penalty0
  (4):\penalty0 28--39, 2006.

\bibitem[Dueck and Scheuer(1990)]{DueckThreshold}
Gunter Dueck and Tobias Scheuer.
\newblock Threshold accepting: a general purpose optimization algorithm
  appearing superior to simulated annealing.
\newblock \emph{J. Comput. Phys.}, 90:\penalty0 161--175, 1990.

\bibitem[Eftimov and Koro{\v{s}}ec(2018)]{eftimov2018impact}
Tome Eftimov and Peter Koro{\v{s}}ec.
\newblock The impact of statistics for benchmarking in evolutionary computation
  research.
\newblock In \emph{Proceedings of the Genetic and Evolutionary Computation
  Conference Companion}, pages 1329--1336, 2018.

\bibitem[Eftimov and Koro{\v{s}}ec(2019)]{eftimov2019identifying}
Tome Eftimov and Peter Koro{\v{s}}ec.
\newblock Identifying practical significance through statistical comparison of
  meta-heuristic stochastic optimization algorithms.
\newblock \emph{Applied Soft Computing}, 85:\penalty0 105862, 2019.

\bibitem[Eftimov et~al.(2017)Eftimov, Koro{\v{s}}ec, and
  Seljak]{eftimov2017novel}
Tome Eftimov, Peter Koro{\v{s}}ec, and Barbara~Korou{\v{s}}i{\'c} Seljak.
\newblock {A Novel Approach to Statistical Comparison of Meta-Heuristic
  Stochastic Optimization Algorithms Using Deep Statistics}.
\newblock \emph{Information Sciences}, 417:\penalty0 186~--~215, 2017.

\bibitem[Eftimov et~al.(2020)Eftimov, Petelin, and
  Koro{\v{s}}ec]{eftimov2020dsctool}
Tome Eftimov, Ga{\v{s}}per Petelin, and Peter Koro{\v{s}}ec.
\newblock Dsctool: A web-service-based framework for statistical comparison of
  stochastic optimization algorithms.
\newblock \emph{Applied Soft Computing}, 87:\penalty0 105977, 2020.

\bibitem[Eggensperger et~al.(2013)Eggensperger, Feurer, Hutter, Bergstra,
  Snoek, Hoos, and {Leyton-Brown}]{eggensperger2013}
Katharina Eggensperger, Matthias Feurer, Frank Hutter, James Bergstra, Jasper
  Snoek, Holger~H. Hoos, and Kevin {Leyton-Brown}.
\newblock {Towards an Empirical Foundation for Assessing Bayesian Optimization
  of Hyperparameters}.
\newblock In \emph{{NIPS Workshop on Bayesian Optimization in Theory and
  Practice}}, volume~10, December 2013.

\bibitem[Eiben and Jelasity(2002)]{Eibe02a}
{\'A}goston~Endre Eiben and M{\'a}rk Jelasity.
\newblock {A Critical Note on Experimental Research Methodology in EC}.
\newblock In \emph{{Proceedings of the 2002 IEEE Congress on Evolutionary
  Computation}}, volume~1, pages 582~--~587. IEEE, 2002.

\bibitem[Eiben and Smit(2011)]{eibe12a}
{\'A}goston~Endre Eiben and Selmar~K Smit.
\newblock {Evolutionary Algorithm Parameters and Methods to Tune Them}.
\newblock In \emph{Autonomous search}, pages 15~--~36. Springer, 2011.

\bibitem[Eiben and Smith(2015)]{eibe03a}
{\'A}goston~Endre Eiben and James~E Smith.
\newblock \emph{{Introduction to Evolutionary Computing}}.
\newblock Natural Computing. Springer, 2 edition, 2015.

\bibitem[Fialho et~al.(2010)Fialho, Costa, Schoenauer, and Sebag]{FialhoCSS10}
{\'{A}}lvaro Fialho, Lu{\'{\i}}s~Da Costa, Marc Schoenauer, and Mich{\`{e}}le
  Sebag.
\newblock Analyzing bandit-based adaptive operator selection mechanisms.
\newblock \emph{Annals of Mathematics and Artificial Intelligence},
  60:\penalty0 25~--~64, 2010.

\bibitem[Finck et~al.(2015)Finck, Hansen, Ros, and Auger]{FHRA2015CDR1}
Steffen Finck, Nikolaus Hansen, Raymond Ros, and Anne Auger.
\newblock {COCO Documentation, Release 15.03}, {November} 2015.
\newblock URL \url{http://coco.lri.fr/COCOdoc/COCO.pdf}.

\bibitem[Fischbach and Bartz-Beielstein(2020)]{Fisc18a}
Andreas Fischbach and Thomas Bartz-Beielstein.
\newblock Improving the reliability of test functions generators.
\newblock \emph{Applied Soft Computing}, 92:\penalty0 106315, 2020.
\newblock ISSN 1568-4946.
\newblock \doi{https://doi.org/10.1016/j.asoc.2020.106315}.
\newblock URL
  \url{http://www.sciencedirect.com/science/article/pii/S1568494620302556}.

\bibitem[Fleming and Wallace(1986)]{Flem86a}
Philip~J. Fleming and John~J. Wallace.
\newblock How not to lie with statistics: The correct way to summarize
  benchmark results.
\newblock \emph{Commun. ACM}, 29\penalty0 (3):\penalty0 218--221, {March} 1986.

\bibitem[Fletcher(1976)]{fletcher1976}
Roger Fletcher.
\newblock Conjugate gradient methods for indefinite systems.
\newblock In \emph{Numerical analysis}, pages 73--89. Springer, 1976.

\bibitem[Fr\'{e}chette et~al.(2016)Fr\'{e}chette, Kotthoff, Michalak, Rahwan,
  Hoos, and Leyton-Brown]{Frechette2016shapley}
Alexandre Fr\'{e}chette, Lars Kotthoff, Tomasz Michalak, Talal Rahwan,
  Holger~H. Hoos, and Kevin Leyton-Brown.
\newblock Using the shapley value to analyze algorithm portfolios.
\newblock In \emph{Proc.~of the Thirtieth AAAI Conference on Artificial
  Intelligence}, pages 3397~–--~3403. AAAI Press, 2016.

\bibitem[Friedman(1937)]{Frie37a}
Milton Friedman.
\newblock The use of ranks to avoid the assumption of normality implicit in the
  analysis of variance.
\newblock \emph{Journal of the american statistical association}, 32\penalty0
  (200):\penalty0 675--701, 1937.

\bibitem[Gallagher(2016)]{GALL2016}
Marcus Gallagher.
\newblock Towards improved benchmarking of black-box optimization algorithms
  using clustering problems.
\newblock \emph{Soft Computing}, 20\penalty0 (10):\penalty0 3835--3849, {March}
  2016.
\newblock \doi{10.1007/s00500-016-2094-1}.
\newblock URL \url{https://doi.org/10.1007/s00500-016-2094-1}.

\bibitem[Garc{\'\i}a et~al.(2009)Garc{\'\i}a, Molina, Lozano, and
  Herrera]{Garc09a}
Salvador Garc{\'\i}a, Daniel Molina, Manuel Lozano, and Francisco Herrera.
\newblock A study on the use of non-parametric tests for analyzing the
  evolutionary algorithms' behaviour: a case study on the cec'2005 special
  session on real parameter optimization.
\newblock \emph{Journal of Heuristics}, 15\penalty0 (6):\penalty0 617, 2009.

\bibitem[Garc\'{\i}a-Mart\'{\i}nez et~al.(2012)Garc\'{\i}a-Mart\'{\i}nez,
  Rodr{\'\i}guez, and Lozano]{GarRodLoz2012soco}
Carlos Garc\'{\i}a-Mart\'{\i}nez, Francisco~J. Rodr{\'\i}guez, and Manuel
  Lozano.
\newblock Arbitrary function optimisation with metaheuristics: No free lunch
  and real-world problems.
\newblock \emph{Soft Computing}, 16\penalty0 (12):\penalty0 2115--2133, 2012.
\newblock \doi{10.1007/s00500-012-0881-x}.

\bibitem[Garden and Engelbrecht(2014)]{GARD2014}
Robert~W. Garden and Andries~P. Engelbrecht.
\newblock Analysis and classification of optimisation benchmark functions and
  benchmark suites.
\newblock In \emph{2014 {IEEE} Congress on Evolutionary Computation ({CEC})}.
  {IEEE}, {July} 2014.
\newblock \doi{10.1109/cec.2014.6900240}.
\newblock URL \url{https://doi.org/10.1109/cec.2014.6900240}.

\bibitem[Gent and Walsh(1994)]{Gent94a}
Ian~P. Gent and Toby Walsh.
\newblock How not to do it.
\newblock In \emph{AAAI Workshop on Experimental Evaluation of Reasoning and
  Search Methods}, 1994.

\bibitem[Gent et~al.(1997)Gent, Grant, MacIntyre, Prosser, Shaw, Smith, and
  Walsh]{GenGraMac1997hownotto}
Ian~P. Gent, Stuart~A. Grant, Ewen MacIntyre, Patrick Prosser, Paul Shaw,
  Barbara~M. Smith, and Toby Walsh.
\newblock How not to do it.
\newblock Technical Report 97.27, School of Computer Studies, University of
  Leeds, May 1997.

\bibitem[Glover(1989)]{glov89a}
Fred Glover.
\newblock Tabu search---part i.
\newblock \emph{ORSA Journal on computing}, 1\penalty0 (3):\penalty0 190--206,
  1989.

\bibitem[Goh et~al.(2015)Goh, Tan, Al-Mamun, and Abbass]{Goh2015}
Sim~Kuan Goh, Kay~Chen Tan, Abdullah Al-Mamun, and Hussein~A. Abbass.
\newblock Evolutionary big optimization ({BigOpt}) of signals.
\newblock In \emph{2015 {IEEE} Congress on Evolutionary Computation ({CEC})}.
  {IEEE}, {May} 2015.
\newblock \doi{10.1109/cec.2015.7257307}.
\newblock URL \url{https://doi.org/10.1109/cec.2015.7257307}.

\bibitem[Goldberg(1989)]{Gol89a}
David~E Goldberg.
\newblock \emph{{Genetic Algorithms in Search, Optimization, and Machine
  Learning}}.
\newblock Addison-Wesley, Reading MA, 1989.

\bibitem[Golden et~al.(1986)Golden, Assad, Wasil, and Baker]{Gold86a}
Bruce~L Golden, Arjang~A Assad, Edward~A Wasil, and Edward Baker.
\newblock Experimentation in optimization.
\newblock \emph{European Journal of Operational Research}, 27\penalty0
  (1):\penalty0 1--16, 1986.

\bibitem[Haftka(2016)]{Haft16b}
Raphael~T. Haftka.
\newblock Requirements for papers focusing on new or improved global
  optimization algorithms.
\newblock \emph{Structural and Multidisciplinary Optimization}, 54\penalty0
  (1):\penalty0 1--1, 2016.

\bibitem[Hains et~al.(2013)Hains, Whitley, Howe, and Chen]{HWHC2013HILSFM}
Doug Hains, L.~Darrell Whitley, Adele~E. Howe, and Wenxiang Chen.
\newblock {Hyperplane Initialized Local Search for MAXSAT}.
\newblock In \emph{{Proc.~of the Genetic and Evolutionary Computation
  Conference}}, pages 805~--~812. {ACM}, 2013.

\bibitem[Hansen(2000)]{Hansen2000invariance}
Nikolaus Hansen.
\newblock Invariance, self-adaptation and correlated mutations in evolution
  strategies.
\newblock In \emph{Proc. of International Conference on Parallel Problem
  Solving from Nature}, pages 355--364. Springer, 2000.
\newblock ISBN 978-3-540-45356-7.

\bibitem[Hansen et~al.(2003)Hansen, M{\"u}ller, and Koumoutsakos]{hans03a}
Nikolaus Hansen, Sibylle~D M{\"u}ller, and Petros Koumoutsakos.
\newblock Reducing the time complexity of the derandomized evolution strategy
  with covariance matrix adaptation (cma-es).
\newblock \emph{Evolutionary computation}, 11\penalty0 (1):\penalty0 1--18,
  2003.

\bibitem[Hansen et~al.(2012)Hansen, Auger, Finck, and Ros]{HAFR2012RPBBOBES}
Nikolaus Hansen, Anne Auger, Steffen Finck, and Raymond Ros.
\newblock Real-parameter black-box optimization benchmarking: Experimental
  setup.
\newblock Technical report, Universit{\'{e}} Paris Sud, INRIA Futurs,
  {\'{E}}quipe TAO, Orsay, France, {March}~24, 2012.
\newblock URL
  \url{http://coco.lri.fr/BBOB-downloads/download11.05/bbobdocexperiment.pdf}.

\bibitem[Hansen et~al.(2016{\natexlab{a}})Hansen, Auger, Brockhoff,
  Tu{\v{s}}ar, and Tu{\v{s}}ar]{HansenABTT16}
Nikolaus Hansen, Anne Auger, Dimo Brockhoff, Dejan Tu{\v{s}}ar, and Tea
  Tu{\v{s}}ar.
\newblock {COCO:} performance assessment.
\newblock \emph{CoRR}, abs/1605.03560, 2016{\natexlab{a}}.
\newblock URL \url{http://arxiv.org/abs/1605.03560}.

\bibitem[Hansen et~al.(2016{\natexlab{b}})Hansen, Auger, Mersmann, Tu\v{s}ar,
  and Brockhoff]{cocoplat}
Nikolaus Hansen, Anne Auger, Olaf Mersmann, Tea Tu\v{s}ar, and Dimo Brockhoff.
\newblock {COCO: A Platform for Comparing Continuous Optimizers in a Black-Box
  Setting}.
\newblock \emph{{arXiv preprint}}, abs/1603.08785v3, August 2016{\natexlab{b}}.
\newblock URL \url{http://arxiv.org/abs/1603.08785v3}.

\bibitem[Hansen et~al.(2020)Hansen, Akimoto, yoshihikoueno, Brockhoff, and
  Chan]{pycmaes}
Nikolaus Hansen, Youhei Akimoto, yoshihikoueno, Dimo Brockhoff, and Matthew
  Chan.
\newblock {CMA-ES}/pycma: r3.0.3, 2020.
\newblock URL \url{https://doi.org/10.5281/zenodo.3764210}.

\bibitem[Hart(2001)]{hart2001mann}
Anna Hart.
\newblock Mann-whitney test is not just a test of medians: differences in
  spread can be important.
\newblock \emph{Bmj}, 323\penalty0 (7309):\penalty0 391--393, 2001.

\bibitem[Heitk{\"o}tter and Beasley(1994)]{heit94a}
J{\"o}rg Heitk{\"o}tter and David Beasley.
\newblock The hitch-hiker's guide to evolutionary computation, 1994.

\bibitem[Hellwig and Beyer(2019)]{HB2018BEAFSORVCOACR}
Michael Hellwig and Hans-Georg Beyer.
\newblock {Benchmarking Evolutionary Algorithms For Single-Objective
  Real-valued Constrained Optimization -- A Critical Review}.
\newblock \emph{{Swarm and Evolutionary Computation}}, 44:\penalty0 927--944,
  2019.

\bibitem[Hooker(1994)]{Hook94a}
John~N Hooker.
\newblock {Needed: An Empirical Science of Algorithms}.
\newblock \emph{Operations research}, 42\penalty0 (2):\penalty0 201~--~212,
  1994.

\bibitem[Hooker(1996)]{Hoo1996joh}
John~N. Hooker.
\newblock Testing heuristics: We have it all wrong.
\newblock \emph{Journal of Heuristics}, 1\penalty0 (1):\penalty0 33--42, 1996.
\newblock \doi{10.1007/BF02430364}.

\bibitem[Hutter et~al.(2009)Hutter, Hoos, Leyton-Brown, and
  St{\"u}tzle]{HutHooLeyStu2009jair}
Frank Hutter, Holger~H. Hoos, Kevin Leyton-Brown, and Thomas St{\"u}tzle.
\newblock {ParamILS:} an automatic algorithm configuration framework.
\newblock \emph{Journal of Artificial Intelligence Research}, 36:\penalty0
  267--306, October 2009.

\bibitem[Hutter et~al.(2011)Hutter, Hoos, and Leyton-Brown]{hutt11}
Frank Hutter, Holger~H. Hoos, and Kevin Leyton-Brown.
\newblock {Sequential Model-Based Optimization for General Algorithm
  Configuration}.
\newblock In \emph{{International Conference on Learning and Intelligent
  Optimization}}, pages 507~--~523. Springer, 2011.

\bibitem[Hutter et~al.(2014)Hutter, L{\'o}pez-Ib{\'a}{\~n}ez, Fawcett,
  Lindauer, Hoos, Leyton-Brown, and St{\"u}tzle]{HutLopFaw2014lion}
Frank Hutter, Manuel L{\'o}pez-Ib{\'a}{\~n}ez, Chris Fawcett, Marius~Thomas
  Lindauer, Holger~H. Hoos, Kevin Leyton-Brown, and Thomas St{\"u}tzle.
\newblock {AClib}: a benchmark library for algorithm configuration.
\newblock In Panos~M. Pardalos, Mauricio G.~C. Resende, Chrysafis Vogiatzis,
  and Jose~L. Walteros, editors, \emph{Learning and Intelligent Optimization,
  8th International Conference, LION 8}, volume 8426 of \emph{Lecture Notes in
  Computer Science}, pages 36--40. Springer, Heidelberg, Germany, 2014.
\newblock \doi{10.1007/978-3-319-09584-4_4}.

\bibitem[Hutter et~al.(2019)Hutter, Kotthoff, and Vanschoren]{HutterBook19}
Frank Hutter, Lars Kotthoff, and Joaquin Vanschoren.
\newblock \emph{{Automated Machine Learning: Methods, Systems, Challenges}}.
\newblock Springer, 2019.

\bibitem[Jesus et~al.(2020)Jesus, Liefooghe, Derbel, and Paquete]{jesus2020}
Alexandre~D. Jesus, Arnaud Liefooghe, Bilel Derbel, and Lu\'{\i}s Paquete.
\newblock {Algorithm Selection of Anytime Algorithms}.
\newblock In \emph{Proc.~of the 2020 Genetic and Evolutionary Computation
  Conference}, pages 850~--~858. ACM, 2020.

\bibitem[Jin and Branke(2005)]{Jin05}
Yaochu Jin and J{\"u}rgen Branke.
\newblock {Evolutionary Optimization in Uncertain Environments -- A Survey}.
\newblock \emph{IEEE Transactions on Evolutionary Computation}, 9\penalty0
  (3):\penalty0 303~--~317, 2005.

\bibitem[Johnson et~al.(1989)Johnson, Aragon, McGeoch, and Schevon]{John89a}
David~S Johnson, Cecilia~R Aragon, Lyle~A McGeoch, and Catherine Schevon.
\newblock {Optimization by Simulated Annealing: An Experimental Evaluation.
  Part I, Graph Partitioning}.
\newblock \emph{{Operations Research}}, 37\penalty0 (6):\penalty0 865~--~892,
  1989.

\bibitem[Johnson et~al.(1991)Johnson, Aragon, McGeoch, and Schevon]{John91a}
David~S Johnson, Cecilia~R Aragon, Lyle~A McGeoch, and Catherine Schevon.
\newblock {Optimization by Simulated Annealing: An Experimental Evaluation.
  Part II, Graph Coloring and Number Partitioning}.
\newblock \emph{{Operations Research}}, 39\penalty0 (3):\penalty0 378~--~406,
  1991.

\bibitem[Johnson(2002{\natexlab{a}})]{J2002ATGTTEAOA}
David~Stifler Johnson.
\newblock {A Theoretician's Guide to the Experimental Analysis of Algorithms}.
\newblock In \emph{Proc.~of a DIMACS Workshop on Data Structures, Near Neighbor
  Searches, and Methodology: Fifth and Sixth DIMACS Implementation Challenges},
  volume~59 of \emph{DIMACS -- Series in Discrete Mathematics and Theoretical
  Computer Science}, pages 215--250, 2002{\natexlab{a}}.

\bibitem[Johnson(2002{\natexlab{b}})]{John96a}
David~Stifler Johnson.
\newblock {A Theoretician's Guide to the Experimental Analysis of Algorithms}.
\newblock \emph{{Data Structures, Near Neighbor Searches, and Methodology:
  Fifth and Sixth DIMACS Implementation Challenges}}, 59:\penalty0 215~--~250,
  2002{\natexlab{b}}.

\bibitem[Johnson and McGeoch(2002)]{JMG2004EAOHFTS}
David~Stifler Johnson and Lyle~A. McGeoch.
\newblock {Experimental Analysis of Heuristics for the STSP}.
\newblock In \emph{{The Traveling Salesman Problem and its Variations}},
  volume~12 of \emph{Combinatorial Optimization}, chapter~9, pages 369~--~443.
  Kluwer Academic Publishers, 2002.

\bibitem[Jones et~al.(1998)Jones, Schonlau, and Welch]{Jones1998}
Donald~R. Jones, Matthias Schonlau, and William~J. Welch.
\newblock Efficient global optimization of expensive black-box functions.
\newblock \emph{Journal of Global Optimization}, 13\penalty0 (4):\penalty0
  455--492, 1998.

\bibitem[Karafotias et~al.(2015)Karafotias, Hoogendoorn, and Eiben]{Kara15b}
Giorgos Karafotias, Mark Hoogendoorn, and {\'{A}}goston~E. Eiben.
\newblock Parameter control in evolutionary algorithms: Trends and challenges.
\newblock \emph{IEEE Transactions on Evolutionary Computation}, 19\penalty0
  (2):\penalty0 167--187, April 2015.

\bibitem[Kauffman(1993)]{93KAU00}
Stuart~A. Kauffman.
\newblock \emph{{The Origins of Order: Self-Organization and Selection in
  Evolution}}.
\newblock Oxford University Press, USA, 1993.

\bibitem[Kennedy and Eberhart(1995)]{kenn95a}
James Kennedy and Russell Eberhart.
\newblock Particle swarm optimization.
\newblock In \emph{Neural Networks, 1995. Proceedings., IEEE International
  Conference on}, volume~4, pages 1942--1948. IEEE, 1995.

\bibitem[Kerschke and Trautmann(2019{\natexlab{a}})]{KerschkeT19}
Pascal Kerschke and Heike Trautmann.
\newblock {Automated Algorithm Selection on Continuous Black-Box Problems by
  Combining Exploratory Landscape Analysis and Machine Learning}.
\newblock \emph{Evolutionary Computation (ECJ)}, 27\penalty0 (1):\penalty0
  99~--~127, 2019{\natexlab{a}}.

\bibitem[Kerschke and Trautmann(2019{\natexlab{b}})]{flacco2019}
Pascal Kerschke and Heike Trautmann.
\newblock {Comprehensive Feature-Based Landscape Analysis of Continuous and
  Constrained Optimization Problems Using the R-package flacco}.
\newblock In \emph{Applications in Statistical Computing}, pages 93~--~123.
  Springer, 2019{\natexlab{b}}.

\bibitem[Kerschke et~al.(2018{\natexlab{a}})Kerschke, Bossek, and
  Trautmann]{KBT2018POSOTAPIARSBOITS}
Pascal Kerschke, Jakob Bossek, and Heike Trautmann.
\newblock {Parameterization of State-of-the-Art Performance Indicators: A
  Robustness Study based on Inexact TSP Solvers}.
\newblock In \emph{{Proceedings of the Genetic and Evolutionary Computation
  Conference Companion}}, pages 1737~--~1744. {ACM}, 2018{\natexlab{a}}.

\bibitem[Kerschke et~al.(2018{\natexlab{b}})Kerschke, Kotthoff, Bossek, Hoos,
  and Trautmann]{KerschkeKBHT18}
Pascal Kerschke, Lars Kotthoff, Jakob Bossek, Holger~H. Hoos, and Heike
  Trautmann.
\newblock {Leveraging TSP Solver Complementarity through Machine Learning}.
\newblock \emph{{Evolutionary Computation (ECJ)}}, 26\penalty0 (4):\penalty0
  597~--~620, December 2018{\natexlab{b}}.

\bibitem[Kerschke et~al.(2019)Kerschke, Hoos, Neumann, and
  Trautmann]{kerschke_automated_2019}
Pascal Kerschke, Holger~H. Hoos, Frank Neumann, and Heike Trautmann.
\newblock {Automated Algorithm Selection: Survey and Perspectives}.
\newblock \emph{{Evolutionary Computation (ECJ)}}, 27:\penalty0 3~--~45, 2019.

\bibitem[Kirkpatrick et~al.(1983)Kirkpatrick, Gelatt, and
  Vecchi]{Kirkpatrick83}
Scott Kirkpatrick, C.~D. Gelatt, and M.~P. Vecchi.
\newblock Optimization by simulated annealing.
\newblock \emph{Science}, 220:\penalty0 671--680, 1983.

\bibitem[Kleijnen(1988)]{Klei88a}
Jack P.~C. Kleijnen.
\newblock {Analyzing Simulation Experiments with Common Random Numbers}.
\newblock \emph{{Management Science}}, 34\penalty0 (1):\penalty0 65~--~74,
  1988.

\bibitem[Kleijnen(2001)]{Klei11a}
Jack P.~C. Kleijnen.
\newblock {Experimental Design for Sensitivity Analysis of Simulation Models}.
\newblock Workingpaper, Operations Research, 2001.

\bibitem[Kleijnen(2015)]{Klei15a}
Jack P.~C. Kleijnen.
\newblock {Design and Analysis of Simulation Experiments}.
\newblock In \emph{{International Workshop on Simulation}}, pages 3--22.
  Springer, 2015.

\bibitem[Kleijnen(2017)]{Klei17a}
Jack P.~C. Kleijnen.
\newblock {Regression and Kriging Metamodels with their Experimental Designs in
  Simulation: A Review}.
\newblock \emph{{European Journal of Operational Research}}, 256\penalty0
  (1):\penalty0 1--16, 2017.

\bibitem[Kliemann and Sanders(2016)]{KliemannS16}
Lasse Kliemann and Peter Sanders.
\newblock \emph{{Algorithm Engineering: Selected Results and Surveys}}, volume
  9220.
\newblock Springer, 2016.

\bibitem[Kruskal and Wallis(1952)]{Krus52a}
William~H Kruskal and W~Allen Wallis.
\newblock Use of ranks in one-criterion variance analysis.
\newblock \emph{Journal of the American statistical Association}, 47\penalty0
  (260):\penalty0 583--621, 1952.

\bibitem[Kumar et~al.(2020)Kumar, Wu, Ali, Mallipeddi, Suganthan, and
  Das]{KWZMSD2020GFRWSOCOC}
Abhishek Kumar, Guohua Wu, Mostafa~Z. Ali, Rammohan Mallipeddi,
  Ponnuthurai~Nagaratnam Suganthan, and Swagatam Das.
\newblock Guidelines for real-world single-objective constrained optimisation
  competition.
\newblock Technical report, 2020.

\bibitem[Larra{\~n}aga and Lozano(2002)]{Larranaga2002EDA}
Pedro Larra{\~n}aga and Jos\'e~A. Lozano.
\newblock \emph{Estimation of Distribution Algorithms: A New Tool for
  Evolutionary Computation}.
\newblock Kluwer Academic Publishers, 2002.

\bibitem[Lehre and Witt(2012)]{LehreW12}
Per~Kristian Lehre and Carsten Witt.
\newblock Black-box search by unbiased variation.
\newblock \emph{Algorithmica}, 64:\penalty0 623--642, 2012.

\bibitem[Levene(1961)]{Leve61a}
Howard Levene.
\newblock Robust tests for equality of variances.
\newblock \emph{Contributions to probability and statistics. Essays in honor of
  Harold Hotelling}, pages 279--292, 1961.

\bibitem[Li et~al.(2017)Li, Jamieson, DeSalvo, Rostamizadeh, and
  Talwalkar]{hyperband}
Lisha Li, Kevin~G. Jamieson, Giulia DeSalvo, Afshin Rostamizadeh, and Ameet
  Talwalkar.
\newblock {Hyperband: A Novel Bandit-Based Approach to Hyperparameter
  Optimization}.
\newblock \emph{Journal of Machine Learning Research}, 18:\penalty0
  185:1--185:52, 2017.

\bibitem[Li et~al.(2006)Li, Emmerich, Eggermont, Bovenkamp, B{\"a}ck, Dijkstra,
  and Reiber]{LI2006}
Rui Li, Michael T.~M. Emmerich, Jeroen Eggermont, Ernst G.~P. Bovenkamp, Thomas
  B{\"a}ck, Jouke Dijkstra, and Johan H.~C. Reiber.
\newblock {Mixed-Integer NK Landscapes}.
\newblock In \emph{{Proc. of Parallel Problem Solving from Nature}}, pages
  42--51. Springer, 2006.

\bibitem[Liao et~al.(2014)Liao, Socha, {Montes de Oca}, St{\"u}tzle, and
  Dorigo]{LiaSocMonStuDor2014}
Tianjun Liao, Krzysztof Socha, Marco~A. {Montes de Oca}, Thomas St{\"u}tzle,
  and Marco Dorigo.
\newblock Ant colony optimization for mixed-variable optimization problems.
\newblock \emph{IEEE Transactions on Evolutionary Computation}, 18\penalty0
  (4):\penalty0 503--518, 2014.

\bibitem[Lindman(1974)]{Lind74a}
Harold~R Lindman.
\newblock \emph{Analysis of variance in complex experimental designs.}
\newblock WH Freeman \& Co, 1974.

\bibitem[Liu et~al.(2020)Liu, Moreau, Preuss, Rozi{\`{e}}re, Rapin, Teytaud,
  and Teytaud]{shiwa}
Jialin Liu, Antoine Moreau, Mike Preuss, Baptiste Rozi{\`{e}}re,
  J{\'{e}}r{\'{e}}my Rapin, Fabien Teytaud, and Olivier Teytaud.
\newblock Versatile black-box optimization.
\newblock \emph{CoRR}, abs/2004.14014, 2020.
\newblock URL \url{https://arxiv.org/abs/2004.14014}.

\bibitem[Liu et~al.(2019)Liu, Gehrlein, Wang, Yan, Cao, Chen, and Li]{Liu19b}
Qunfeng Liu, William~V. Gehrlein, Ling Wang, Yuan Yan, Yingying Cao, Wei Chen,
  and Yun Li.
\newblock {Paradoxes in Numerical Comparison of Optimization Algorithms}.
\newblock \emph{{IEEE Transactions on Evolutionary Computation}}, pages 1--15,
  2019.

\bibitem[L{\'o}pez-Ib{\'a}{\~n}ez et~al.(2016)L{\'o}pez-Ib{\'a}{\~n}ez,
  Dubois-Lacoste, P{\'e}rez~C{\'a}ceres, St{\"u}tzle, and
  Birattari]{LopDubPerStuBir2016irace}
Manuel L{\'o}pez-Ib{\'a}{\~n}ez, J{\'e}r{\'e}mie Dubois-Lacoste, Leslie
  P{\'e}rez~C{\'a}ceres, Thomas St{\"u}tzle, and Mauro Birattari.
\newblock The {\Rpackage{irace}} package: Iterated racing for automatic
  algorithm configuration.
\newblock \emph{Operations Research Perspectives}, 3:\penalty0 43--58, 2016.
\newblock \doi{10.1016/j.orp.2016.09.002}.

\bibitem[Malan and Engelbrecht(2013)]{13MAL00}
Katherine~Mary Malan and Andries~Petrus Engelbrecht.
\newblock {A Survey of Techniques for Characterising Fitness Landscapes and
  Some Possible Ways Forward}.
\newblock \emph{{Information Sciences (JIS)}}, 241:\penalty0 148~--~163, 2013.

\bibitem[Matou\v{s}ek(2009)]{Mat99}
Ji\v{r}{\'i} Matou\v{s}ek.
\newblock \emph{Geometric Discrepancy}.
\newblock Springer, Berlin, 2 edition, 2009.

\bibitem[Mayo and Spanos(2006)]{Mayo06a}
Deborah~G Mayo and Aris Spanos.
\newblock Severe testing as a basic concept in a neyman--pearson philosophy of
  induction.
\newblock \emph{The British Journal for the Philosophy of Science}, 57\penalty0
  (2):\penalty0 323--357, 2006.

\bibitem[McClymont and Keedwell(2011)]{MCCL2011}
Kent McClymont and Ed~Keedwell.
\newblock {Benchmark Multi-Objective Optimisation Test Problems with Mixed
  Encodings}.
\newblock In \emph{{Proceedings of the 2011 IEEE Congress on Evolutionary
  Computation}}, pages 2131~--~2138. IEEE, 2011.

\bibitem[McDermott(2020)]{McDermott2020nfl}
James McDermott.
\newblock When and why metaheuristics researchers can ignore "no free lunch"
  theorems.
\newblock \emph{{SN} Computer Science}, 1\penalty0 (60):\penalty0 1--18, 2020.
\newblock \doi{10.1007/s42979-020-0063-3}.

\bibitem[McGeoch(1986)]{McGe86a}
Catherine~C McGeoch.
\newblock \emph{{Experimental Analysis of Algorithms}}.
\newblock PhD thesis, Carnegie Mellon University, Pittsburgh PA, 1986.

\bibitem[McGeoch(1996)]{Mcge96a}
Catherine~C McGeoch.
\newblock Toward an experimental method for algorithm simulation.
\newblock \emph{INFORMS Journal on Computing}, 8\penalty0 (1):\penalty0 1--15,
  1996.

\bibitem[{McKay, Michael D and Beckman, Richard J and Conover, William
  J}(2000)]{LHS}
{McKay, Michael D and Beckman, Richard J and Conover, William J}.
\newblock {A Comparison of Three Methods for Selecting Values of Input
  Variables in the Analysis of Output from a Computer Code}.
\newblock \emph{Technometrics}, 42\penalty0 (1):\penalty0 55~--~61, 2000.

\bibitem[Mersmann et~al.(2010)Mersmann, Preuss, and Trautmann]{Mers10a}
Olaf Mersmann, Mike Preuss, and Heike Trautmann.
\newblock {Benchmarking Evolutionary Algorithms: Towards Exploratory Landscape
  Analysis}.
\newblock In \emph{Proc. of International Conference on Parallel Problem
  Solving from Nature}, pages 73~--~82. Springer, 2010.

\bibitem[Mersmann et~al.(2011)Mersmann, Bischl, Trautmann, Preuss, Weihs, and
  Rudolph]{mersmann_exploratory_2011}
Olaf Mersmann, Bernd Bischl, Heike Trautmann, Mike Preuss, Claus Weihs, and
  G{\"u}nter Rudolph.
\newblock {Exploratory Landscape Analysis}.
\newblock In \emph{{Proc.~of the 13th Annual Conference on Genetic and
  Evolutionary Computation}}, pages 829~--~836. ACM, 2011.

\bibitem[Mersmann et~al.(2013)Mersmann, Bischl, Trautmann, Wagner, Bossek, and
  Neumann]{Mersmann2013tspinstances}
Olaf Mersmann, Bernd Bischl, Heike Trautmann, Markus Wagner, Jakob Bossek, and
  Frank Neumann.
\newblock {A Novel Feature-Based Approach to Characterize Algorithm Performance
  for the Traveling Salesperson Problem}.
\newblock \emph{Annals of Mathematics and Artificial Intelligence}, 69\penalty0
  (2):\penalty0 151~--~182, 2013.

\bibitem[Mladenovi\'{c} and Hansen(1997)]{VariableNeighborhoodSearch}
Nenad Mladenovi\'{c} and Pierre Hansen.
\newblock Variable neighborhood search.
\newblock \emph{Comput. Oper. Res.}, 24\penalty0 (11):\penalty0 1097–1100,
  1997.
\newblock ISSN 0305-0548.
\newblock \doi{10.1016/S0305-0548(97)00031-2}.
\newblock URL \url{https://doi.org/10.1016/S0305-0548(97)00031-2}.

\bibitem[Montgomery(2017)]{Mont17a}
Douglas~C Montgomery.
\newblock \emph{{Design and Analysis of Experiments}}.
\newblock John Wiley \& Sons, 9 edition, 2017.

\bibitem[Mor{\'e} and Wild(2009)]{More09a}
Jorge~J Mor{\'e} and Stefan~M Wild.
\newblock Benchmarking derivative-free optimization algorithms.
\newblock \emph{SIAM Journal on Optimization}, 20\penalty0 (1):\penalty0
  172--191, 2009.

\bibitem[Mor{\'e} et~al.(1981)Mor{\'e}, Garbow, and Hillstrom]{More81a}
Jorge~J Mor{\'e}, Burton~S Garbow, and Kenneth~E Hillstrom.
\newblock {Testing Unconstrained Optimization Software}.
\newblock \emph{{ACM Transactions on Mathematical Software (TOMS)}}, 7\penalty0
  (1):\penalty0 17~--~41, 1981.

\bibitem[{Mu{\~n}oz Acosta} et~al.(2015{\natexlab{a}}){Mu{\~n}oz Acosta},
  Kirley, and Halgamuge]{munoz_exploratory_2015}
Mario~Andr{\'e}s {Mu{\~n}oz Acosta}, Michael Kirley, and Saman~K. Halgamuge.
\newblock {Exploratory Landscape Analysis of Continuous Space Optimization
  Problems Using Information Content}.
\newblock \emph{{IEEE Transactions on Evolutionary Computation (TEVC)}},
  19\penalty0 (1):\penalty0 74~--~87, 2015{\natexlab{a}}.

\bibitem[{Mu{\~n}oz Acosta} et~al.(2015{\natexlab{b}}){Mu{\~n}oz Acosta}, Sun,
  Kirley, and Halgamuge]{munoz2015_as}
Mario~Andr{\'e}s {Mu{\~n}oz Acosta}, Yuan Sun, Michael Kirley, and Saman~K.
  Halgamuge.
\newblock {Algorithm Selection for Black-Box Continuous Optimization Problems:
  A Survey on Methods and Challenges}.
\newblock \emph{{Information Sciences (JIS)}}, 317:\penalty0 224~--~245,
  2015{\natexlab{b}}.

\bibitem[M{\"u}hlenbein and Paa{\ss}(1996)]{EDAMuehlenbein}
H.~M{\"u}hlenbein and G.~Paa{\ss}.
\newblock From recombination of genes to the estimation of distributions i.
  binary parameters.
\newblock In Hans-Michael Voigt, Werner Ebeling, Ingo Rechenberg, and Hans-Paul
  Schwefel, editors, \emph{Proc. of Parallel Problem Solving from Nature
  (PPSN'96)}, pages 178--187. Springer, 1996.
\newblock ISBN 978-3-540-70668-7.

\bibitem[M{\"u}ller-Hannemann and Schirra(2010)]{Muel10a}
Matthias M{\"u}ller-Hannemann and Stefan Schirra.
\newblock \emph{{Algorithm Engineering: Bridging the Gap Between Algorithm
  Theory and Practice}}.
\newblock Springer, 2010.

\bibitem[Mu{\~{n}}oz and Smith-Miles(2017)]{MS2017PAOC}
Mario~A. Mu{\~{n}}oz and Kate~A. Smith-Miles.
\newblock {Performance Analysis of Continuous Black-Box Optimization Algorithms
  via Footprints in Instance Space}.
\newblock \emph{Evolutionary Computation}, 25\penalty0 (4):\penalty0 529--554,
  {December} 2017.

\bibitem[Nazzal et~al.(2012)Nazzal, Mollaghasemi, Hedlund, and
  Bozorgi]{Nazz11a}
Dima Nazzal, Mansooreh Mollaghasemi, H~Hedlund, and A~Bozorgi.
\newblock {Using Genetic Algorithms and an Indifference-Zone Ranking and
  Selection Procedure Under Common Random Numbers for Simulation Optimisation}.
\newblock \emph{{Journal of Simulation}}, 6\penalty0 (1):\penalty0 56~--~66,
  2012.

\bibitem[Nelder and Mead(1965)]{Neld65a}
John~A Nelder and Roger Mead.
\newblock A simplex method for function minimization.
\newblock \emph{The computer journal}, 7\penalty0 (4):\penalty0 308--313, 1965.

\bibitem[Neumann et~al.(2019)Neumann, Gao, Wagner, and
  Neumann]{Neumann2019diversity}
Aneta Neumann, Wanru Gao, Markus Wagner, and Frank Neumann.
\newblock Evolutionary diversity optimization using multi-objective indicators.
\newblock In \emph{Proceedings of the Genetic and Evolutionary Computation
  Conference}, GECCO ’19, page 837–845, New York, NY, USA, 2019.
  Association for Computing Machinery.
\newblock ISBN 9781450361118.
\newblock \doi{10.1145/3321707.3321796}.
\newblock URL \url{https://doi.org/10.1145/3321707.3321796}.

\bibitem[Neumann and Witt(2010)]{NeumannW10}
Frank Neumann and Carsten Witt.
\newblock \emph{Bioinspired {C}omputation in {C}ombinatorial {O}ptimization --
  {A}lgorithms and {T}heir {C}omputational {C}omplexity}.
\newblock Springer, 2010.

\bibitem[Nguyen et~al.(2012)Nguyen, Yang, and Branke]{Nguyen2012}
Trung~Thanh Nguyen, Shengxiang Yang, and Juergen Branke.
\newblock Evolutionary dynamic optimization: A survey of the state of the art.
\newblock \emph{Swarm and Evolutionary Computation}, 6:\penalty0 1--24,
  {October} 2012.
\newblock \doi{10.1016/j.swevo.2012.05.001}.
\newblock URL \url{https://doi.org/10.1016/j.swevo.2012.05.001}.

\bibitem[Nosek et~al.(2018)Nosek, Ebersole, DeHaven, and
  Mellor]{NosEbeHav2018preregistration}
Brian~A. Nosek, Charles~R. Ebersole, Alexander~C. DeHaven, and David~T. Mellor.
\newblock The preregistration revolution.
\newblock \emph{Proceedings of the National Academy of Sciences}, 115\penalty0
  (11):\penalty0 2600--2606, March 2018.
\newblock ISSN 0027-8424, 1091-6490.
\newblock \doi{10.1073/pnas.1708274114}.

\bibitem[Olson and Moore(2016)]{olson2016tpot}
Randal~S Olson and Jason~H Moore.
\newblock {TPOT}: A tree-based pipeline optimization tool for automating
  machine learning.
\newblock In \emph{Workshop on automatic machine learning}, pages 66--74, 2016.

\bibitem[Olson et~al.(2017)Olson, La~Cava, Orzechowski, Urbanowicz, and
  Moore]{olson2017pmlb}
Randal~S Olson, William La~Cava, Patryk Orzechowski, Ryan~J Urbanowicz, and
  Jason~H Moore.
\newblock Pmlb: a large benchmark suite for machine learning evaluation and
  comparison.
\newblock \emph{BioData mining}, 10\penalty0 (1):\penalty0 1--13, 2017.

\bibitem[Orzechowski et~al.(2018)Orzechowski, La~Cava, and
  Moore]{orzechowski2018we}
Patryk Orzechowski, William La~Cava, and Jason~H Moore.
\newblock Where are we now? a large benchmark study of recent symbolic
  regression methods.
\newblock In \emph{Proceedings of the Genetic and Evolutionary Computation
  Conference}, pages 1183--1190, 2018.

\bibitem[Orzechowski et~al.(2020)Orzechowski, Magiera, and
  Moore]{orzechowski2020benchmarking}
Patryk Orzechowski, Franciszek Magiera, and Jason~H Moore.
\newblock Benchmarking manifold learning methods on a large collection of
  datasets.
\newblock In \emph{European Conference on Genetic Programming (Part of
  EvoStar)}, pages 135--150. Springer, 2020.

\bibitem[Packebusch and Mertens(2016)]{Packebusch_2016_LABS}
Tom Packebusch and Stephan Mertens.
\newblock Low autocorrelation binary sequences.
\newblock \emph{Journal of Physics A: Mathematical and Theoretical},
  49\penalty0 (16):\penalty0 165001, 2016.
\newblock \doi{10.1088/1751-8113/49/16/165001}.
\newblock URL \url{https://doi.org/10.1088%2F1751-8113%2F49%2F16%2F165001}.

\bibitem[Paenke et~al.(2006)Paenke, Branke, and Jin]{Paen06a}
Ingo Paenke, J{\"u}rgen Branke, and Yaochu Jin.
\newblock Efficient search for robust solutions by means of evolutionary
  algorithms and fitness approximation.
\newblock \emph{IEEE Trans. Evolutionary Computation}, 10\penalty0
  (4):\penalty0 405--420, 2006.

\bibitem[Pan and Yang(2010)]{Pan10a}
Sinno~Jialin Pan and Qiang Yang.
\newblock {A Survey on Transfer Learning}.
\newblock \emph{{IEEE Transactions on Knowledge and Data Engineering}},
  22\penalty0 (10):\penalty0 1345--1359, Oct 2010.

\bibitem[Pesarin(2001)]{Pesa01a}
Fortunato Pesarin.
\newblock \emph{Multivariate permutation tests: with applications in
  biostatistics}, volume 240.
\newblock Wiley Chichester, 2001.

\bibitem[Plackett and Burman(1946)]{Plac46a}
Robin~L Plackett and J~Peter Burman.
\newblock The design of optimum multifactorial experiments.
\newblock \emph{Biometrika}, 33\penalty0 (4):\penalty0 305--325, 1946.

\bibitem[Polyakovskiy et~al.(2014)Polyakovskiy, Bonyadi, Wagner, Michalewicz,
  and Neumann]{Polyakovskiy2014ttplib}
Sergey Polyakovskiy, Mohammad~Reza Bonyadi, Markus Wagner, Zbigniew
  Michalewicz, and Frank Neumann.
\newblock {A Comprehensive Benchmark Set and Heuristics for the Traveling Thief
  Problem}.
\newblock In \emph{Proc. of the 2014 Annual Conference on Genetic and
  Evolutionary Computation}, pages 477~–--~484. ACM, 2014.
\newblock ISBN 9781450326629.

\bibitem[Popper(1959)]{Popp59a}
Karl~Raimund Popper.
\newblock \emph{{The Logic of Scientific Discovery}}.
\newblock Hutchinson \& Co, 2 edition, 1959.

\bibitem[Popper(1975)]{Popp79a}
Karl~Raimund Popper.
\newblock \emph{{Objective Knowledge: An Evolutionary Approach}}.
\newblock Oxford University Press, 1975.

\bibitem[Preuss(2015)]{Preu15a}
Mike Preuss.
\newblock \emph{Experimentation in Evolutionary Computation}, pages 27--54.
\newblock Springer, 2015.

\bibitem[Price(1997)]{P1997DEVTFOT2I}
Kenneth~V. Price.
\newblock {Differential Evolution vs.~The Functions of the
  2\textsuperscript{nd} \mbox{ICEO}}.
\newblock In \emph{{Proc. of the IEEE International Conference on Evolutionary
  Computation}}, pages 153~--~157. {IEEE}, 1997.

\bibitem[Pukelsheim(1993)]{Puke93a}
F.~Pukelsheim.
\newblock \emph{Optimal Design of Experiments}.
\newblock Wiley, New York NY, 1993.

\bibitem[Rapin and Teytaud(2018)]{nevergrad}
Jeremy Rapin and Olivier Teytaud.
\newblock {Nevergrad - A gradient-free optimization platform}.
\newblock \url{https://GitHub.com/FacebookResearch/Nevergrad}, 2018.

\bibitem[Rosenbrock(1960)]{Rose60a}
Howard~Harry Rosenbrock.
\newblock {An Automatic Method for Finding the Greatest or Least Value of a
  Function}.
\newblock \emph{{The Computer Journal}}, 3\penalty0 (3):\penalty0 175~--~184,
  1960.

\bibitem[Rowe and Vose(2011)]{ABB}
Jonathan Rowe and Michael Vose.
\newblock Unbiased black box search algorithms.
\newblock In \emph{Proc. of Genetic and Evolutionary Computation Conference},
  pages 2035~--~2042. ACM, 2011.

\bibitem[Roy(2001)]{Roy01a}
Ranjit~K Roy.
\newblock \emph{Design of experiments using the Taguchi approach: 16 steps to
  product and process improvement}.
\newblock John Wiley \& Sons, 2001.

\bibitem[Sachdeva et~al.(2020)Sachdeva, Neumann, and
  Wagner]{sachdeva2020dynamic}
Ragav Sachdeva, Frank Neumann, and Markus Wagner.
\newblock The dynamic travelling thief problem: Benchmarks and performance of
  evolutionary algorithms, 2020.

\bibitem[Santner et~al.(2003)Santner, Williams, Notz, and Williams]{Sant03a}
Thomas~J Santner, Brian~J Williams, William~I Notz, and Brain~J Williams.
\newblock \emph{The design and analysis of computer experiments}, volume~1.
\newblock Springer, 2003.

\bibitem[Schwefel(1975)]{Schw75a}
Hans-Paul Schwefel.
\newblock \emph{{Evolutionsstrategie und numerische Optimierung}}.
\newblock PhD thesis, Technische Universit{\"a}t Berlin, Fachbereich
  Verfahrenstechnik, Berlin, Germany, 1975.

\bibitem[Shanno(1970)]{shanno1970}
David~F Shanno.
\newblock Conditioning of quasi-newton methods for function minimization.
\newblock \emph{Mathematics of computation}, 24\penalty0 (111):\penalty0
  647--656, 1970.

\bibitem[Shapiro and Wilk(1965)]{Shap65a}
Samuel~Sanford Shapiro and Martin~B Wilk.
\newblock An analysis of variance test for normality (complete samples).
\newblock \emph{Biometrika}, 52\penalty0 (3/4):\penalty0 591--611, 1965.

\bibitem[Sheskin(2003)]{Shes03a}
David~J Sheskin.
\newblock \emph{Handbook of parametric and nonparametric statistical
  procedures}.
\newblock crc Press, 2003.

\bibitem[Shi and Eberhart(1998)]{shi98a}
Yuhui Shi and Russell Eberhart.
\newblock {A Modified Particle Swarm Optimizer}.
\newblock In \emph{{Proc. of the 1998 IEEE International Conference on
  Evolutionary Computation, within the IEEE World Congress on Computational
  Intelligence}}, pages 69--73. IEEE, 1998.

\bibitem[Shir et~al.(2018)Shir, Doerr, and B{\"{a}}ck]{ShirDB18}
Ofer~M. Shir, Carola Doerr, and Thomas B{\"{a}}ck.
\newblock {Compiling a Benchmarking Test-Suite for Combinatorial Black-Box
  Optimization: A Position Paper}.
\newblock In \emph{Proc. of Genetic and Evolutionary Computation Conference},
  pages 1753~--~1760. ACM, 2018.

\bibitem[{\v S}kvorc et~al.(2020){\v S}kvorc, Eftimov, and Koro{\v
  s}ec]{SEK2020UTPS}
Urban {\v S}kvorc, Tome Eftimov, and Peter Koro{\v s}ec.
\newblock {Understanding the Problem Space in Single-Objective Numerical
  Optimization Using Exploratory Landscape Analysis}.
\newblock \emph{{Applieed Soft Computing (ASOC)}}, 90:\penalty0 106138, 2020.

\bibitem[Smith-Miles and Bowly(2015)]{SB2015GNTI}
Kate Smith-Miles and Simon Bowly.
\newblock Generating new test instances by evolving in instance space.
\newblock \emph{Computers {\&} Operations Research}, 63:\penalty0 102--113,
  {November} 2015.
\newblock \doi{10.1016/j.cor.2015.04.022}.
\newblock URL \url{https://doi.org/10.1016/j.cor.2015.04.022}.

\bibitem[Smith-Miles and Tan(2012)]{ST2012MAF}
Kate Smith-Miles and Thomas~T. Tan.
\newblock {Measuring Algorithm Footprints in Instance Space}.
\newblock In \emph{{2012 IEEE Congress on Evolutionary Computation}}. {IEEE},
  2012.

\bibitem[Socha and Dorigo(2008)]{soch08a}
Krzysztof Socha and Marco Dorigo.
\newblock Ant colony optimization for continuous domains.
\newblock \emph{European journal of operational research}, 185\penalty0
  (3):\penalty0 1155--1173, 2008.

\bibitem[Stork et~al.(2020)Stork, Eiben, and Bartz-Beielstein]{stork18a}
J{\"o}rg Stork, A.~E. Eiben, and Thomas Bartz-Beielstein.
\newblock A new taxonomy of global optimization algorithms, Nov 2020.
\newblock ISSN 1572-9796.
\newblock URL \url{https://doi.org/10.1007/s11047-020-09820-4}.

\bibitem[Talbi(2009)]{T2009MFDTI}
El-Ghazali Talbi.
\newblock \emph{Metaheuristics: From Design to Implementation}.
\newblock John Wiley \& Sons Inc., {July} 2009.
\newblock ISBN 978-0-470-27858-1.

\bibitem[Tanabe and Ishibuchi(2020)]{TI2020AETURWMOOPS}
Ryoji Tanabe and Hisao Ishibuchi.
\newblock {An Easy-to-Use Real-World Multi-Objective Optimization Problem
  Suite}.
\newblock \emph{{Applied Soft Computing (ASOC)}}, 89:\penalty0 106078, 2020.

\bibitem[Tsutsui et~al.(1996)Tsutsui, Ghosh, and Fujimoto]{tsutsui1996robust}
Shigeyoshi Tsutsui, Ashish Ghosh, and Yoshiji Fujimoto.
\newblock {A Robust Solution Searching Scheme in Genetic Search}.
\newblock In \emph{{International Conference on Parallel Problem Solving from
  Nature}}, pages 543~--~552. Springer, 1996.

\bibitem[Tukey(1977)]{Tuke77a}
John~Wilder Tukey.
\newblock \emph{{Exploratory Data Analysis}}, volume~2.
\newblock Reading, MA, 1977.

\bibitem[Tu\v{s}ar et~al.(2019)Tu\v{s}ar, Brockhoff, and Hansen]{Tusar2019}
Tea Tu\v{s}ar, Dimo Brockhoff, and Nikolaus Hansen.
\newblock Mixed-integer benchmark problems for single- and bi-objective
  optimization.
\newblock In \emph{Proceedings of the Genetic and Evolutionary Computation
  Conference}, pages 718~–--~726. ACM, 2019.

\bibitem[Ueno et~al.(2016)Ueno, Rhone, Hou, Mizoguchi, and
  Tsuda]{ueno2016combo}
Tsuyoshi Ueno, Trevor~David Rhone, Zhufeng Hou, Teruyasu Mizoguchi, and Koji
  Tsuda.
\newblock {COMBO}: an efficient {B}ayesian optimization library for materials
  science.
\newblock \emph{Materials discovery}, 4:\penalty0 18--21, 2016.

\bibitem[Ve{\v{c}}ek et~al.(2014)Ve{\v{c}}ek, Mernik, and
  {\v{C}}repin{\v{s}}ek]{vevcek2014chess}
Niki Ve{\v{c}}ek, Marjan Mernik, and Matej {\v{C}}repin{\v{s}}ek.
\newblock A chess rating system for evolutionary algorithms: A new method for
  the comparison and ranking of evolutionary algorithms.
\newblock \emph{Information Sciences}, 277:\penalty0 656--679, 2014.

\bibitem[Volz et~al.(2019)Volz, Naujoks, Kerschke, and
  Tu{\v{s}}ar]{volz2019single}
Vanessa Volz, Boris Naujoks, Pascal Kerschke, and Tea Tu{\v{s}}ar.
\newblock {Single- and Multi-Objective Game-Benchmark for Evolutionary
  Algorithms}.
\newblock In \emph{{Proceedings of the Genetic and Evolutionary Computation
  Conference}}, pages 647~--~655. ACM, 2019.

\bibitem[Vrielink and van~den Berg(2019)]{vrielink2019fireworks}
Wouter Vrielink and Daan van~den Berg.
\newblock Fireworks algorithm versus plant propagation algorithm.
\newblock In \emph{IJCCI}, pages 101--112, 2019.

\bibitem[Wagner(2010)]{Wagn10a}
Tobias Wagner.
\newblock A subjective review of the state of the art in model-based parameter
  tuning.
\newblock In Thomas Bartz-Beielstein, Marco Chiarandini, Luis Paquete, and Mike
  Preuss, editors, \emph{Workshop on Experimental Methods for the Assessment of
  Computational Systems (WEMACS 2010)}, Algorithm Engineering Report, pages
  1--13. TU Dortmund, Faculty of Computer Science, Algorithm Engineering
  (Ls11), 2010.

\bibitem[Wang et~al.(2020)Wang, Vermetten, Ye, Doerr, and
  B{\"{a}}ck]{IOHanalyzer}
Hao Wang, Diederick Vermetten, Furong Ye, Carola Doerr, and Thomas B{\"{a}}ck.
\newblock Iohanalyzer: Performance analysis for iterative optimization
  heuristic.
\newblock \emph{CoRR}, abs/2007.03953, 2020.
\newblock URL \url{https://arxiv.org/abs/2007.03953}.
\newblock A benchmark data repository is available through the web-based GUI at
  \url{iohprofiler.liacs.nl/}.

\bibitem[{Weinand} et~al.(2020){Weinand}, {S{\"o}rensen}, {San Segundo},
  {Kleinebrahm}, and {McKenna}]{Wein20a}
Jann~Michael {Weinand}, Kenneth {S{\"o}rensen}, Pablo {San Segundo}, Max
  {Kleinebrahm}, and Russell {McKenna}.
\newblock {Research trends in combinatorial optimisation}.
\newblock \emph{arXiv e-prints}, art. arXiv:2012.01294, {December} 2020.

\bibitem[Weise(2019)]{W2019JRDAIOTJSSP}
Thomas Weise.
\newblock jsspinstancesandresults: Results, data, and instances of the job shop
  scheduling problem, 2019.
\newblock URL \url{http://github.com/thomasWeise/jsspInstancesAndResults/}.
\newblock A meta-study of 145 algorithm setups from literature on the JSSP.

\bibitem[Weise and Wu(2018)]{Weise2018wmodel}
Thomas Weise and Zijun Wu.
\newblock Difficult features of combinatorial optimization problems and the
  tunable w-model benchmark problem for simulating them.
\newblock In \emph{Proceedings of the Genetic and Evolutionary Computation
  Conference Companion}, GECCO ’18, pages 1769~--–~1776. ACM, 2018.
\newblock ISBN 9781450357647.

\bibitem[Weise et~al.(2014)Weise, Chiong, Tang, L{\"a}ssig, Tsutsui, Chen,
  Michalewicz, and Yao]{WCTLTCMY2014BOAAOSFFTTSP}
Thomas Weise, Raymond Chiong, Ke~Tang, J{\"o}rg L{\"a}ssig, Shigeyoshi Tsutsui,
  Wenxiang Chen, Zbigniew Michalewicz, and Xin Yao.
\newblock Benchmarking optimization algorithms: An open source framework for
  the traveling salesman problem.
\newblock \emph{{{IEEE} Computational Intelligence Magazine ({CIM})}},
  9:\penalty0 40--52, {August} 2014.

\bibitem[Whitley et~al.(1996)Whitley, Rana, Dzubera, and Mathias]{WhitleyRDM96}
L.~Darrell Whitley, Soraya~B. Rana, John Dzubera, and Keith~E. Mathias.
\newblock {Evaluating Evolutionary Algorithms}.
\newblock \emph{Artificial Intelligence (AIJ)}, 85\penalty0 (1-2):\penalty0
  245~--~276, 1996.

\bibitem[Whitley et~al.(2002)Whitley, Watson, Howe, and Barbulescu]{Whit02b}
L.~Darrell Whitley, Jean-Paul Watson, Adele Howe, and Laura Barbulescu.
\newblock {Testing, Evaluation and Performance of Optimization and Learning
  Systems}.
\newblock In Ian~C. Parmee, editor, \emph{Adaptive Computing in Design and
  Manufacture V}, pages 27~--~39, London, 2002. Springer.

\bibitem[Wilcoxon(1945)]{Wilc45a}
Frank Wilcoxon.
\newblock Individual comparisons by ranking methods.
\newblock \emph{Biometrics bulletin}, 1\penalty0 (6):\penalty0 80--83, 1945.

\bibitem[Wol\-pert and Macready(1997)]{Wolp97a}
David~H. Wol\-pert and William~G. Macready.
\newblock {No Free Lunch Theorems for Optimization}.
\newblock \emph{IEEE Transactions on Evolutionary Computation}, 1\penalty0
  (1):\penalty0 67--82, {April} 1997.

\bibitem[Wu et~al.(2017)Wu, Mallipeddi, and Suganthan]{WMS2017PDAECFTCCOCRPO}
Guohua Wu, Rammohan Mallipeddi, and Ponnuthurai~Nagaratnam Suganthan.
\newblock Problem definitions and evaluation criteria for the {CEC~2017}
  competition on constrained real-parameter optimization.
\newblock Technical report, National University of Defense Technology,
  Changsha, Hunan, PR China and Kyungpook National University, Daegu, South
  Korea and Nanyang Technological University, Singapore, {September} 2017.

\bibitem[Xu et~al.(2008)Xu, Hutter, Hoos, and
  Leyton-Brown]{XuHutHooLey2008jair}
Lin Xu, Frank Hutter, Holger~H. Hoos, and Kevin Leyton-Brown.
\newblock {SATzilla:} portfolio-based algorithm selection for {SAT}.
\newblock \emph{Journal of Artificial Intelligence Research}, 32:\penalty0
  565--606, June 2008.

\end{thebibliography}
\end{small}

\end{document}